\RequirePackage{snapshot}
\RequirePackage{fix-cm}
\documentclass[smallcondensed]{svjour3}

\usepackage[utf8]{inputenc} 
\usepackage{hyperref}       
\usepackage{url}            
\usepackage{booktabs}       
\usepackage{amsfonts}       
\usepackage{amsmath}
\usepackage{amssymb}
\usepackage{nicefrac}       
\usepackage{microtype}      
\usepackage{subfig}
\usepackage{algorithm}
\usepackage[noend]{algpseudocode} 
\usepackage{array}
\usepackage{color}
\usepackage[export]{adjustbox}
\usepackage{multirow}
\usepackage[numbers]{natbib}

\usepackage{xcolor}
\usepackage{soul}

\setulcolor{blue}
\setstcolor{red}
\sethlcolor{green}

\definecolor{mpigreen} {RGB} {0, 129, 122}
\usepackage{graphicx}
\usepackage{subfig}

\newcommand{\argmax}{\operatornamewithlimits{argmax}}

\usepackage{bm}
\newcommand{\nodeset}[1]{\bm{\mathsf{#1}}}

\definecolor{lacamdarklilac5} {RGB} {51, 10, 102}
\definecolor{lacamgold5} {RGB} {255, 87, 0}
\definecolor{violet} {RGB} {119, 111, 178}
\definecolor{petroil2} {RGB} {36, 165, 175}
\definecolor{petroil4} {RGB} {30, 132, 149}
\definecolor{petroil6} {RGB} {23, 101, 115}
\definecolor{gold2} {RGB} {255, 130, 0}
\definecolor{gold4} {RGB} {250, 100, 0}
\definecolor{gold6} {RGB} {245, 90, 0}

\newcommand{\highlight}[2][yellow]{\mathchoice%
  {\colorbox{#1}{\textcolor{white}{$\displaystyle#2$}}}%
  {\colorbox{#1}{\textcolor{white}{$\textstyle#2$}}}%
  {\colorbox{#1}{\textcolor{white}{$\scriptstyle#2$}}}%
  {\colorbox{#1}{\textcolor{white}{$\scriptscriptstyle#2$}}}}%

\begin{document}







\title{Visualizing and understanding Sum-Product Networks}
\titlerunning{Visualizing and understanding Sum-Product Networks}  

\authorrunning{Antonio Vergari \and Nicola Di Mauro \and Floriana Esposito}

\author{Antonio Vergari\and Nicola Di Mauro\and Floriana Esposito
}
\institute{Antonio Vergari \at Dept. of Empirical Inference, Max Planck Institute for Intelligent Systems, T\"ubingen, Germany
  \and Nicola Di Mauro\and Floriana Esposito
  \at Dept. of Computer Science, University of Bari, Bari, Italy \\
  \email{antonio.vergari@tuebingen.mpg.de, nicola.dimauro@uniba.it, floriana.esposito@uniba.it} \\}

\journalname{Machine Learning Journal}
\date{August 2018}


\maketitle              

\begin{abstract}
  Sum-Product Networks (SPNs) are  deep tractable probabilistic
  models by which several kinds of inference queries can be answered
  exactly and in a tractable time.
  %
  %
  They have been largely used as black box density
  estimators, assessed  by comparing their
  likelihood scores on different tasks.
  In this paper we explore and exploit the inner representations learned by
  SPNs.
  By taking a closer look at the inner workings of SPNs, we aim to
  better understand what and how meaningful the representations they learn are, as in a classic Representation Learning framework.
  %
  %
  We firstly propose an interpretation of SPNs as Multi-Layer
  Perceptrons, we then devise several criteria to extract
  representations from SPNs and finally we empirically evaluate them
  in several (semi-)supervised tasks showing they are competitive
  against classical feature extractors like RBMs, DBNs and deep probabilistic autoencoders, like MADEs and VAEs.
  %
  %
  
\end{abstract}

\section{Introduction}

Density estimation is the unsupervised task of learning an estimator for a joint probability distribution over a set of random variables 
(RVs) from a set of samples. 
%
Such an estimator can be used to do
\emph{inference}---computing the probability of queries over those RVs. 
Many machine learning (ML) problems can be reframed as different kinds of
probabilistic inference tasks, e.g., classifying a target RV can be solved
by Most Probable Explanation (MPE) inference~\cite{Koller2009}.
%
Density estimation can be
thought as one of the most general tasks in ML.
%
%

Sum-Product Networks
(SPNs)~\cite{Poon2011} are tractable density estimators
compiling a joint probability distribution into a deep architecture.
While for classical density estimators such as Probabilistic Graphical
Models (PGMs)~\cite{Koller2009}, like
 Markov Networks (MNs) and Bayesian Networks (BNs), performing exact
 inference is generally unfeasible---it is exponential in the model treewidth---for SPNs
many kinds of queries like  marginal and conditional probabilities 
 are computable in linear time
in the size of the network~\cite{Poon2011}.
This is achievable by the presence of structural constraints like \emph{decomposability} and
 \emph{completeness}~\cite{Poon2011,Peharz2015a}, regarding the
 \emph{scope} of the nodes---the RVs appearing in the distributions
 modeled by those nodes.
%
SPNs have been successfully
employed in several applications, such as computer vision~\cite{Gens2012,Peharz2013},
speech recognition~\cite{Peharz2014a}, natural language processing~\cite{Cheng2014,Molina2017}
and
activity recognition~\cite{Amer2015}. 
%
The task of learning an SPN has been tackled both in the
weight~\cite{Poon2011,Rashwan2016,Zhao2016a}
and structure learning
scenarios~\cite{Gens2013,Rooshenas2014,Dennis2015,Rahman2016,DBLP:journals/corr/HsuKP17}. 

Up to now, however, SPNs have  been evaluated only as \emph{black box} inference
machines, i.e., only their output---the answer to a probabilistic query---is actually
exploited in the task considered.
In this paper we aim to \emph{uncover the inner workings} of these probabilistic
models by i) extending them
towards Representation Learning (RL)~\cite{6472238} and ii) bridging
themselves closer to deep neural models.
%
%
By leveraging the learned inner representations of a
model, RL approaches aim
to disentangle and uncover different explanatory
factors behind the data~\cite{6472238}.
Usually, one employs these \emph{embeddings} as features in 
predictive tasks later, e.g., Restricted Boltzmann
Machines (RBMs)~\cite{Smolensky1986} have been employed as feature
extractors after being unsupervisedly
trained~\cite{Coates2011,Marlin2010}, or representations from 
neural autoencoders---trained  to reconstruct the data---are
classically used as highly predictive features~\cite{Vincent2010,Hinton2006}.
%
%

In this work, we investigate how to extract and exploit the
representations learned by SPNs when trained unsupervisedly as density
estimators.
Specifically, we try to answer the following questions:
\textbf{Q1}) What are the inner representations learned by SPNs?;
\textbf{Q2}) How can representations at different levels of
  abstraction be extracted from an SPN? How and why are RL approaches
  for classical deep neural models unsuitable for SPNs?;
\textbf{Q3}) Are SPN representations competitive
  with those extracted from other neural models for predictive tasks?

To do so, we make the following contributions.
First, we propose a natural interpretation of SPNs as sparse,
labeled and generative Multi Layer Perceptrons (MLPs) that are
 arranged in a graph (\textbf{Q1}).
Then, we try to better understand SPN representations 
by devising sampling routines in order to visually inspect
their generated samples.
Moreover, we visualize what each neuron has learned
by providing a probabilistic formulation of visualizing samples
maximizing the neuron activations in 
MLPs~\cite{Erhan2009,Yosinski2015} (\textbf{Q1}).
Additionally, 
since extracting representations at different levels of abstraction in a layer-wise fashion---as usually done in
MLPs~\cite{6472238}---is inadequate in SPNs, due to the aforementioned structural constraints,
we  devise
several alternative criteria to build embeddings,  like arranging nodes by type or scope
length or aggregating them by scope (\textbf{Q2}).
Finally, we demonstrate that the SPNs embeddings
 are
competitive with other classical feature extractors such as
RBMs, their deep
counterparts Deep Belief Networks (DBNs)~\cite{Salakhutdinov2009}, and deep probabilistic autoencoders~\cite{Germain2015,Kingma2013} when evaluated on several
(semi-)supervised tasks (\textbf{Q3}).

By answering the aforementioned questions we both provide
  practitioners with several routines to effectively exploit any learned SPN as
  a feature extractor, and suggest when and why to prefer one
  routine over another.
At the same time, we hope to attract those in the deep learning
community that are not familiar with such models by highlighting the differences---and
advantages---of SPNs w.r.t. classical neural models for RL.
Ultimately, we argue that SPNs are not only expressive and tractable probabilistic models 
but also  provide
rich part-based representations.
They naturally provide this without the need of being retrained
and without requiring to manually specify an architecture beforehand or
imposing an embedding size a priori, thus classifying as
promising candidates for RL.

\section{Related Work}
\label{section:rel}

The theoretical properties of SPNs have been thoroughly
investigated, while their node interactions and
practical interpretability have received little or no attention.
For instance, \cite{Delalleau2011} investigates the representational power of SPNs through a theoretical analysis that compares deep
 vs.  shallow architectures.
\citet{Martens2014} demonstrate how
expressive efficiency in SPNs correlates to their depth.
In~\cite{Rooshenas2014}, SPNs are demonstrated to be estimators equivalent  to
Arithmetic Circuits over discrete finite domains.
In~\cite{Peharz2015a}, it was shown that consistency---a less
strict constraint than decomposability---does not
lead to exponentially more compact networks.
In~\cite{Zhao2015} it is demonstrated how SPNs
are equivalent to bipartite BNs with Algebric Decision Diagrams
modeling their conditional probability tables.



Visualizations provide important tools to assess a model from a
qualitatively perspective, and have proven to be complementary to
quantitative analysis.
%
%
The most common and simplest technique for (not only deep) generative
model is to visualize sampled
instances~\cite{Larochelle2011,Germain2015}.
Recently, the need to better understand the successes of deep
models more
in depth lead to studies focused on particular architectures,
for instance Convolutional Neural Networks in~\cite{Zeiler2014} and 
 Recurrent Neural Networks, even more recently, in ~\cite{Karpathy2015}.
%
%
%
%
In this paper, we follow the work in~\cite{Erhan2009} to visualize
the feature learned by each neuron from an arbitrary layer as the input instance maximizing its
activation. 
Extensions of~\cite{Erhan2009} explored how
to impose natural image priors to visualize features from deep models
learned on image data: e.g., in~\cite{Yosinski2015}.
In~\cite{Simonyan2013}, on the other hand, the optimization problem is recast as
finding the best image maximizing a class score and computing a
saliency map for a query image sample, given a class.
 With MPE inference with SPNs we can efficiently solve an optimization
problem similar to~\cite{Erhan2009},
effectively showing that the learned features are \emph{part-based}
representations.
%


Representation Learning (RL)~\cite{6472238} works have extensively studied how to extract
useful features in unsupervised, semi-supervised and
supervised settings from both deep and shallow models.
%
RBMs have been extensively employed as robust feature extractors in
several studies, both as generative and discriminative
models~\cite{Larochelle2008,Marlin2010}.
They also inspired successful
autoregressive models like the Neural Autoregressive Distribution
Estimator (NADE)~\cite{Larochelle2011}.
%
For all these neural density estimators the structure is fixed
a priori or after a hyperparameter selection for the number of hidden
layers and hidden nodes per layer.
With SPNs, efficient structure learning is
possible.
Moreover, the
extracted representations can be assessed against the learned
structure and \emph{vice versa}, due to their recursive definition.
Masked Autoencoder Distribution
Estimators (MADEs) have been introduced in~\cite{Germain2015} as the autoencoder
variant of NADEs.
Empirically, they have been proven to be
highly competitive in terms of likelihood scores while providing tractable
 complete evidence inference.
The autoregressive property in MADEs binds inner neurons to be connected only to other neurons whose direct input
respects the order dependencies among RVs, making them sparsely connected.
Differently from SPNs, where each node outputs a
  probability, MADEs only do this in the last layer.
Similarly to MADEs, variational autoencoders (VAEs)~\cite{Kingma2013}
are generative autoencoders, but differently from MADEs they are
tailored towards compressing and learning untangled representations of
the data through a variational approach to Bayesian inference.
While VAEs have recently gained momentum as generative models, their inference
capabilities, contrary to SPNs, are limited and restricted to Monte
Carlo estimates relying on the generated samples.

W.r.t. all the above mentioned neural models, one can learn one SPN
structure from
data and obtain a highly versatile probabilistic model capable of
performing a wide variety of inference queries efficiently and at the
same time
providing very informative feature representations, as we will see in
the following sections.

\section{Sum-Product Networks}
\label{section:spn}

We denote RVs by upper-case letters, e.g., $X$, and
ordered sets of RVs by their bold variants, e.g., $\mathbf{X}$.
We denote a \emph{sample} for $\mathbf{X}$ as $\mathbf{x} \sim \mathbf{X}$,
and a single value from it as $x_{j}$.
We define a set of $m$ samples---a \emph{dataset}---as 
$\{\mathbf{x}^{i}\}_{i=1}^{m}$.
Let $\mathbf{Q}\subseteq \mathbf{X}$, then  $\mathbf{x}_{|\mathbf{Q}}$
denotes the \emph{marginal} sample $\mathbf{q}\sim \mathbf{Q}$, i.e., the restriction of $\mathbf x$ to $\mathbf{Q}$.


%
%
%

  A \emph{Sum-Product Network} (SPN) $S$ over RVs $\mathbf X$
  is a probabilistic model defined via a
  rooted directed acyclic graph (DAG).
  %
  Let $\nodeset{S}$ be the set of all nodes in $S$ and
  $\mathsf{ch}(n)$ denote the set of children of a node $n
  \in \nodeset{S}$.
  The DAG structure recursively defines a distribution $S_n$ for each
  node $n \in \nodeset{S}$.
  To a \emph{leaf} node $n$, i.e., $\mathsf{ch}(n) = \emptyset$, 
is associated 
a computationally tractable distribution $\phi_{n} \triangleq S_n$ over $\mathsf{sc}(n)\subseteq\mathbf{X}$, 
where $\mathsf{sc}(n)$ denotes the \emph{scope} of $n$.~\footnote{For discrete (resp. continuous) RVs, $\phi_{n}$ represents a probability mass function (resp. density function). 
We will generically refer to both as \emph{probability distribution functions} (pdfs).}
%

An inner node  $n$ is either a \emph{sum} or \emph{product} node and
its scope is recursively defined as 
$\mathsf{sc}(n) = \bigcup_{c\in\mathsf{ch}(n)}\mathsf{sc}(c)$.
A sum  node $n$ outputs a non-negative weighted sum over its children: 
$S_n = \sum_{c \in \mathsf{ch}(c)} w_{nc} \, S_c$.
A product node $n$ outputs a product over its children:
$S_n = \prod_{c \in \mathsf{ch}(c)} S_c$.
The \emph{distribution} encoded by an SPN $S$ is the normalized output of its root, and it depends both on the structure of $S$ and its
parameters---the set of sum-weights and the leaf
distributions parameters---denoted as $\mathbf{w}$.  
Let $\nodeset{S}^{\oplus}$ (resp. $\nodeset{S}^{\otimes}$) be the set
 of all sum (resp. product) nodes in $S$. 
%
An example of SPNs is shown in Figure~\ref{fig:spn-layered}, where
 the direction of the model edges are graphically omitted to avoid clutter.

%
%
%

In order to allow for efficient inference, an SPN $S$ is required to be \emph{complete}, i.e., 
$\forall n\in  \nodeset{S}^{\oplus}$, $\forall c_{1}, c_{2}\in \mathsf{ch}(n):  
\mathsf{sc}(c_{1})=\mathsf{sc}(c_{2})$, 
and \emph{decomposable}, i.e.,  $\forall n\in  \nodeset{S}^{\otimes}, \forall c_{1}, c_{2}\in \mathsf{ch}(n), c_{1}\neq c_{2}:
 \mathsf{sc}(c_{1})\cap\mathsf{sc}(c_{2})=\emptyset$ \cite{Poon2011,Peharz2015a}.
Moreover, we assume SPNs to be \emph{locally normalized} \cite{Peharz2015a}: $\forall n\in \nodeset{S}^{\oplus}, \sum_{c\in\mathsf{ch}(n)}w_{nc}=1$.
W.l.o.g., we also assume SPNs to have alternate node types, which we call \emph{alternate} SPNs, i.e., each
product (resp. sum) node can have as child a sum (resp. product) node~\cite{Vergari2015}.

Computing the \emph{exact} probability of complete evidence $\mathbf{x} \sim \mathbf{X}$ consists of a single bottom-up
evaluation of $S$:
each leaf $n$ evaluates $\phi_{n}(\mathbf{x}_{|\mathsf{sc}(n)})$ and 
subsequently, each inner node computes the probability
$S_{n}(\mathbf{x}_{|\mathsf{sc}(n)})$---or short-hand
$S_{n}(\mathbf{x})$---before passing it to its parent, till the root.
This computation is guaranteed to be tractable
as long as the network size---$|S|$, the number of edges in it---is
polynomial in $|\mathbf{X}|$.

Even \emph{exact marginal inference} can be computed in linear time
w.r.t. $|S|$ in a
complete and decomposable SPN $S$~\cite{Poon2011,Peharz2015a}: 
%
to compute the query $p(\mathbf{Q=q}),
\mathbf{Q}\subset\mathbf{X}$, one evaluates $\phi_{n}$ for each leaf
$n$ by marginalizing RVs in $\mathsf{sc}(n)$ not in $\mathbf{Q}$,
then propagating the outputs bottom-up, as before.
%
%
Consequently, also exact conditionals are computable
in linear time, since $p(\mathbf{Q}|\mathbf{E}) = p(\mathbf{Q},
\mathbf{E})/p(\mathbf{E})$, for $\mathbf{Q},
\mathbf{E}\subset\mathbf{X}$.

%


Exact MPE inference is hard in general SPNs~\cite{Peharz2016,DBLP:conf/uai/ConatyCM17}.
However, reasonable approximations for MPE solutions can be found in linear
time in general SPNs~\cite{Poon2011,Peharz2016}.
%
%
Given an SPN $S$ over RVs $\mathbf{X}$, to find an MPE assignment
$\mathbf{q}^{*}=\argmax_{\mathbf{q}\sim \mathbf{Q}}p(\mathbf{E},\mathbf{q})$
for some RVs $\mathbf{E}, \mathbf{Q} \subset\mathbf{X},
\mathbf{E}\cap\mathbf{Q}=\emptyset, \mathbf{E}\cup \mathbf{Q}=\mathbf{X}$,
$S$ is transformed into a Max-Product Network (MPN) $M$,
by replacing each sum node $n$ with a \emph{max node}
computing $\max_{c\in\mathsf{ch}(n)}w_{nc}M_{c}(\mathbf{x})$
 and each leaf distribution $\phi_n$ with a maximizing distribution $\phi_{n}^{M}$ \cite{Peharz2016}.
In a first bottom-up step,
one computes $M(\mathbf{x}_{|\mathbf{E}})$.
A top-down step traces back the MPE
solution for RVs $\mathbf{Q}$.
Starting from the root and following only the max output
child  of a max node and all the children 
of a product node, an 
\emph{induced tree} is grown.
Taking the $\argmax$ over its leaves retrieves the MPE solution~\cite{Poon2011}.


%
%
%

The structure of SPNs can be effectively learned from data
by leveraging the \emph{probabilistic semantics} of sum nodes as
\emph{mixture models} over their child distributions
and product nodes being \emph{factorizations of independent components}~\cite{Poon2011,Peharz2015a}.
In particular, a categorical
latent RV $H_{n}$, having values in
$\{1,\dots,|\mathsf{ch}(n)|\}$, can be associated to each sum node
$n$.
Network weights $w_{nk}$ can be
interpreted as the probabilities 
of choosing the $k$-th child branch from
sum node $n$, having taken the path from the root up to
$n$.
Several constraint-based algorithms exploit this perspective and perform variants of hierarchical
co-clustering~\cite{Gens2013,Rooshenas2014,Dennis2015,Vergari2015}.
To introduce a decomposable product node, RVs
are clustered by some statistical independence test, while complete sum
nodes are introduced by clustering samples
in sub-populations.
%
%
%
The first learner  adopting such a schema is
\textsf{LearnSPN}~\cite{Gens2013}, which greedily induces 
\emph{tree-shaped} SPNs by recursively splitting the data matrix top
down along its axis.
For each call on a submatrix, column splits add child nodes to product
nodes, while those on rows extend sum node. 
RVs are checked for independency by means of a $G$-test and a product node is inserted in the network if the test is passed with threshold $\rho$
.
A sum node
$n$ is inserted over $k$ child nodes if a clustering step over the
rows produced $k$ different clusters.
The weights $w_{nc}$ are
directly estimated as the proportions of samples falling into each
cluster $c$.
In this
way, no weight learning step is needed after the network is fully
grown.
The learning
process stops when the number of samples in a partition falls under a threshold $\mu$.
Then leaves are introduced as univariate distributions whose parameters are smoothed with a coefficient $\alpha$.
As they are considered to be independent one from another, a product
node is put on top of them.

Here we adopt such a structure learning approach because i) it is simple
and yet effective~\cite{Gens2013,Vergari2015};
ii) it does not require designing or fixing a priori
a network structure; iii) it allows us to automatically determine the size of the
representations we extract from SPNs in Section~\ref{section:rle};
iv) finally, by performing hierarchical co-clustering, \textsf{LearnSPN}
acts as a \emph{recursive data crawler},
providing the rich part-based representations we visualize in Section~\ref{section:vis}.

%
%
%

While latent RVs associated to sum nodes suggest a natural way to
exploit SPNs as \emph{generative} models,
to the best of our knowledge, they have not been employed in the
literature to sample.
We use a
simple sampling scheme for SPNs, effectively adopting it to visually inspect what a network
has learned in Section~\ref{section:rle}.
To generate one sample $\mathbf{x}$ from the pdf over $\mathbf{X}$ encoded by an SPN
$S$,
one traverses $S$ top-down and induces a tree similarly to MPE
inference:
at each sum node $n$,
 the child $c$ to follow is randomly chosen with
probability proportional to $w_{nc}$.
Product node children are followed all at once.
To draw a sample $\mathbf{q}$ from the conditional distribution
$p(\mathbf{Q}|\mathbf{e})$, one chooses the sum child branch $c$
proportionally to $w_{nc}S_{c}(\mathbf{e})$, instead.
%
Again, the leaves of the induced tree form a partition over all $\mathbf{X}$.
A complete sample is generated by sampling from these leaf distributions.

\section{Interpreting Sum-Product Networks as neural networks}
\label{section:int}


\begin{figure}[!t]
  \centering
  \includegraphics[width=0.8\columnwidth]{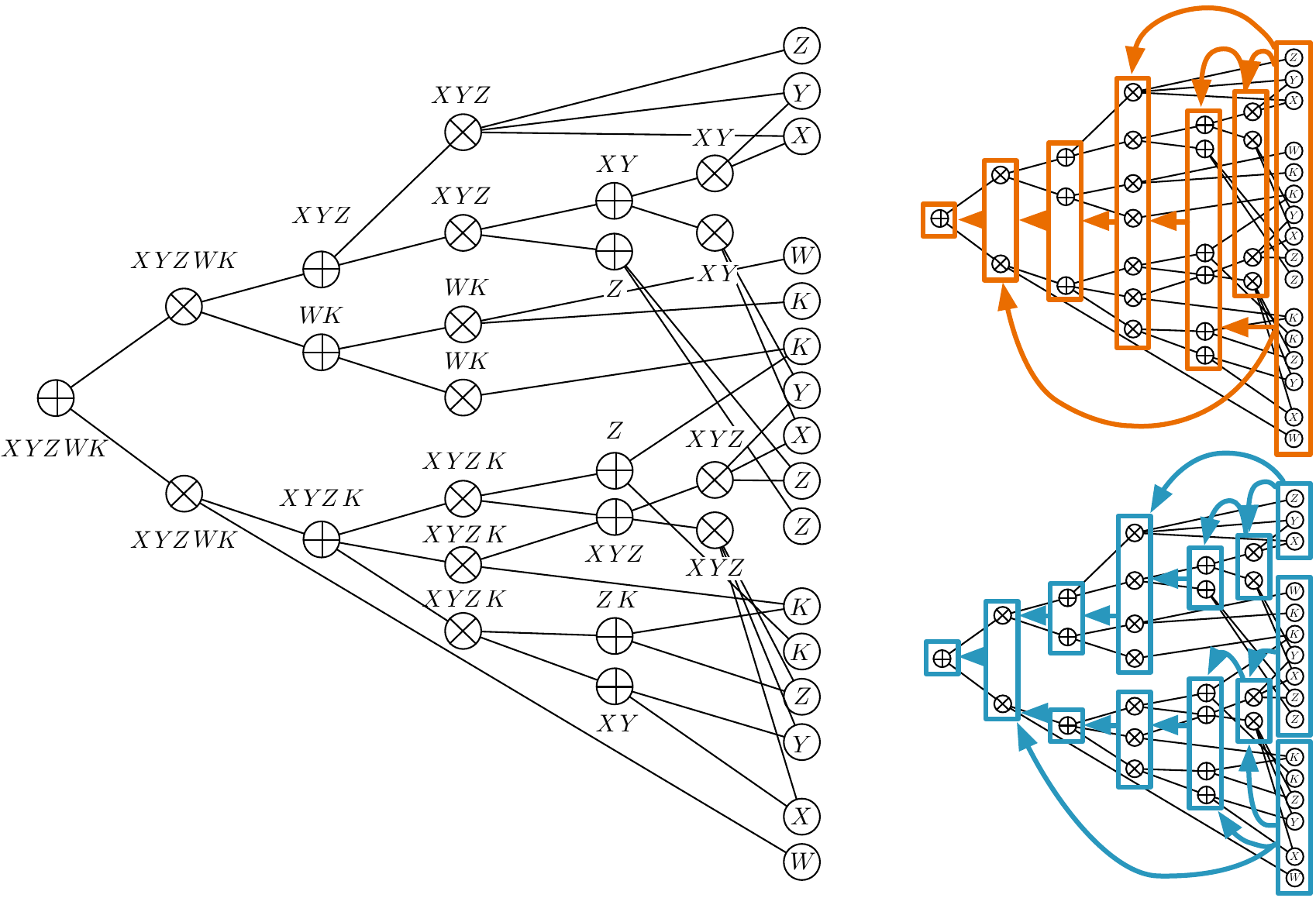}
  \caption{\emph{Layered representation of SPNs.}
    On the left, an example of a complete and decomposable SPN over
    RVs $X, Y, Z, W$.
    Leaves are represented as labelled circles and inner nodes
    have their scope associated.
  On the right, two possible layered representations featuring more
  (bottom) or less (top) sparse weight connections.}
  \label{fig:spn-layered}
\end{figure}

Similarly to Arithmetic  Circuits (ACs)~\cite{Darwiche2003,Rooshenas2014}, 
SPNs are \emph{computational graphs} in which the  operations computed
to evaluate a pdf are rearranged into an efficient data structure.
%
Differently from ACs, the latent RVs semantics in SPNs  allows for \emph{direct} structure
learning~\cite{DBLP:conf/icml/ChoiD17}.
SPNs and ACs are \emph{not} classical PGMs.
%
%
Edges in SPNs deterministically determine the evaluation of the nodes
in the DAG, 
which represent \emph{computational units}, while
in PGMs nodes represents RVs and edges the statistical dependencies
among them.
%


As computational graphs, SPNs can be seen as feedforward deep
neural networks (DNNs) in which hidden neurons can only compute sum and
products and input neurons are pdfs.
%
We argue that the peculiarity of SPNs as DNNs lies in them being a)
\emph{labelled}, b) \emph{constrained} and c) retaining a \emph{fully probabilistic} semantics.
The scope function labels each network node by a set of RVs,
enabling a \emph{direct encoding} of the input space~\cite{6472238}.
The DAG \emph{constrained topology}, due to completeness and
decomposability of scopes, determines \emph{sparse} and \emph{local} connections,
similar to convolutional networks.
%
%
Moreover, like RBMs, but differently
from deep estimators like
NADEs and MADEs~\cite{Germain2015},
each neuron activation, i.e., $S_{n}(\mathbf{x})$, is still a valid probability
value by definition.
These properties suggest each hidden neuron
to act as \emph{probabilistic part-based feature extractor}, which we
investigate in Section~\ref{section:vis}.

We propose an interpretation of SPNs as
sparse Multi Layer Perceptrons (MLPs) 
whose layers are
arranged in a DAG. 
A classic \emph{sequential} MLP consists of an input layer, a series of hidden layers and
an output layer. 
A hidden layer of $s$ neurons is a function of its input $\mathbf{x}\in\mathbb{R}^{r}$:
 $h(\mathbf{x}) =\sigma(\mathbf{W}\mathbf{x}+ \mathbf{b})$, 
 with $\sigma$ being a nonlinear activation, e.g., ReLU~\cite{6472238}, and
 $\mathbf{W}\in\mathbb{R}^{s\times r}$ a linear transformation with
 bias $\mathbf{b}\in\mathbb{R}^{s}$.
 %
 
%

To reframe an SPN as an MLP one first has to \emph{group nodes into layers}
containing nodes of the same type.
Each layer can receive input connections from
multiple layers (including the input layer),
and whose adjacent input and
output layers are made up of nodes of a different type.
Moreover, one layer can feed multiple layers with its output.
%
These layers lend themselves to
be arranged in a DAG based on their multiple input and output
connections.

The input layer still computes the pdfs of the leaf distributions.
The output of each hidden layer, based on its type, can be computed as
follows.
Let $\mathbf{S}(\mathbf{x})\in\mathbb{R}^{s}$ denote the output of a
generic SPN hidden layer with $s$ nodes: $\mathbf{S}(\mathbf{x})=\langle
S_{1}(\mathbf{x}),\dots , S_{s}(\mathbf{x})\rangle$.
A sum layer receiving $r$ input nodes would output $\mathbf{S}(\mathbf{x}) = \log(\mathbf{W}\mathbf{x})$ where
$\mathbf{W}\in\mathbb{R}_{+}^{s\times r}$ is the weight matrix defining the sparse connections: $\mathbf{W}_{(ij)}=w_{ij}$ if there is an edge between nodes $i$ and $j$, and $0$ otherwise.
For locally normalized SPNs, we want 
$\mathbf W \cdot \mathbf 1_r = \mathbf 1_s$.
A product layer, instead, would compute
$\mathbf{S}(\mathbf{x}) = \exp(\mathbf{P}\mathbf{x})$, with $\mathbf{P}\in\{0,
1\}^{s\times r}$ being a sparse connection matrix: $\mathbf{P}_{(ij)}=1$ if there is an edge between nodes $i$ and $j$, 0 otherwise.
%
In this reparameterization $exp$ and $log$ functions act as
non-linear functions $\sigma$ and the signals between layers switch from
the domains of probabilities to log-probabilities and vice versa.
%
%
The absence of a bias term $\mathbf{b}$ is due to dealing with
normalized probabilities.

%
Grouping all nodes at a certain depth in a single layer 
 leads to
 sequential DAGs with \emph{very sparse weight matrices}.
On the other hand, grouping only sibling nodes in a layer increases
the number of layers in the DAG arrangement.
%
%
In general, grouping nodes into layers in the
DAG is somehow \emph{arbitrary}: 
one can always break them up or merge them together to reduce or
enhance sparsity on the  matrices $\mathbf{W}$ and $\mathbf{P}$.
In Figure~\ref{fig:spn-layered} the same tree-shaped SPN is rearranged into a more
sequential architecture.
The advantages such a reparameterization offers are:
i) better understanding SPNs as DNNs, highlighting the role of
nonlinearities in SPNs; 
ii) allowing for efficient GPU implementations;~\footnote{As done in our code, 
  available at \url{https://github.com/arranger1044/spyn-repr}.} 
iii) paving the way to structure learning as constrained optimization---learning the sparse $\mathbf{P}$ and
$\mathbf{W}$ indeed  determines the DAG of $S$;  and
iv) questioning what are the representations learned from an SPN and
how to extract them from it like for classic MLPs.

\section{Extracting Representations from SPNs}
\label{sec:spn-repr}

A new feature representation for a set of samples---usually called \emph{embedding} when it is
continuous and dense---is a transformation of such a set to a new
geometric space.
%
The main aim of Representation Learning (RL) approaches is to extract
\emph{meaningful} feature representations, such that they can better
explain the latent factors underlying the data or be effectively
\emph{reused} in other predictive tasks~\cite{6472238}.
%
%
In this section, we discuss how to employ SPNs for RL, following our interpretation of SPNs as
peculiar DNNs, and how classical depth-based feature extraction
criteria are unsatisfactory for SPNs.
Furthermore, looking at SPNs under a RL lens can help
better understanding them as probabilistic models, as well.
For deep architectures, it is common practice to employ the top hidden layer
activations as the learned
representations~\cite{6472238,Yosinski2015}.
%
The rationale behind this \emph{layer-wise} extraction criterion is that such representations are
arranged in a \emph{hierarchy of abstractions} at different levels of
granularity, correlated with the depth of a layer, with
the top layers providing the most complex and meaningful features~\cite{Erhan2009,Zeiler2014,Yosinski2015}. 
%
%
We are looking for an analogous and reasonable criterion to filter node
activations in SPNs.
%
%
Unfortunately, employing the MLP reparameterization introduced in Section~\ref{section:int} 
does not guarantee that the layer-wise or depth-wise criteria would
produce representations at different levels of abstraction in SPNs.
We actually deem them inadequate, due to the peculiar constrained structure in SPNs.
%

\begin{figure}[!t]
  \centering
    \includegraphics[width=0.3\columnwidth]
    {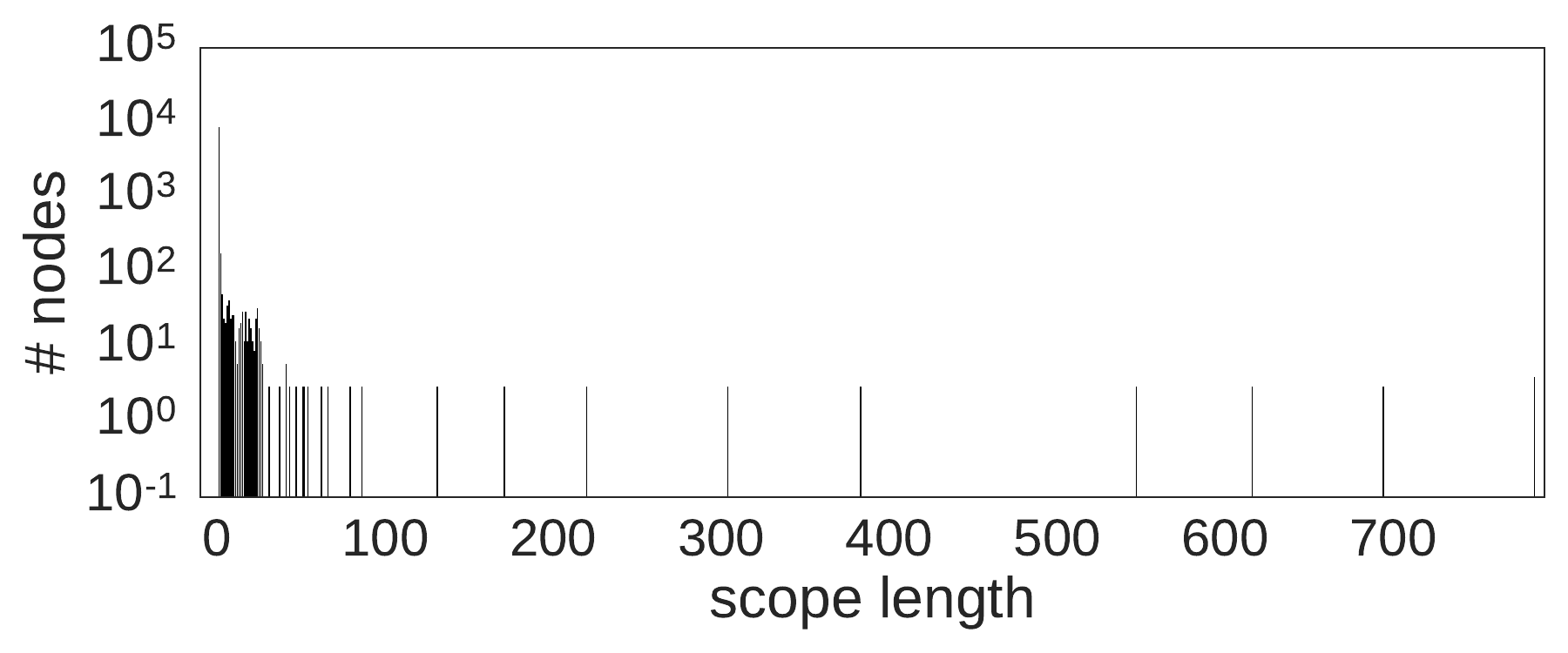}\hspace{5pt}
    \includegraphics[width=0.3\columnwidth]
    {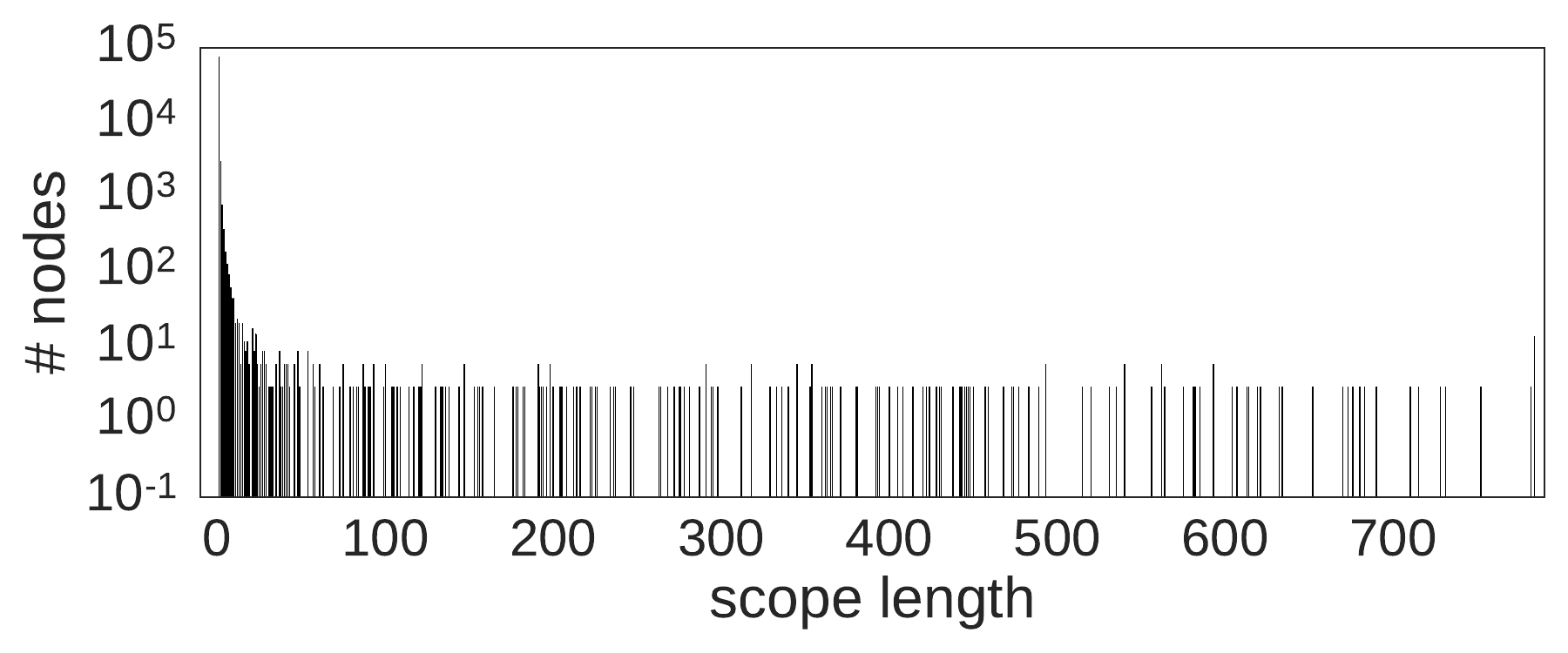}\hspace{5pt}
    \includegraphics[width=0.3\columnwidth]
    {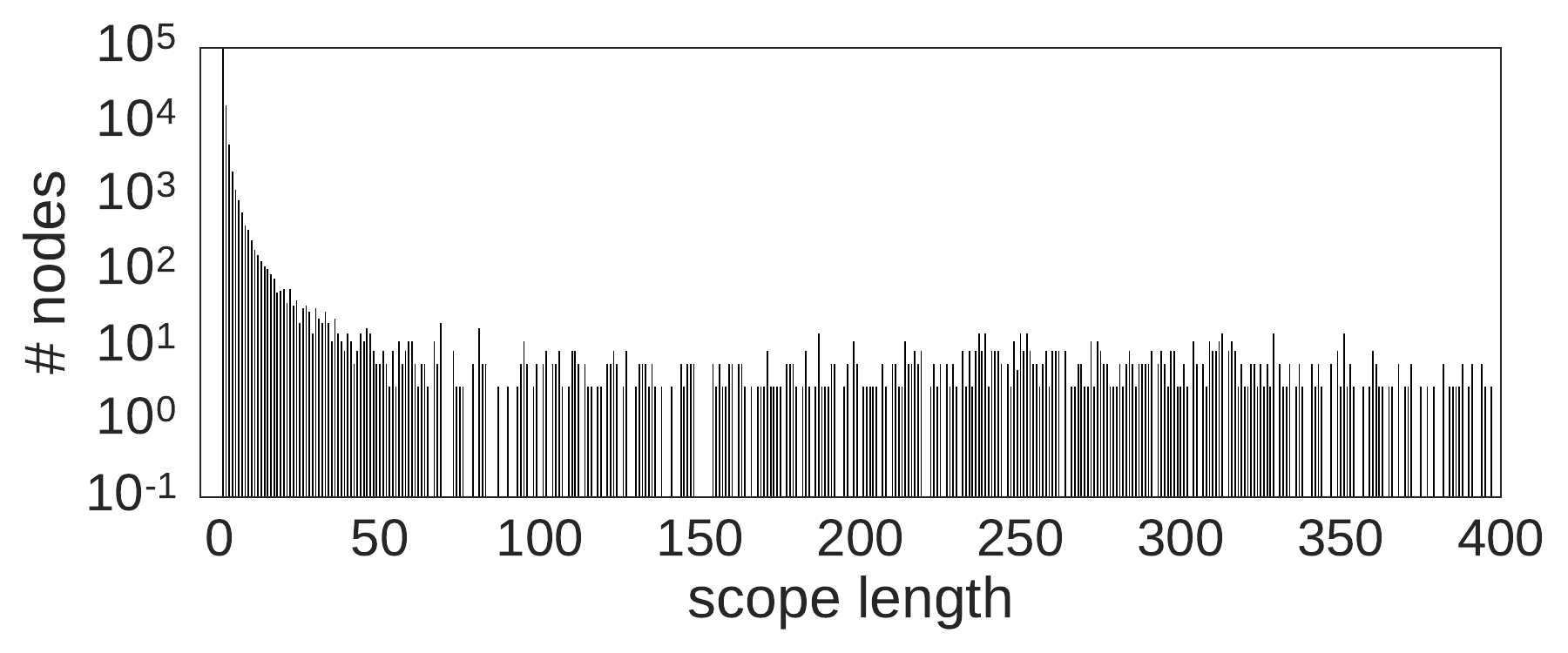}
  \\[1pt]
  \subfloat[\textsf{REC}]{\includegraphics[width=0.3\columnwidth]
    {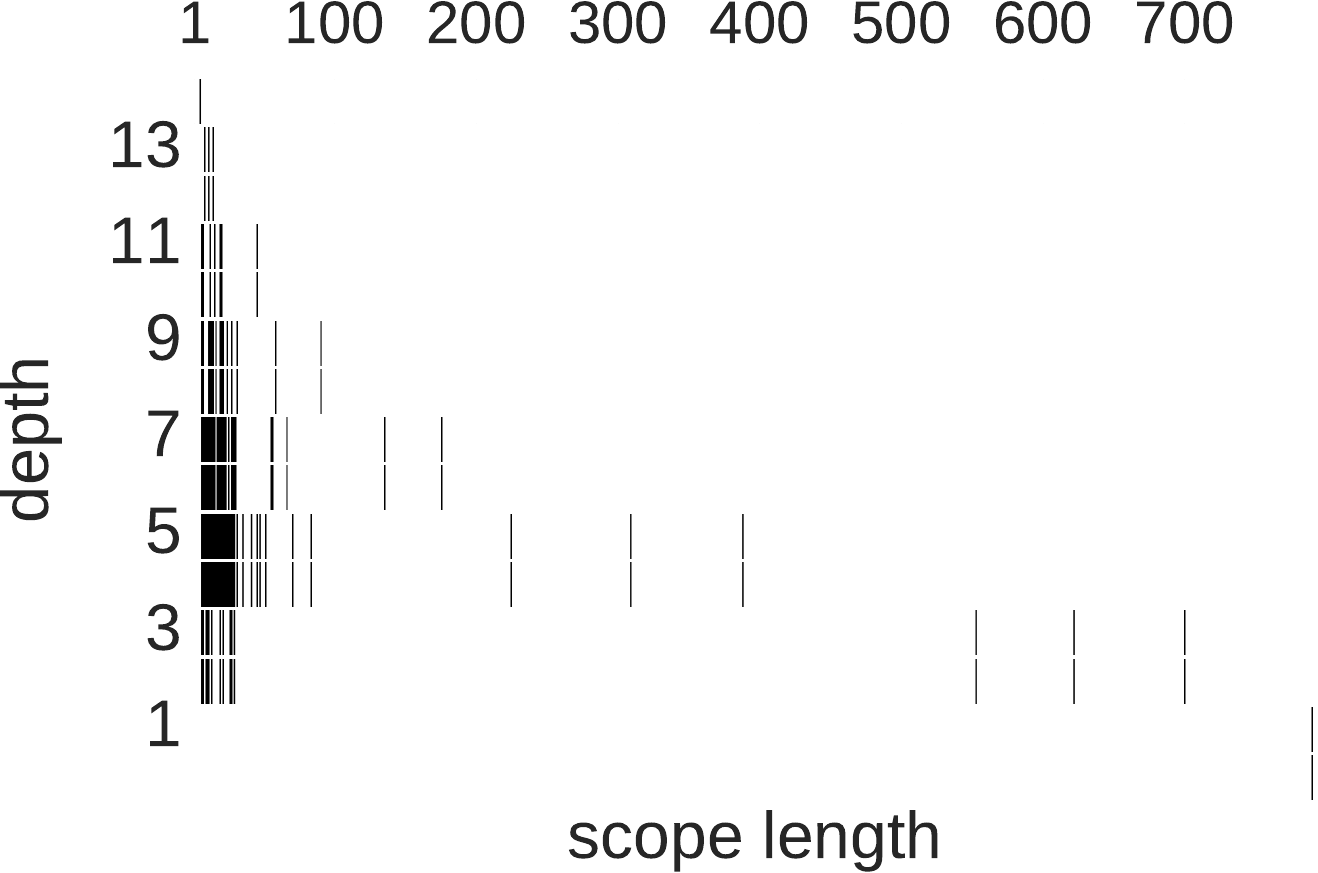}
    \label{fig:layer-scope-rec}}\hspace{5pt}
  \subfloat[\textsf{CON}]{\includegraphics[width=0.3\columnwidth]
    {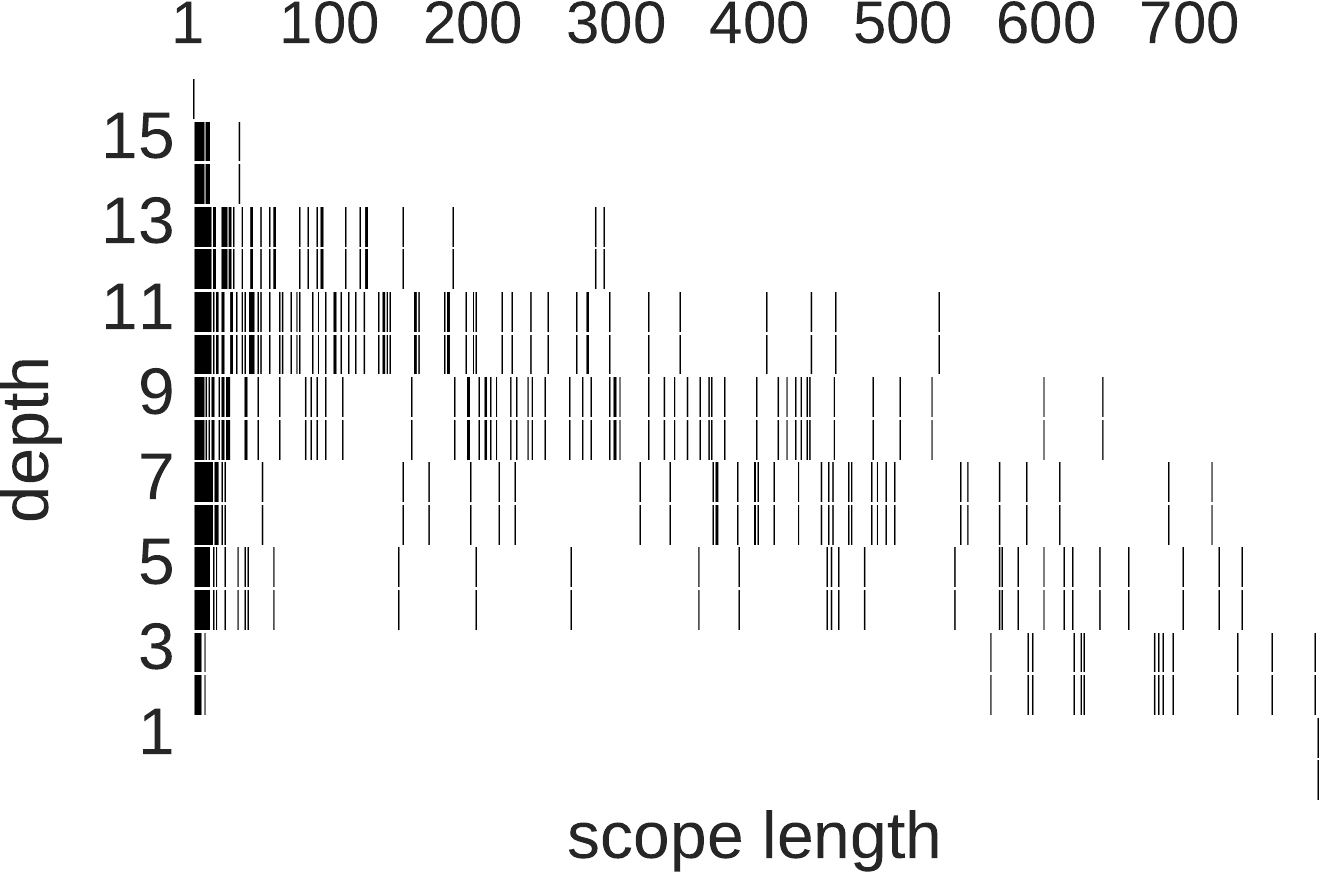}
    \label{fig:layer-scope-con}}\hspace{5pt}
  \subfloat[\textsf{BMN}]{\includegraphics[width=0.3\columnwidth]
    {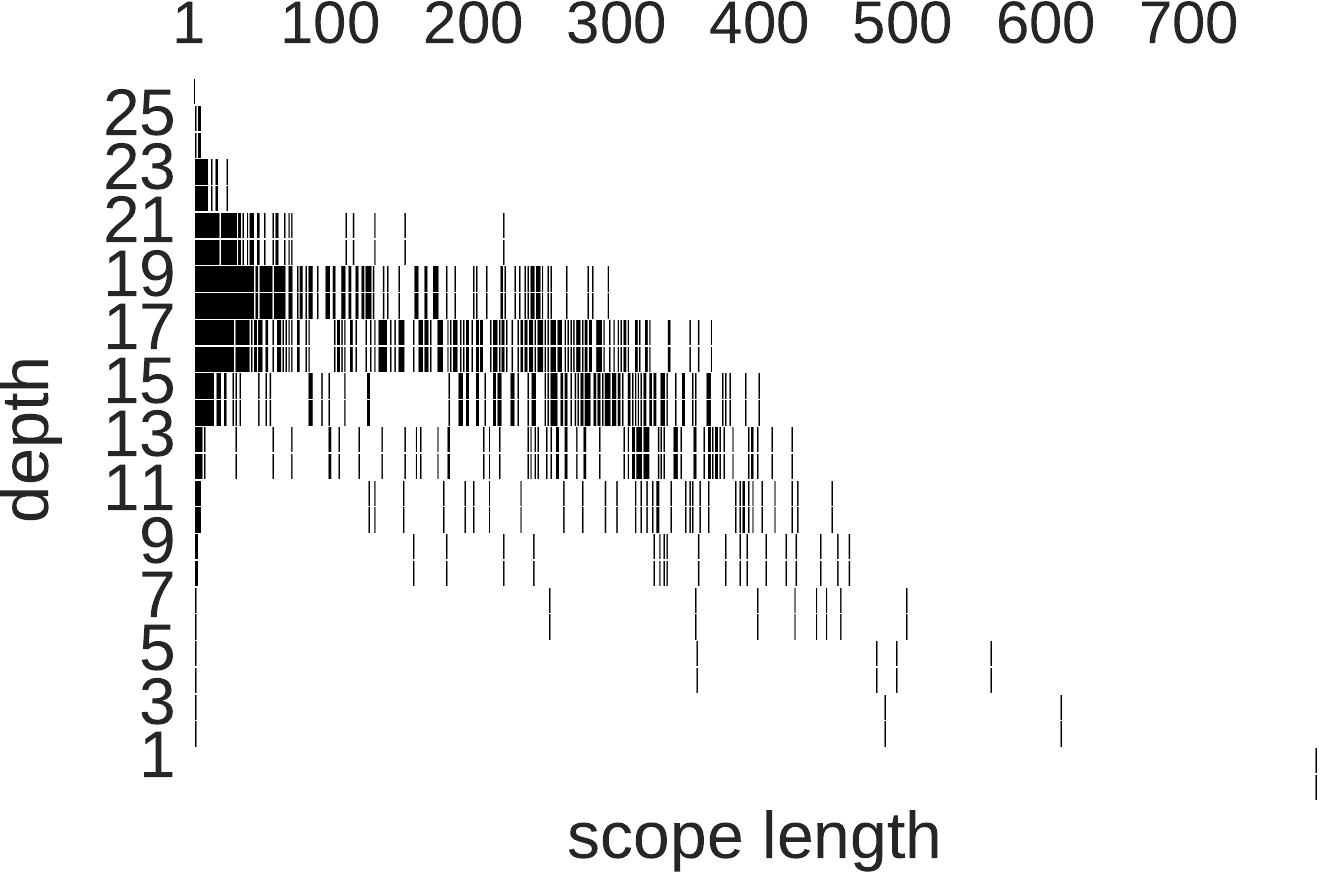}
    \label{fig:layer-scope-bmn}}
  \caption{\emph{Scope length distributions}. Scope length distributions for \textsf{SPN-III} models on
    \textsf{REC} (Figs.~\ref{fig:layer-scope-rec}),
    \textsf{CON} (Figs.~\ref{fig:layer-scope-con}),
    \textsf{BMN} (Figs.~\ref{fig:layer-scope-bmn})
    (cf. Section~\ref{section:rle} for model and dataset details).
    as a histogram of their possible values (from 1 to $|\mathbf{X}|$) against
the number of nodes having that scope length and belonging to a
certain depth.
    A long tail distribution for number of nodes w.r.t. scope lengths
    is visible (top).
    Very different scope lengths are grouped at the same layer depth (bottom, a bar
    indicates there is \emph{at least}
    one node of the corresponding scope length at that depth).}
  \label{fig:hist-scopes}
\end{figure}

%
As a first motivation, consider that the top layers in an SPN would
comprise significant fewer nodes w.r.t. lower layers.
Second, the
choice of any other layer in the DAG would be somehow
\emph{arbitrary}.  
Even the depth of a layer seems an unsatisfactory criterion, since 
 nodes with very different scopes, hence encoding parts of the input
 space at very different granularities, may still
 share the same depth.
To confirm these claims,
 we visualize the network topology of the SPN models employed in our
experiments (see Section~\ref{section:rle})
w.r.t. the scope information associated to their nodes.
Let $S$ be an SPN over RVs $\mathbf{X}$. 
We define the \emph{scope length} of a node $n\in\nodeset{S}$  as
$|\mathsf{sc}(n)|$.
The scope length of $S$ is $|\mathbf{X}|$.
We plot the scope lengths in Figure~\ref{fig:hist-scopes}.
A \emph{long tail} effect is
visible:
80\% or more of the nodes in each model have a scope length of 1 to 3.
Additionally, top layers indeed comprise very few nodes---as expected
on tree-shaped SPNs as learned by LearnSPN-like algorithms.
Furthermore, nodes at the same depth level can show a high
variance of scope lengths.
These visualizations support the inadequacy of extracting representations from SPNs
by collecting activations by depth.

\begin{figure}[!t]
  \centering
  \subfloat[]{
    \includegraphics[width=0.2\columnwidth]
    {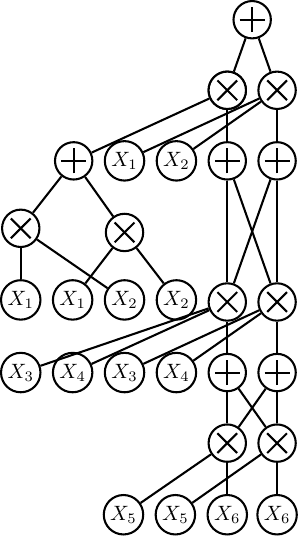} \label{fig:spn-emb-I}}
  \subfloat[]{
    \includegraphics[width=0.21\columnwidth]
    {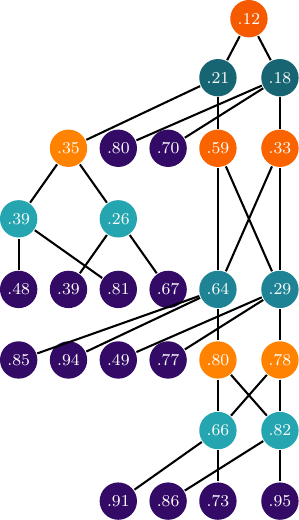} \label{fig:spn-emb-I-eval}}
  \subfloat[]{
    \includegraphics[width=0.25\columnwidth]
    {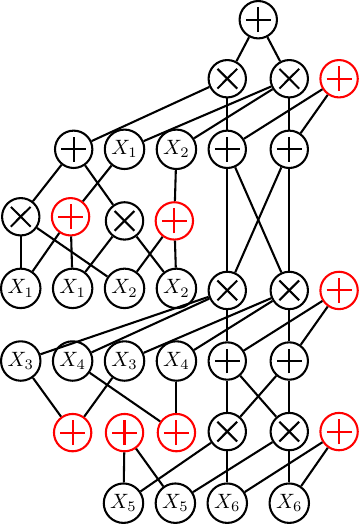} \label{fig:spn-emb-II}}
  \subfloat[]{
    \includegraphics[width=0.25\columnwidth]
    {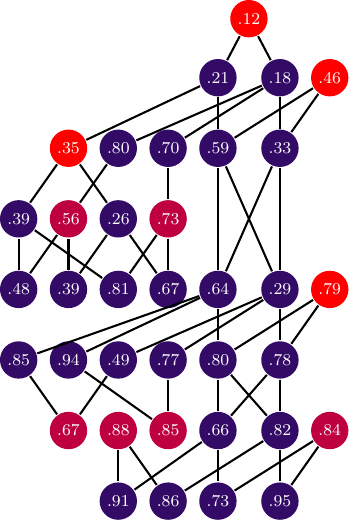} \label{fig:spn-emb-II-eval}}\\[10pt]
  \begin{minipage}[t]{1.0\linewidth}
    \flushleft
    \begin{minipage}[c]{0.05\linewidth}
      \vspace{-2pt}\subfloat[]{\label{fig:spn-emb-full}}
    \end{minipage}
\tiny
      $\mathbf{e}^{i}=
      \langle\highlight[gold6]{.12}\highlight[petroil6]{.21}\highlight[petroil6]{.18}\highlight[gold2]{.35}\highlight[gold4]{.59}\highlight[gold4]{.33}\highlight[petroil2]{.39}\highlight[petroil2]{.26}\highlight[petroil4]{.64}\highlight[petroil4]{.29}\highlight[gold2]{.80}\highlight[gold2]{.78}\highlight[petroil2]{.66}\highlight[petroil2]{.82}\rangle$
      \end{minipage}\\[5pt]
  \raggedright\begin{minipage}[t]{0.42\linewidth}
    \flushleft
    \begin{minipage}[c]{0.05\linewidth}
      \vspace{-2pt}\subfloat[]{\label{fig:spn-emb-sum}}
    \end{minipage}\hspace{12pt}{\tiny
      $\mathbf{e}^{i}_{\mathsf{sum}}=
      \langle\highlight[gold6]{.12}\highlight[gold2]{.35}\highlight[gold4]{.59}\highlight[gold4]{.33}\highlight[gold2]{.80}\highlight[gold2]{.78}\rangle$
      }\end{minipage}\begin{minipage}[t]{0.5\linewidth}
    \flushleft
    \begin{minipage}[c]{0.05\linewidth}
      \vspace{-2pt}\subfloat[]{\label{fig:spn-emb-prod}}
    \end{minipage}\hspace{12pt}{\tiny
      $\mathbf{e}^{i}_{\mathsf{prod}}=
      \langle\highlight[petroil6]{.21}\highlight[petroil6]{.18}\highlight[petroil2]{.39}\highlight[petroil2]{.26}\highlight[petroil4]{.64}\highlight[petroil4]{.29}\highlight[petroil2]{.66}\highlight[petroil2]{.82}\rangle$
      }\end{minipage}\\[5pt]
  \raggedright\begin{minipage}[t]{0.45\linewidth}
    \flushleft
    \begin{minipage}[c]{0.05\linewidth}
      \vspace{-2pt}\subfloat[]{\label{fig:spn-emb-scope-s}}
    \end{minipage}\hspace{12pt}{\tiny
      $\mathbf{e}^{i}_{\mathsf{S}}=
      \langle\highlight[gold2]{.35}\highlight[petroil2]{.39}\highlight[petroil2]{.26}\highlight[gold2]{.80}\highlight[gold2]{.78}\highlight[petroil2]{.66}\highlight[petroil2]{.82}\rangle$
      }\end{minipage}\begin{minipage}[t]{0.3\linewidth}
    \flushleft
    \begin{minipage}[c]{0.05\linewidth}
      \vspace{-2pt}\subfloat[]{\label{fig:spn-emb-scope-m}}
    \end{minipage}\hspace{12pt}{\tiny
      $\mathbf{e}^{i}_{\mathsf{M}}=
      \langle\highlight[gold4]{.59}\highlight[gold4]{.33}\highlight[petroil4]{.64}\highlight[petroil4]{.29}\rangle$
      }\end{minipage}\begin{minipage}[t]{0.25\linewidth}
    \flushleft
    \begin{minipage}[c]{0.05\linewidth}
      \vspace{-2pt}\subfloat[]{\label{fig:spn-emb-scope-l}}
    \end{minipage}\hspace{12pt}{\tiny
      $\mathbf{e}^{i}_{\mathsf{L}}=
      \langle\highlight[gold6]{.12}\highlight[petroil6]{.21}\highlight[petroil6]{.18}\rangle$
      }\end{minipage}\\[5pt]
  \raggedright
  \begin{minipage}[t]{0.61\linewidth}
    \flushleft
    \begin{minipage}[c]{0.05\linewidth}
      \vspace{-2pt}\subfloat[]{\label{fig:spn-emb-aggr}}
    \end{minipage}\hspace{12pt}{\tiny
      $\mathbf{e}^{i}_{\mathsf{aggr}}=\langle\highlight[red]{.12}\highlight[red]{.46}\highlight[red]{.35}\highlight[purple]{.56}\highlight[purple]{.73}\highlight[red]{.79}\highlight[purple]{.67}\highlight[purple]{.88}\highlight[purple]{.85}\highlight[purple]{.84}\rangle$}
  \end{minipage}
  \begin{minipage}[t]{0.36\linewidth}
    \flushleft
    \begin{minipage}[c]{0.05\linewidth}
      \vspace{-2pt}\subfloat[]{\label{fig:spn-emb-aggr-l}}
    \end{minipage}\hspace{12pt}{\tiny
      $\mathbf{e}^{i}_{\mathsf{aggr-l}}=\langle\highlight[red]{.12}\highlight[red]{.46}\highlight[red]{.35}\highlight[red]{.79}\rangle$
    }
  \end{minipage}
    \caption{\emph{Extracting embeddings with SPNs}.
      By feeding the SPN $S$ in (a) a sample
    $\mathbf{x}^{i}$, one evaluates it and collects its node activations (b).
    We devise several filtering criteria to build embeddings: 
    collecting all inner node activations (e); filtering by node type,
    obtaining sum (f) and
    product only embeddings (g);
    filtering nodes by \textsf{S}mall ($|\mathsf{sc}(n)|=2$) (h),
    \textsf{M}edium ($|\mathsf{sc}(n)|=4$) (i) and \textsf{L}arge
    ($|\mathsf{sc}(n)|=6$) (j) scope
    lengths;
    by aggregating all nodes by similar scopes (k) as represented by the red sum
    nodes
    introduced in (c) and evaluated in (d),
    or only inner nodes (l).}
  \label{fig:spn-emb}
\end{figure}

Therefore, we start investigate 
alternative criteria to extract embeddings from an SPN.
The simplest would be by collecting
\emph{all} inner nodes outputs---the longest
embedding for a given SPN (excluding the overabundant leaves).
Nevertheless, even this heuristic is somehow still unsatisfactory: i) it
treats all neurons equally, despite their different roles in the
network, and ii) embeddings of such a size can easily suffer from the
curse of dimensionality when employed as features in predictive tasks.
Therefore, we propose several additional criteria \emph{to filter nodes} from this full
embedding, to better understand the influence of the network topology
over the extracted representations and also to investigate an effective way
to reduce the size of an embedding.

We first devise filtering activations by \emph{node type} (i), to assess the role of
sum versus product nodes as feature extractors.
Then, we argue that a hierarchy of representations at
different levels of abstractions for SPNs can be captured by
how the scope information decomposes along the network structure.
We therefore correlate the complexity
of a representation learned by a node to its scope length,
creating embeddings by filtering nodes having
\emph{comparable scope lengths} (ii).
We further investigate how the scope information correlates to
the level of abstraction of the representations by aggregating activations
from nodes sharing the \emph{same scope information} (iii) by
leveraging the recursive definition of SPNs.
Figure~\ref{fig:spn-emb} depicts all the different embeddings we
propose to extract from one SPN.

Before proceeding with an empirical evaluation of \emph{how} meaningful the proposed
embedding extraction criteria are, we try to gain a deeper
understanding of \emph{what} the representations learned by SPNs are,
by leveraging
visualization techniques.

\section{Visualizing SPN representations}
\label{section:vis}

To investigate the hypothesis  of SPNs learning a hierarchy of
part-based representations, we visualize the \emph{representations
learned by single neurons} in the network.
We do this by exploiting both the direct encoding to the input space that the scope
function provides, and the ability of SPNs to perform MPE inference
efficiently (even though approximately).

For DNNs, one generally assumes the feature learned by the $i$-th neuron
 in the $j$-th layer
to be approximated by the representation in the
input space $\mathbf{x}^{*}\sim\mathbf{X}$ \emph{maximizing its activation} $h_{i}^{j}$~\cite{Erhan2009,Yosinski2015}.
To obtain such a representation, one can compute the bounded norm  solution of the following non-convex problem:
\begin{equation}
  \label{eq:vis-sgd}
  \mathbf{x}^{*} = \argmax_{\mathbf{x},
    ||\mathbf{x}||=\gamma}h_{ij}(\mathbf{x};\boldsymbol\Theta),
\end{equation}
solvable through stochastic gradient descent after
fixing the network parameters $\boldsymbol\Theta$,
even though it is only feasible for a limited number of layers and not
guaranteed to converge~\cite{Erhan2009}.
%

SPNs lend themselves to an analogous problem formulation whose 
solution can be found without expensive iterative optimization.
Since in an SPN $S$ each inner node $n$ recursively defines a
probabilistic distribution over its scope $\mathsf{sc}(n)$,
maximizing its
activation reduces to find its MPE assignment over its scope---the \emph{mode} of the
distribution $S_{n}$.
%
%
Hence, one can reframe the problem in
Equation~\ref{eq:vis-sgd} as:
\begin{equation}
  \label{eq:vis-mpe}
  \mathbf{x}^{*}_{|\mathsf{sc}(n)} = \argmax_{\mathbf{x}}S_{n}(\mathbf{x}_{|\mathsf{sc}(n)}; \mathbf{w}).
\end{equation}

This suggests that even the
scope alone conveys semantics about the learned representations.
Indeed, for image samples
the visualization of the scope of each node corresponds to a shape against a background.
The meaningfulness of such representations therefore correlates
to the scope arrangement in the network.

%
%
%

\begin{figure}
  \centering{
  \begin{minipage}[c]{0.05\columnwidth}
    \subfloat[]{ \ \label{fig:mpe-filters-ocr}}
  \end{minipage}
  \begin{minipage}[c]{0.9\columnwidth}
    \centering
    \includegraphics[width=0.7\columnwidth]{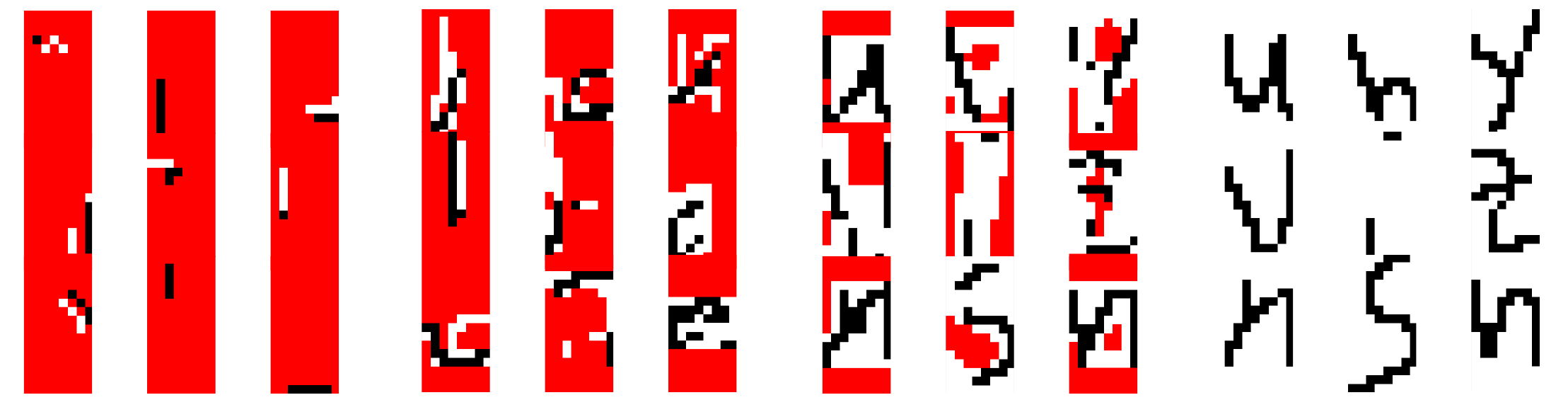}
  \end{minipage}\\
  \begin{minipage}[c]{0.05\columnwidth}
    \subfloat[]{ \ \label{fig:mpe-filters-cal}}
  \end{minipage}
  \begin{minipage}[c]{0.9\columnwidth}
    \centering
    \includegraphics[width=0.7\columnwidth]{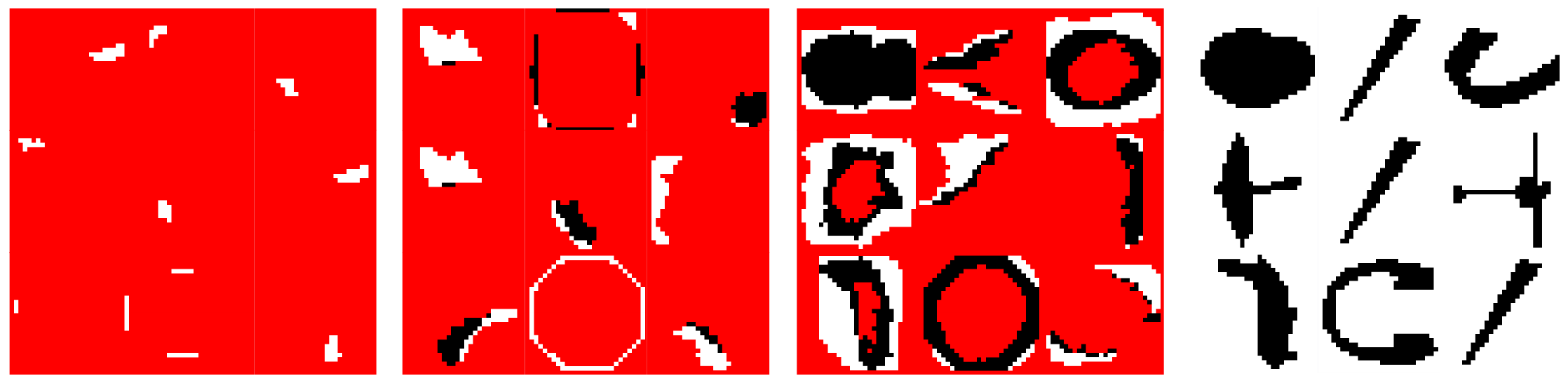}
  \end{minipage}\\
  \begin{minipage}[c]{0.05\columnwidth}
    \subfloat[]{ \ \label{fig:mpe-filters-bmn}}
  \end{minipage}
  \begin{minipage}[c]{0.9\columnwidth}
        \centering
    \includegraphics[width=0.7\columnwidth]{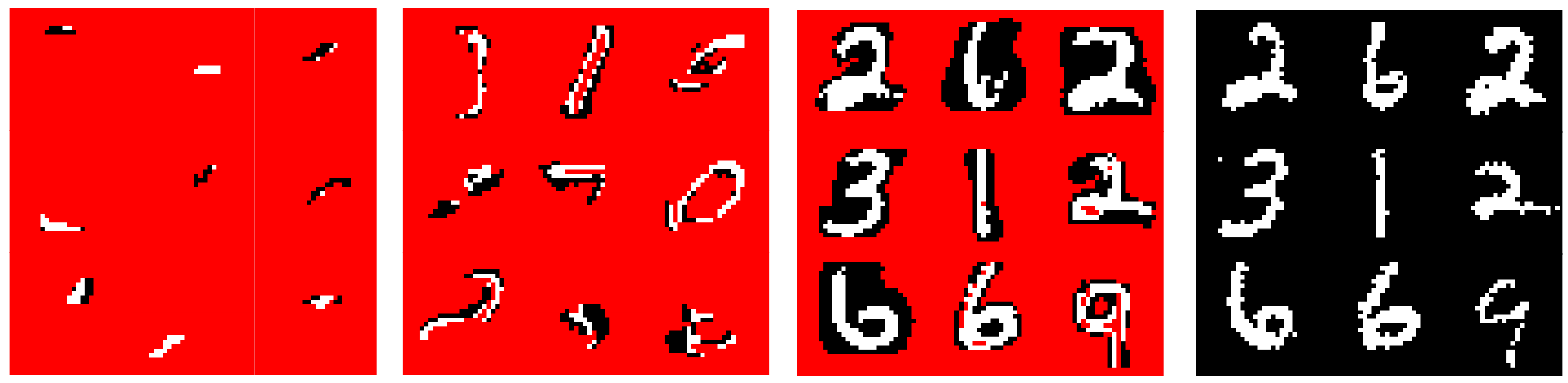}
  \end{minipage}
  }
  \caption{\emph{Representations visualization.}
    Visualization of representations of nodes sharing a similar
scope length of increasing size ranges: small, medium and large (columns 1 to 3). For each scope range, representations are extracted from \textsf{SPN-III}
    models, according to Eq. \ref{eq:vis-mpe}, on
    \textsf{OCR} (a), \textsf{CAL} (b) and \textsf{BMN} (c) from 9 randomly selected nodes. For each node $n$, pixels corresponding to  variables in \textsf{sc}($n$) are colored black (resp. white) when their MPE assignment is $1$ (resp. $0$); pixels corresponding to variables not belonging to \textsf{sc}($n$) are colored red, highlighting how recognizable \emph{per se} are the shapes induced by scopes.
    Column 4 shows the training images nearest to those in column 3.}
  \label{fig:mpe-filters}
\end{figure}

To verify the validity of the scope
length heuristics as a proxy for the abstraction level of
a representation, we inspect the representations of nodes in
Figure~\ref{fig:mpe-filters}.
There, visualizations for the representations learned have
  been selected by inspecting the scope length distributions of the
considered model (see Figure~\ref{fig:hist-scopes}), devising ranges for (S)mall,
(M)edium and (L)arge scope lengths (see Section~\ref{section:rle-fil})
for the largest SPN models we learned in Section~\ref{section:rle}.
From there, we randomly extracted 9 neurons for each scope length range.~\footnote{The randomness of the selection is visible in Figure~\ref{fig:mpe-filters} for the mid level
representations for CAL---two nodes not only share the same scope but the same most
activating input image as well. This might happen if a sum node and one of its maximising
children are both chosen. Clearly, this highly structured part-based
hierarchy cannot be equally visualized on smaller networks,
e.g. \textsf{SPN-I} model on \textsf{CAL} (see Section~\ref{section:rle}).}

Even if 
spatial autocorrelation
is not taken into account while learning the structure of the SPN
considered (see Section~\ref{section:rle}), node scopes naturally
capture recognizable part shapes.
Clearly, this is not the case for higher-level features, i.e., features
associated to scope lengths covering almost all the image.


%
%
%

The compositionality of the
learned representations is evident through the different levels of the hierarchy: the longer the scope
the higher the level of abstraction.
The visualized features indeed resemble part-based
filters at different levels of complexity: from pixel blobs to shape
contours, to full shapes comprising background parts.
%
We can therefore confirm the role of SPN nodes as probabilistic
part-based filters. 
%
Lastly, by comparing the higher-level features extracted to the nearest
training images one can inspect if 
they are their exact reconstructions.
This is not the case for the features in Figure~\ref{fig:mpe-filters},
even though they appear to be very
specialized filters.
We evaluate how this relates to the predictive performances of these
representations in Section~\ref{section:rle}.

\section{Representation Learning with SPNs}
\label{section:rle}

Here we empirically evaluate SPNs as feature extractors in
a classical RL framework.
We exploit embeddings extracted by different filtering criteria---leading
to different feature sets---to train a
classifier to predict unseen target RVs.
%
We use the accuracy of such a classifier as a proxy
measure to assess the \emph{usefulness and effectiveness} of these representations~\cite{6472238}.
We point out how we are not interested in achieving state-of-the-art
accuracy scores on the dataset employed.
Instead, our aim is to investigate how competitively the embeddings
filtered by the proposed criteria rank
when compared against themselves and the ones extracted by commonly
employed \emph{generative} models for RL
like RBMs~\cite{Smolensky1986}, DBNs~\cite{Salakhutdinov2009}, 
MADEs~\cite{Germain2015}, and recently introduced VAEs~\cite{Kingma2013}.

\begin{table}
  \caption{Datasets statistics: classes ($c$), dimensions ($n$) and
    samples number ($m$).}
  \label{data-stats}
  \centering
  \setlength{\tabcolsep}{3pt}  
  \small
  \begin{tabular}{lrlr|lrlr}
    \toprule
     & $c$ & $n$ & $m$ & &  $c$
    & $n$ & $m$ \\
    \midrule
    \textsf{REC} & 2 & 784 ($28\times 28$) &1000/200/50000 &
    \textsf{CAL} & 101 &784 ($28\times 28$) & 4100/2264/2307\\
    \textsf{CON} & 2 & 784 ($28\times 28$) &6666/1334/50000 &
    \textsf{BMN} & 10 & 784 ($28\times 28$) & 50000/10000/10000\\
    \textsf{OCR} & 26 & 128 ($16\times 8$) & 32152/10000/10000\\

    \bottomrule
    \end{tabular}
\end{table}

\subsection{Experimental setting}
\begin{figure}
  \centering
  \begin{minipage}[c]{0.03\columnwidth}
    \
  \end{minipage}
  \begin{minipage}[c]{0.96\columnwidth}
    \ \hspace{25pt} \small\textbf{\textsf{SPN-III}} \hfill
    \textbf{\textsf{MADE-1k}} \hfill
    \textbf{\textsf{DBN-1k}} \hfill
    {\textbf{\textsf{VAE-1k}}} \hspace{26pt}
  \end{minipage}  \\[2pt]  
  \begin{minipage}[c]{0.03\columnwidth}
    {\raisebox{6pt}{\rotatebox[origin=c]{90}{\small\textbf{\textsf{REC}}}}}
  \end{minipage}
  \begin{minipage}[c]{0.96\columnwidth}
    \includegraphics[height=12pt,trim={0 30mm 0 0},clip]{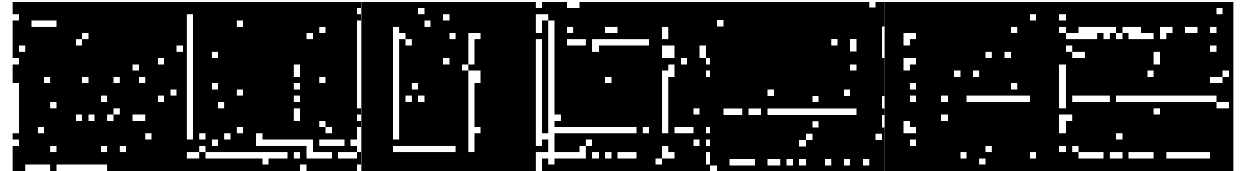} \hfill
    \includegraphics[height=12pt]{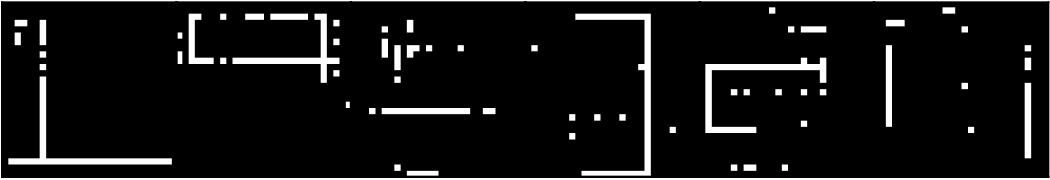} \hfill
    \includegraphics[height=12pt]{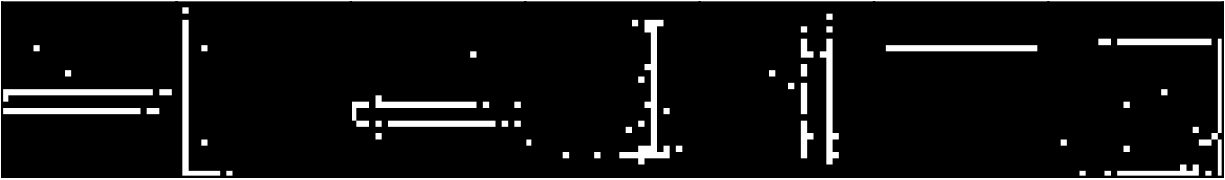} \hfill
    \includegraphics[height=12pt]{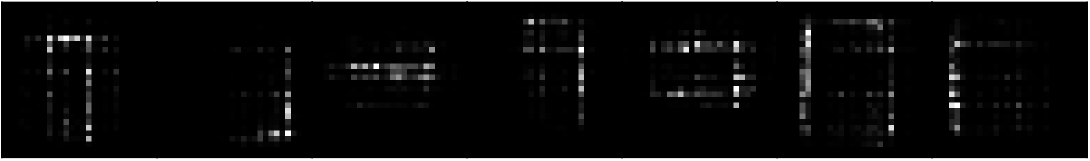}      \\
      \includegraphics[height=12pt,trim={0 0 0 30mm},clip]{rectangles-50-7-samples.pdf} \hfill
      \includegraphics[height=12pt]{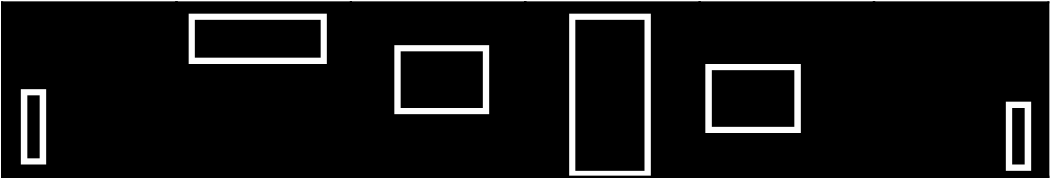} \hfill
      \includegraphics[height=12pt]{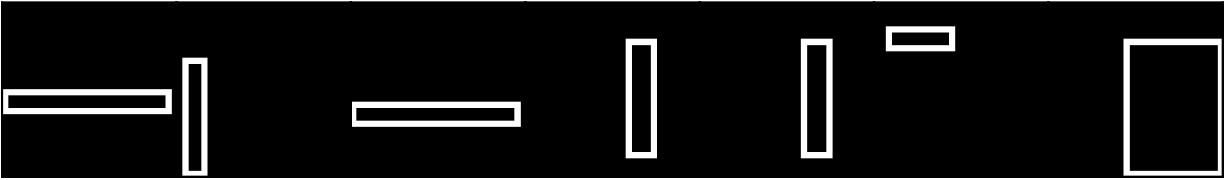} \hfill
      \includegraphics[height=12pt]{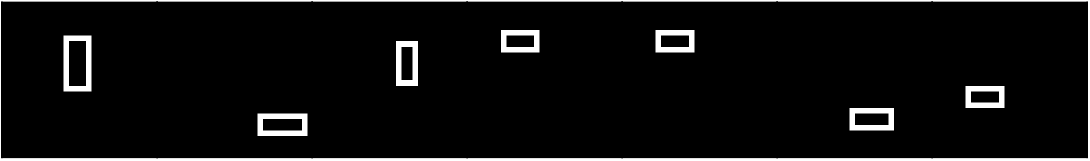}
  \end{minipage}      
    \\[5pt]
  \begin{minipage}[c]{0.03\columnwidth}
    {\raisebox{6pt}{\rotatebox[origin=c]{90}{\small\textbf{\textsf{CON}}}}}       
  \end{minipage}
  \begin{minipage}[c]{0.96\columnwidth}
    \includegraphics[height=12pt,trim={0 30mm 0 0},clip]{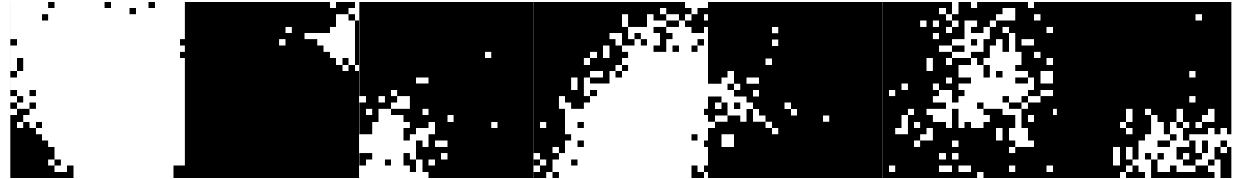} \hfill
    \includegraphics[height=12pt]{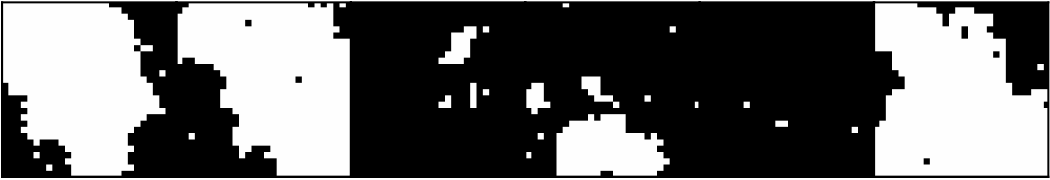} \hfill
    \includegraphics[height=12pt]{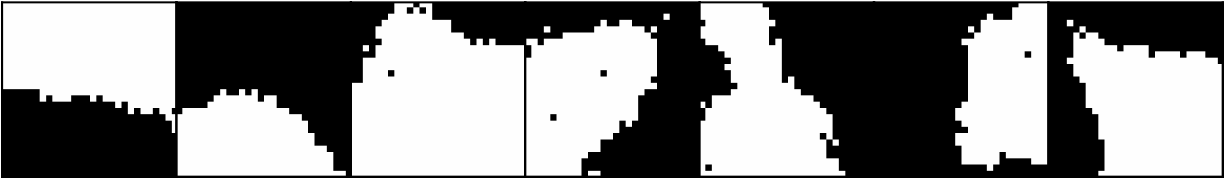} \hfill
    \includegraphics[height=12pt]{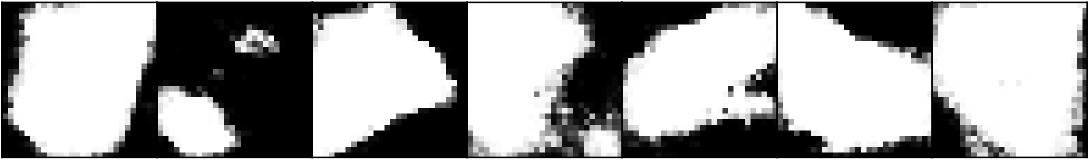}\\
    \includegraphics[height=12pt,trim={0 0 0 30mm},clip]{convex-50-7-samples.pdf} \hfill
    \includegraphics[height=12pt]{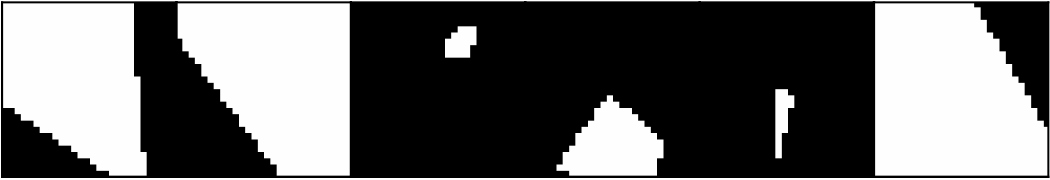} \hfill
    \includegraphics[height=12pt]{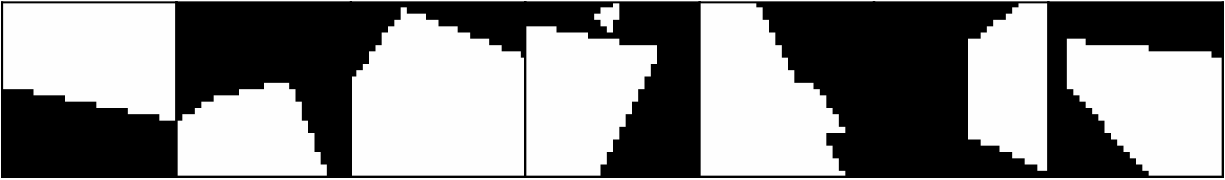} \hfill
    \includegraphics[height=12pt]{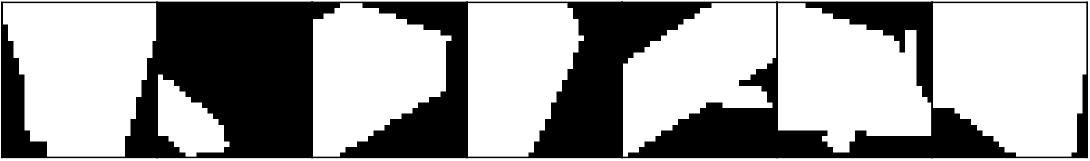}
  \end{minipage}
    \\[5pt]
  \begin{minipage}[c]{0.03\columnwidth}
    {\raisebox{6pt}{\rotatebox[origin=c]{90}{\small\textbf{\textsf{OCR}}}}}    
  \end{minipage}
  \begin{minipage}[c]{0.96\columnwidth}
    \includegraphics[height=12pt,trim={0 30mm 0 0},clip]{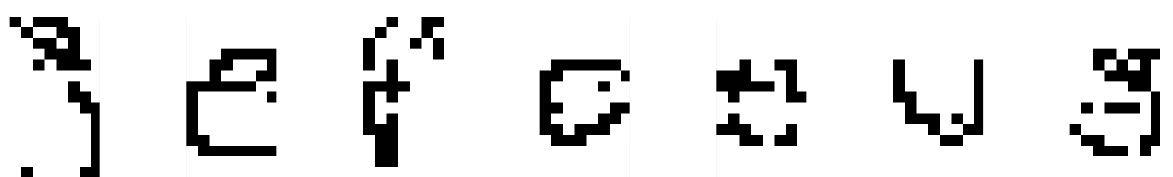} \hfill
    \includegraphics[height=12pt]{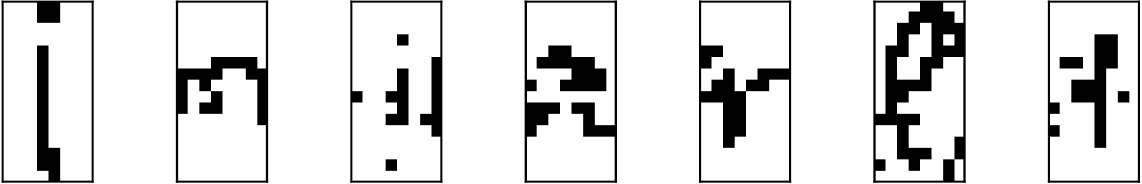} \hfill
    \includegraphics[height=12pt]{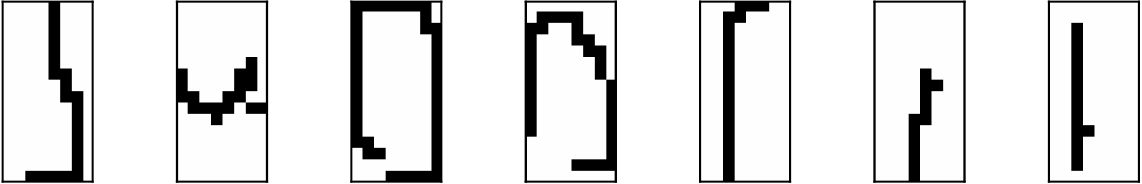} \hfill
    \includegraphics[height=12pt]{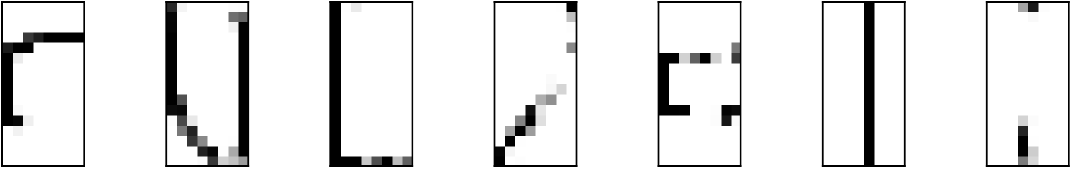}\\
    \includegraphics[height=12pt,trim={0 0 0 30mm},clip]{ocr_letters-50-7-samples.pdf} \hfill
    \includegraphics[height=12pt]{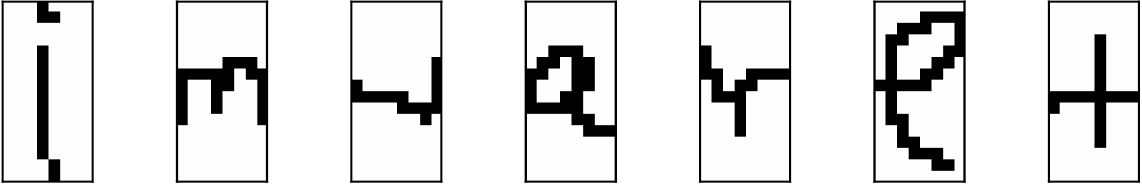} \hfill
    \includegraphics[height=12pt]{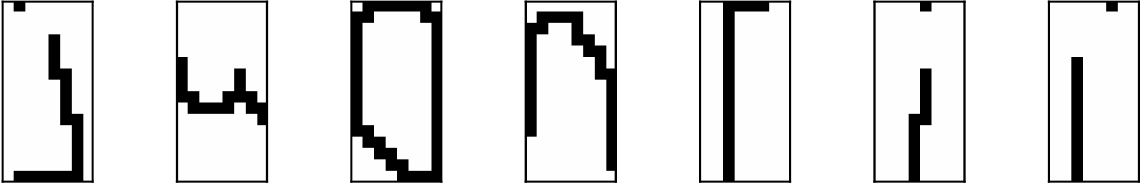} \hfill
    \includegraphics[height=12pt]{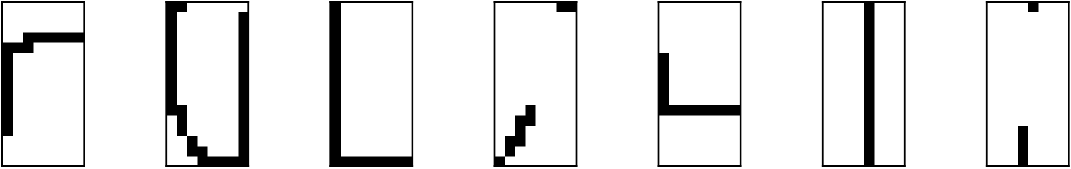}
  \end{minipage}
    \\[5pt]
  \begin{minipage}[c]{0.03\columnwidth}
    {\raisebox{6pt}{\rotatebox[origin=c]{90}{\small\textbf{\textsf{CAL}}}}}
  \end{minipage}
  \begin{minipage}[c]{0.96\columnwidth}
    \includegraphics[height=11pt,trim={0 30mm 0 0},clip]{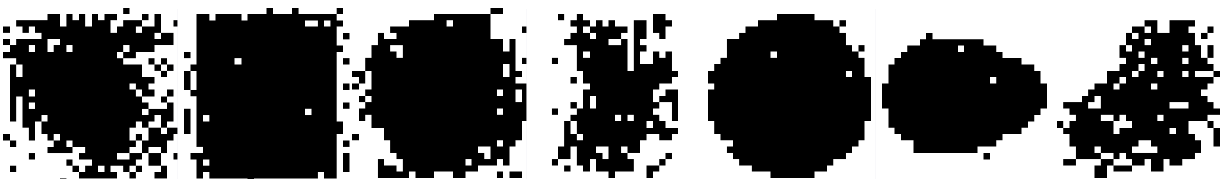} \hfill
    \includegraphics[height=11pt]{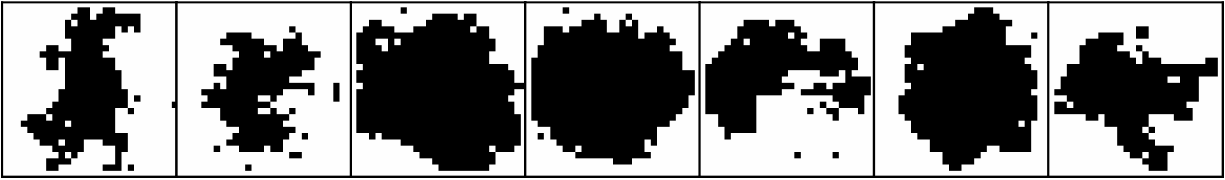} \hfill
    \includegraphics[height=11pt]{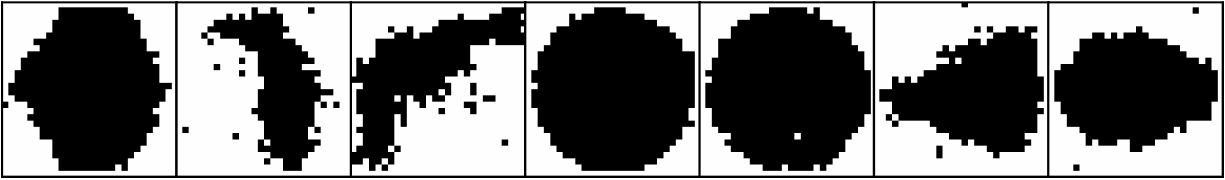} \hfill
    \includegraphics[height=11pt]{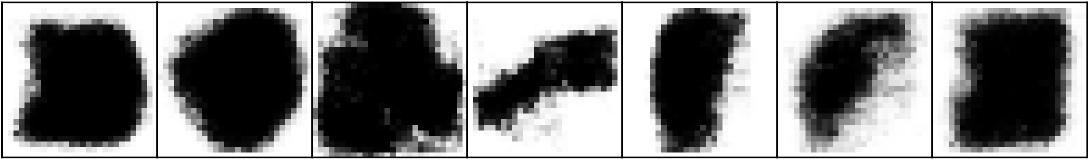}\\
    \includegraphics[height=11pt,trim={0 0 0 30mm},clip]{caltech101-50-7-samples.pdf} \hfill
    \includegraphics[height=11pt]{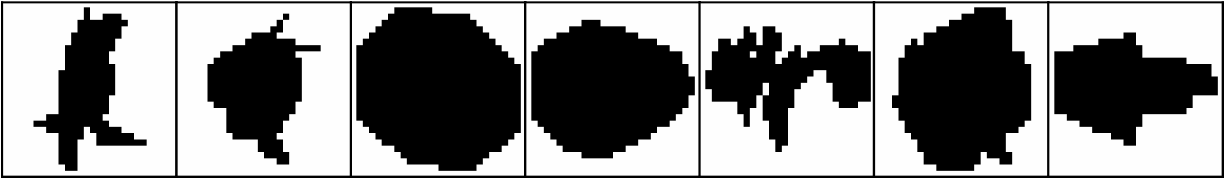} \hfill
    \includegraphics[height=11pt]{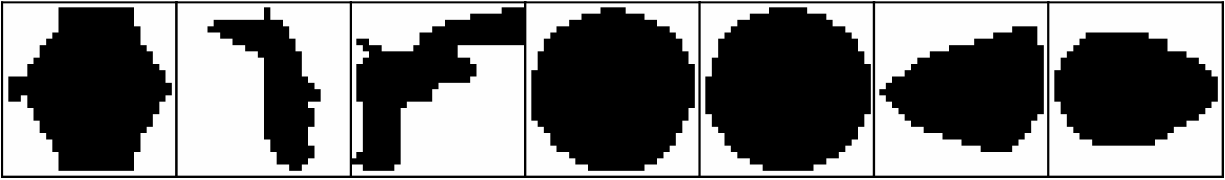} \hfill
    \includegraphics[height=11pt]{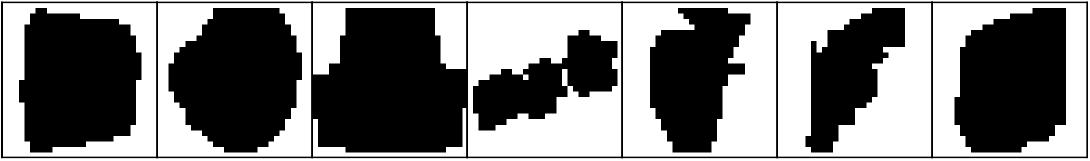}
  \end{minipage}
    \\[5pt]
  \begin{minipage}[c]{0.03\columnwidth}
          {\raisebox{6pt}{\rotatebox[origin=c]{90}{\small\textbf{\textsf{BMN}}}}}
  \end{minipage}
  \begin{minipage}[c]{0.96\columnwidth}
    \includegraphics[height=11pt,trim={0 30mm 0 0},clip]{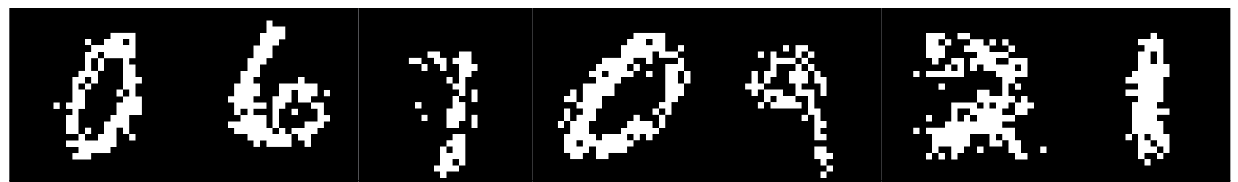} \hfill
    \includegraphics[height=11pt]{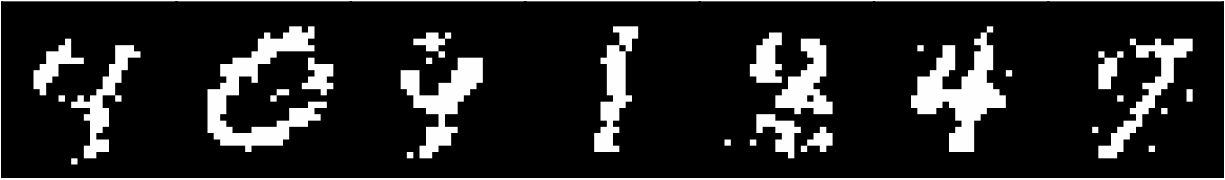} \hfill
    \includegraphics[height=11pt]{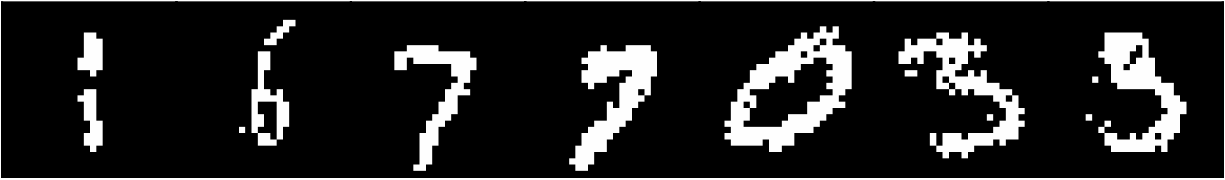} \hfill
    \includegraphics[height=11pt]{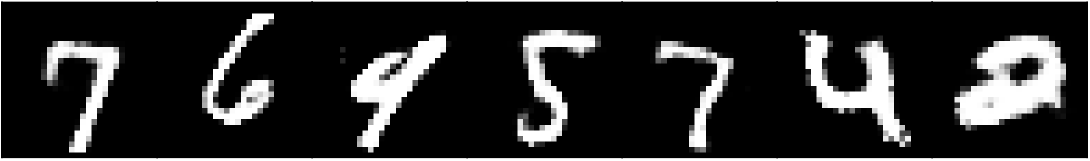}\\
    \includegraphics[height=11pt,trim={0 0 0 30mm},clip]{bmnist-50-7-samples.pdf} \hfill
    \includegraphics[height=11pt]{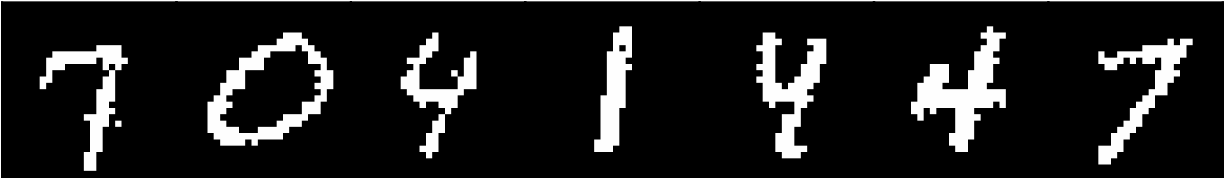} \hfill
    \includegraphics[height=11pt]{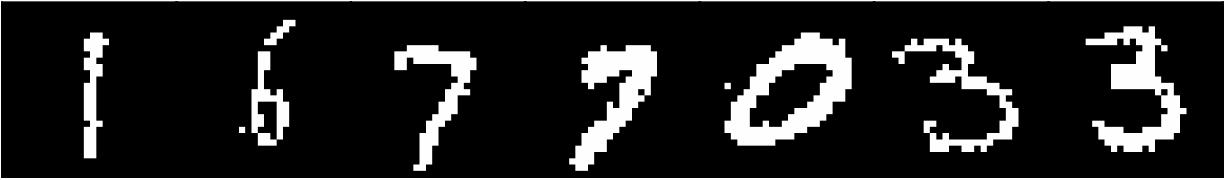} \hfill
    \includegraphics[height=11pt]{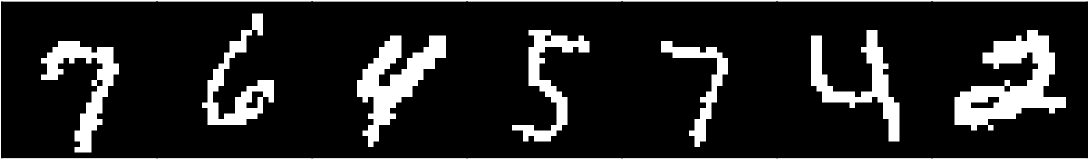}
  \end{minipage}
  \caption{\emph{Sampling.} Seven samples from \textsf{SPN-III}
    (1st col), \textsf{MADE-1k} (2nd col), \textsf{DBN-1k} (3rd col),
    and {\textsf{VAE-1k}} (4th col) models on the first row, and  their nearest neighbor images in the training set on the row below for 
    \textsf{REC},
    \textsf{CON},
    \textsf{OCR},
    \textsf{CAL}, and
    \textsf{BMN}.}
  \label{fig:samples}
\end{figure}

We employ five standard image classification
benchmarks for evaluating DNNs on disentangling many factors of variations~\cite{Larochelle2007}:
Rectangles (\textsf{REC})~\cite{Larochelle2007}, Convex (\textsf{CON})~\cite{Larochelle2007}, OCR Letters (\textsf{OCR})~\cite{Larochelle2011}, Caltech-101 Silhouettes (\textsf{CAL})~\cite{Marlin2010} and a binarized version
of MNIST (\textsf{BMN})~\cite{Larochelle2007}.
We preprocessed them as in their original works and report their
statistics  in Table~\ref{data-stats}.
Image samples are shown in Figure~\ref{fig:samples}.
Each dataset is a collection $\{\mathbf x^i\sim\mathbf{X}, y^i\sim
Y\}_{i=1}^m$ comprising samples in the original feature space
$\mathbf{X}$ and labeled by RV $Y$.
We learn all our reference models on the $\mathbf{X}$ alone---in an
\emph{unsupervised} and \emph{generative} way---discarding the class
information $Y$.
%

As competitors, we employ RBMs as they have been extensively proven to be solid feature
extractors~\cite{Larochelle2008,Marlin2010};
their deep version, DBNs~\cite{Salakhutdinov2009}, to measure the
influence of latent RVs layered in a hierarchy as for SPNs;
MADEs~\cite{Germain2015} as generative models
that are also autoencoders---thus innately suited for RL---which,
similarly to SPNs, exhibit a constrained and
labeled structure, as imposed by the autoregressive property;
and lastly VAEs~\cite{Kingma2013} as autoencoders
trained to be generative models.
For a fair comparison, and to avoid numerical issues, we collect
embeddings in the \emph{log} domain for all models dealing with probabilities. 
%
We also want to learn networks with different \emph{model
  capacities} in order to analyze how different structures, learned by
the same algorithm, affect the usefulness of differently sized
embeddings.

%
We learn RBMs having $500$, $1000$ and
$5000$ hidden units---providing embeddings of respective sizes---denoting them as \textsf{RBM-5h}, \textsf{RBM-1k} and
\textsf{RBM-5k}.
To generate embeddings, we evaluate the conditional
probabilities of the hidden units given each sample.
We train them using the Persistent Constrastive Divergence 
algorithm~\cite{Marlin2010}.
We select the learning hyperparameters by a grid search looking  for the learning rate in $\{0.1,
0.01\}$, the batch size in $\{20, 100\}$ and the number of epochs in
$\{10, 20, 30\}$ by comparing the best validation set pseudo-log-likelihoods.
%
%

%
For MADEs (resp. DBNs) we build architectures comprising 500 and 1000
hidden units up to 3
hidden layers, and we denote them as \textsf{MADE-5h} and
\textsf{MADE-1k} (resp. \textsf{DBN-5h} and \textsf{DBN-1k}).
For both DBN and MADE models, we extract embedding by concatenating
all the activations of all the nodes from all the layers, obtaining
embeddings of sizes 3000 and 4500, respectively.

Similarly, for VAEs we stack up to three levels in the encoder
comprising 500 or 1000 units each, denoting them as \textsf{VAE-5h} and
\textsf{VAE-1k}, but investigate different compression factors
$\{0.7, 0.8, 0.9\}$ for the bottleneck layer.
Again, we experimented with extracting representations from the
bottleneck layer alone or by concatenating all the layers of the
encoder, ultimately finding the latter to provide far more accurate
predictions.

For MADEs we employ the ADADELTA method to schedule learning rates with
decay rate $0.95$; we set to 30 the max number of worsening
iterations on the validation and a batch size of
100. 
We initialize weights by SVD.
Other hyperparameters are selected by a grid
search guided by the best validation set log-likelihoods.
We look for gradient dumping coefficients in $\{10^{-5}, 10^{-7},
10^{-9}\}$; we either set no mask cycling, and we set their
maximum number to 300, or we cycle over 32 random masks;
we investigate both ReLus and softplus as nonlinearities.

We select the DBNs hyperparameters by performing a grid search 
for the learning rate in $\{0.1, 0.01\}$, the batch size in $\{20, 100\}$
and the epoch numbers in $\{10, 20, 30\}$.
For VAEs we employed the ADAM method as an optimizer, running it up to 1000
epochs with a patience of 50, performing a grid search for batch size
$\{20, 100, 256\}$ and learning rate in $\{0.01, 0.001\}$ and setting
$\beta_1=0.9$, $\beta_2=0.999$ and no decay.
We investigate both ReLus and softplus as nonlinearities.

Differently from RBMs, DBNs, MADEs and VAEs, we can directly learn the
structure of our SPN models from data (see Section~\ref{section:spn}).
However, we do not have a direct way to control embedding sizes except for
regularizing the structure learning phase.
%
%
We employ \textsf{LearnSPN-b}~\cite{Vergari2015}, a variant of
\textsf{LearnSPN}, as a structure learner.
With the aim of slowing down the greedy hierarchical co-clustering
process, \textsf{LearnSPN-b} always splits samples and RVs into two
clusters, thus achieving deeper and more compact structures~\cite{Vergari2015}.
%
%
%
%
For each dataset we learn three differently regularized architectures
by early stopping, varying paramater $\mu\in\{500, 100, 50\}$, and denote them as \textsf{SPN-I}, \textsf{SPN-II} and \textsf{SPN-III} models respectively.
For all models we fix the
pairwise statistical independence test threshold $\rho$ always to 20 except for
\textsf{OCR}, for which it is 15.
%
%
We then perform a grid search to select the best  leaf distribution
smoothing factor 
$\alpha\in\{0.1, 0.2, 0.5, 1.0, 2.0\}$.
Table~\ref{tab:models} reports the learned SPN structural statistics.

\begin{table}[!t]
  \caption[datasets]{Structural statistics for the \textsf{SPN}
    reference architectures on \textsf{REC}, \textsf{CON}, \textsf{OCR},
    \textsf{CAL} and \textsf{BMN} datasets,  like number
 of nodes by type (sum, product, leaf), 
of unique scopes and the number of nodes for certain scope
lengths, since they correspond to the sizes of the embeddings
as filtered in Section~\ref{section:rle-fil}.}
  \centering
  \footnotesize
  \setlength{\tabcolsep}{4pt}  
  \begin{tabular}{l l l r r r r r r r r r}
    \toprule
    & & $\mu$ & depth & edges & sum  & prod  & leaves & unique &
                                                                      \multicolumn{3}{c}{
                                                                       scope length}\\
    & & &  &  & nodes & nodes & & scopes &\textsf{S} & \textsf{M} & \textsf{L}\\
    \midrule
    \multirow{3}{*}{\textsf{REC}} &\textsf{SPN-I} & $500$        & 5
                    & 2240     & 5  &   10       &2226 & 789 & 3&6&6\\
    & \textsf{SPN-II}   & $100$ & 15& 8145 & 163  &  327  &7656&946&108&354&28 \\
    & \textsf{SPN-III}  & $50$  & 15& 9424 & 265 &  531 &8629&1045& 231 & 537 & 28\\
    \midrule
    \multirow{3}{*}{\textsf{CON}} &\textsf{SPN-I} & $500$        & 7
                    & 13019         & 13  &   33       &12974&797 & 6&0&40\\
    & \textsf{SPN-II}   & $100$     & 15             & 50396
                            & 308  &  627        &49462&1083&573&90&272 \\
    & \textsf{SPN-III}  & $50$       & 17             & 81330
                            & 1872 &  3755        &75704 &2439& 3849 & 1302 & 476\\
    \midrule
    \multirow{3}{*}{\textsf{OCR}} & \textsf{SPN-I} & $500$ & 17 & 7848 & 64 & 163 &7622&191 & 18 & 42 &167\\
    & \textsf{SPN-II}  & $100$   & 23  & 35502  & 1972 & 4005  &29526&1537&3465&2033&479 \\
    & \textsf{SPN-III} & $50$    & 23  & 48548  & 4069 & 8200 &36280&2159 & 8844 &2940 &485 \\
    \midrule
    \multirow{3}{*}{\textsf{CAL}} & \textsf{SPN-I} & $500$        & 9
                        & 8102 & 10 & 22 &8071&794 & 3 & 0& 29\\
    & \textsf{SPN-II} & $100$        & 17    & 32267    & 206 & 415 &31647&987 & 387 & 63 & 171\\
    & \textsf{SPN-III} & $50$        & 19    & 53121    & 1821 & 3645 &47656 &2340& 3777 & 1434 & 255\\
    \midrule
    \multirow{3}{*}{\textsf{BMN}} &\textsf{SPN-I} & $500$        & 19
                        & 47215         & 184  &   370       &46662 &967&
                                                                      99&33&422\\
    & \textsf{SPN-II}   & $100$     & 25             & 168424
                                   & 5493  &  10990        &151942&4487&10049&5034&1400 \\
    & \textsf{SPN-III}  & $50$       & 27             & 198573
                                   & 10472 &6172&  20948        &6172 &
                                                                     21992
    & 8031 & 1397\\
    \bottomrule
  \end{tabular}
   \label{tab:models}
 \end{table}





At first, we peek at the effect of different model capacities over the representations learned by all
models by employing them as generative models and \emph{visually inspecting
  samples generated from them}.
For SPNs we employ the sampling scheme we introduced in Section~\ref{section:spn}.
We check if  models have learned
representations just able to reconstruct the training
set~\cite{Larochelle2011,Germain2015},
by comparing samples against the nearest,
in the sense of the Euclidean distance, training samples.
Samples from our least regularized SPN, \textsf{SPN-III}, are compared
against those from \textsf{DBN-1k} and
\textsf{MADE-1k} models in Figure~\ref{fig:samples}.
%
%
The presence of noise is evident for all models and datasets, with
\textsf{REC} and \textsf{CON} being the hardest datasets.
\textsf{DBN-1k} generated images are generally more recognizable,
however they are very close to their training counterparts.
This might suggest a form of overfitting for DBN models.
The proximity of the generated samples w.r.t. training images is even
more prominent for \textsf{VAE-1k} models, but expected in this case,
as they are trained to explicitly reconstruct their inputs.
 Additionally, we note how \textsf{SPN-III} struggles
to capture some straight lines on \textsf{REC},
differently from \textsf{DBN-1k}, hinting at 
SPNs  not modeling some spatial correlations in the data.
The extent to which these conjectures will affect the predictive power
of the extracted embeddings
 is investigated in Sections~\ref{sec:node-act}, \ref{section:rle-fil}, \ref{section:rle-sca} and \ref{section:rle-semi}.

%
%


\subsection{Supervised representation learning with SPNs}
\label{sec:node-act}

Generally, we are interested in representing a sample
$\mathbf{x}^{i}\sim\mathbf{X}$ as an \emph{embedding} $\mathbf{e}^{i}$ in a new $d$-dimensional
space $\mathbf{E}_{\mathbf{X}}\subseteq\mathbb{R}^{d}$
through a transformation $f_{\theta}$ provided by some model $\theta$, i.e., $f_{\theta}(\mathbf{x}^{i})=\mathbf{e}^{i}$.
For an SPN $S$, $f_{S}$ clearly is determined by the structure and
parameters of $S$.
Specifically, let $\nodeset{N}=\{n_{j}\}_{j=1}^{d}\subseteq\nodeset{S}$
 be a set of nodes in $S$, filtered by a certain criterion.
We build $\mathbf{e}^{i}$ by collecting the activations of nodes in $\nodeset{N}$, i.e.,
$e_{j}^{i}=S_{n_{j}}(\mathbf{x}^{i})$.
{
  Therefore an embedding is the collection of the features expressed
  by a set of neurons, which could be potentially be visualized
  individually as showed in Section~\ref{section:vis}, by leveraging
  the fact that \emph{they are indeed probabilities}.}

%

In all  experiments we train a
\emph{linear} classifier  to predict
$Y$ from the embeddings extracted by our SPN and competitor models.
The rationale here is to inspect if the
new geometric space $\mathbf{E}_{\mathbf{X}}$ has disentangled the
input space enough to let a linear separator easily discriminate the
 classes in $Y$~\cite{6472238}.
We employ a logistic
regressor with an L2 regularizer in a one-versus-rest setting.
%
%
We determine the L2 regularization
coefficient $C$ for each experiment in $\{0.0001, 0.001, 0.01, 0.1,
1.0\}$. 
As the simplest baseline, denoted as \textsf{LR}, we apply such a classifier
directly on the original feature space $\mathbf{X}$.
{To test the statistical significance of the differences for the accuracies reported in Tables~\ref{tab:model-accs}, \ref{tab:model-filter-accs}, \ref{tab:model-scope-aggr} and \ref{tab:model-accs-ssl}, we applied a 
paired signed rank Wilcoxon test, using a p-value of 0.05 for each possible pair of
competitors. A result is rendered in bold if it is statistically better than all other non-bold
results for a certain experimental setting.~\footnote{Two or more results for a single experimental
setting are in bold if they were not deemed to be statistically different, and all of them are
significantly better than all the rest.}}

\begin{table}
  \centering
  \small
  \caption[datasets]{Test set accuracy scores for full embeddings
    extracted with different \textsf{SPN},
    \textsf{RBM}, \textsf{DBN}, \textsf{MADE} and {\textsf{VAE}} models, compared to
    the baseline \textsf{LR} model on all
    datasets. Bold values denote significantly better scores
    than all the others for a dataset.}
  \setlength{\tabcolsep}{2pt}  
  \begin{tabular}{l c@{\hskip 6pt} c c c@{\hskip 6pt} c c c@{\hskip
    6pt} c c@{\hskip 6pt} c c@{\hskip 6pt} c c}
    \toprule
    
     &  & \multicolumn{3}{c@{\hskip 12pt}}{\textsf{SPN}}
    & \multicolumn{3}{c@{\hskip 12pt}}{\textsf{RBM}} &
                                         \multicolumn{2}{c@{\hskip 12pt}}{\textsf{DBN}}
    & \multicolumn{2}{c@{\hskip 2pt}}{\textsf{MADE}}& \multicolumn{2}{c@{\hskip 2pt}}{\textsf{VAE}}\\
    \textsf{data}& \textsf{LR}&\scriptsize\textsf{I}&\scriptsize\textsf{II} & \scriptsize\textsf{III}
    & \scriptsize\textsf{5h} & \scriptsize\textsf{1k} & \scriptsize\textsf{5k} & \scriptsize\textsf{5h} & \scriptsize\textsf{1k}&
    \scriptsize\textsf{5h} & \scriptsize\textsf{1k} & \scriptsize\textsf{5h} & \scriptsize\textsf{1k}\\
  
    \midrule
    \textsf{REC} & 69.28 & 77.31& \textbf{97.77}& 97.66& 94.22& 96.10&
                                                                       96.36
                                            & 94.47 & 95.46&82.38& 86.60&95.80&96.57\\
    \textsf{CON} & 53.48 & 67.48& 78.31& \textbf{84.69}& 67.55& 75.37& 79.15 & 81.12&81.61&71.96& 76.90&79.45&78.89\\
    \textsf{OCR} & 75.58 & 82.60& \textbf{89.95}& \textbf{89.94}& 86.07& 87.96& 88.76 &87.48&88.21& 84.40&84.18&86.38&86.89\\
    \textsf{CAL} & 62.67& 59.17 & 65.19& 66.62& 67.36& 68.88&                                                                       67.71 &69.53& \textbf{69.60}& 61.94&64.76&64.62&64.67\\
    \textsf{BMN} & 90.62& 95.15& \textbf{97.66}& 97.59& 96.09& 96.80& 97.47 & 97.06&97.51&94.10&95.18&96.98&97.10\\
    \bottomrule
  \end{tabular}
  \label{tab:model-accs}
\end{table}

%
We are {firstly} interested in evaluating full embeddings comprising all nodes
in a network, {as the initial, simplest and uninformative attempt at measuring the
predictive power of probability activations in SPN.}
Clearly, considering leaf node activations would result in
too large embedding sizes (see Table~\ref{tab:models} and
Section~\ref{section:int}).
Therefore, as the first filtering criterion, we consider \emph{full
embeddings comprising only
  inner nodes} in $S$, i.e., $\nodeset{N}=\nodeset{S}^{\oplus}\cup\nodeset{S}^{\otimes}$. 
We devise a way to efficiently include the contribution of leaf
nodes in Section~\ref{section:rle-sca}.
{More generally, larger embedding sizes, even if it would definitely help a linear
classifier better discriminate classes,
 could also let it suffer from the curse of dimensionality.
}

Test accuracy scores for all  datasets, for
\textsf{LR}, \textsf{SPN}, \textsf{RBM}, \textsf{DBN}, \textsf{MADE} and \textsf{VAE} models are reported in
Table~\ref{tab:model-accs}. 
Some datasets are inherently harder than others: if the \textsf{LR} baseline
scores $90.6$\% of accuracy on the 10 classes of \textsf{BMN}---indicating the original
feature representations to be disentangled enough---on the binary
\textsf{CON} dataset it scores only $53.5$\%.
All embeddings perform better
than the LR baseline, proving their effectiveness in disentangling factors
of variations, the only exceptions being
\textsf{SPN-I} and \textsf{MADE-5h} models on
\textsf{CAL}. However,
while the latter embeddings comprise 500 values,
the former only 32 (see Table~\ref{tab:models}), therefore acting as a
remarkable compressor for the original 784-dimensional space.
%
%
Generally, there is a constant improvement by adopting
less regularized SPN models,
even if the accuracy difference between \textsf{SPN-II} and
\textsf{SPN-III} models can be negligible at times.
%

On all datasets, with the exception of \textsf{CAL}, accuracies of SPN embeddings
are better or competitive w.r.t. all other models.
While additional hidden layers in DBNs make them slightly better than
RBMs, the same is not true for MADE models, which underperform on almost
all datasets.
{Remarkably, with the exception of \textsf{REC}, VAEs also underperform
w.r.t. RBMs and DBNs.}

%
It is remarkable how effective SPN embeddings are, considering the
simple and greedy way in which both the network structures and
parameters are unsupervisedly learned.
%
The best accuracy on \textsf{REC}, $97.77\%$ held by \textsf{SPN-II} is very close to the best 
score achieved by a \emph{fully supervised} learner
in~\cite{Larochelle2007}: $97.85\%$ by an SVM.
On \textsf{CON}, 
\textsf{SPN-III} scores a significantly higher accuracy than the
best supervised model
in~\cite{Larochelle2007}: $84.69\%$ versus the $81.59\%$ achieved
by stacked autoencoders.
This proves the practical utility of SPNs trained as generative models
 when plugged into predictive tasks:
 one obtains an expressive and tractable density estimator and, \emph{at the
 same time and without retraining it}, can effectively  extract
effectively competitive features from it.
{It is worth asking if this performance gain is due  
to better modeling the data distributions. 
The answer is negative, since MADE log-likelihoods are higher that SPN
ones.~\footnote{The trained RBM and DBN models do not allow to compute comparable
  log-likelihoods and comparing pseudo-log-likelihoods is not immediate.} 
%
We argue the effectiveness of SPN embeddings lies in the hierarchical
part-based representations they provide, which are confirmed by the
visual inspection of our models, as provided in
Section~\ref{section:vis}.
The positive effect of dealing with part-based representations in
predictive tasks has been, indeed confirmed more than once in the
literature, e.g. in~\cite{Agarwal2004,Felzenszwalb2010}.
This, in turn, relates to how SPNs are learned
by $\mathsf{LearnSPN}$-like algorithms:
while performing a form of hierarchical co-clustering over the data matrix,
they implicitly discover meaningful ways to discriminate among data at different
levels of granularity. 
}

\subsection{Filtering embeddings by node type and scope length}
\label{section:rle-fil}

It worth looking for the nodes in an SPN most responsible for the
surprisingly high accuracy scores obtained in
Section~\ref{sec:node-act}.
{We do this as a means to
reduce the size of SPN embeddings---as simply collecting all node
activations is an unsatisfactory criterion easily suffering from the
curse of dimensionality---and also to assess the importance of the
representations at different levels of abstraction, confirming the scope
length heuristics we propose.
}
Note that selecting only a subset of nodes of the
network as feature extractors is not the same as having a network
composed only by those nodes---the contributions of the nodes filtered out are still present, even if
indirectly, in the output activations of the collected nodes.

We apply the following filtering criteria to the
inner node embeddings extracted previously.
At first, we filter them by node type, to
evaluate whether there is a pattern in \emph{{sum}} ($\nodeset{N}=\nodeset{S}^{\oplus}$) versus \emph{{product node embeddings}} ($\nodeset{N}=\nodeset{S}^{\otimes}$).
Orthogonally, we filter nodes w.r.t. their {\emph{scope length}}
according to the heuristics
about the hierarchy of abstractions as presented in
Section~\ref{section:int}.
Based on the visualization on the scope length distributions provided
there,
 we define (\textsf{S})mall scope lengths, comprising 2 to 3 RVs;
scopes of (\textsf{M})edium length containing up to 100 RVs for all
datasets, except for  \textsf{OCR} where it is 50;
and lastly, (\textsf{L})arge length embeddings including all remaining
lengths.
We filter in this way  only to embeddings
from \textsf{SPN-III} models, as their scope length distributions have shown the highest variance.
Test accuracy results for the five filtering criteria are reported in Table~\ref{tab:model-filter-accs}.
%
For SPNs with fewer nodes, the product nodes seem to contribute
the most to the scored performance.
On the other hand, when the model capacity is enough, e.g., with
\textsf{SPN-III} models, \emph{{sum
nodes act as efficient compressors}}, greatly reducing the embedding
size (cf. Table~\ref{tab:models}) and
preserving the accuracy achieved by the full embeddings, or even
improving it.
{This behavior is similar to what happens to
max pooling in convolutional neural architectures, even though here we have \emph{aggregations by
weighted averages} as we are dealing with mixtures of valid probabilities.
}
More generally, a \emph{{holistic effect}} can be observed, sum and
product nodes
perform better together than when considered separately, even if slightly,
and even when the
size of a full embedding could suffer from the curse of dimensionality.
As reported in Table~\ref{tab:model-filter-accs}, embeddings from the smallest scope lengths
are always the
less accurate than both the full version and the ones filtered from longer scope lengths.
Even if they are the embeddings with the largest size (cf. Table~\ref{tab:models}), the meaningfulness of the extracted features
is minimal, as conjectured in Section~\ref{section:int}.
However, also \emph{{the contribution of the higher-level features is less
prominent}}.
This confirms the intuition we had through the filter
visualizations in Section~\ref{section:vis}: high level features in our reference models may be too 
specialized.
As a result, in general, selecting only mid-level features proved itself to be a meaningful
way to extract compressed, but still accurate, embeddings.

Filtered SPN embeddings
are smaller than the RBM, DBN, MADE and {VAE} counterparts
while their accuracies are comparable
or better on three datasets out of five.
The filtering process
also improves the scores reported in Section~\ref{sec:node-act}
against fully-supervised models: e.g., the $97.80\%$ accuracy on \textsf{BMN} achieved by the sum nodes of
\textsf{SPN-III}, or the $98.45\%$ scored by \textsf{M} scope length
embeddings on \textsf{REC}.



\begin{table}
  \caption[datasets]{Test set accuracy scores for the embeddings
     filtered by node
    type (columns 2-7) and for \textsf{SPN-III} embeddings filtered by \textsf{S}mall
    , \textsf{M}edium and \textsf{L}arge scope
    lengths (columns 8-10).
    Bold values denote significantly better scores than all the others.
    $\blacktriangle$ indicates a better score than competitor embeddings
    with greater or equal size. $\triangledown$ indicates worse
    scores than competitor embeddings with smaller or equal size.
  The last columns  report the accuracies 
for the \textsf{SPN-III} embeddings filtered by the scope length ranges
\textsf{S}, \textsf{M}, \textsf{L}.}
  \centering
  \small
  \setlength{\tabcolsep}{3pt}  
  \begin{tabular}{l l l l l l l | l l l}
    \toprule
    & \multicolumn{2}{c}{\textsf{SPN-I}} &
                                           \multicolumn{2}{c}{\textsf{SPN-II}}
    & \multicolumn{2}{c}{\textsf{SPN-III}} & \multicolumn{3}{c}{\textsf{SPN-III}}\\
    
    \textsf{dataset}&sum  & prod& sum& prod& sum& prod & \textsf{S} & \textsf{M} & \textsf{L}\\
    \midrule
    \textsf{REC} & 72.46& 62.25& $\mathbf{98.03}^{\blacktriangle}$& $97.06^{\blacktriangle}$& $\mathbf{98.00}^{\blacktriangle}$& $97.04^{\blacktriangle}$ & 88.73&$\mathbf{98.45}^{\blacktriangle}$& 93.91\\
    \textsf{CON} & 62.36& 64.03& $77.13^{\blacktriangle}$& $76.07^{\blacktriangle}$& $\mathbf{83.59}^{\blacktriangle}$& $82.06^{\blacktriangle}$ &$70.51^{\triangledown}$&77.18&$\mathbf{83.32}^{\blacktriangle}$\\
    \textsf{OCR} & 74.19& 81.58& $89.73^{\blacktriangle}$& $88.78^{\blacktriangle}$& $\mathbf{90.02}^{\blacktriangle}$& 89.32 & $87.22^{\triangledown}$& $\mathbf{89.29}^{\blacktriangle}$& $88.19^{\blacktriangle}$\\
    \textsf{CAL} & 38.19& 56.95& 62.64& 64.80& $\mathbf{66.58}^{\triangledown}$& $66.40^{\triangledown}$& $63.37^{\triangledown}$& $\mathbf{66.23}^{\triangledown}$& $66.10$\\
    \textsf{BMN} & 93.50& 94.75& $97.67$& $96.90^{\triangledown}$& \textbf{97.80}& $97.20^{\triangledown}$ & $96.02^{\triangledown}$& $\mathbf{97.42^{\triangledown}}$& 97.38\\
    \bottomrule
  \end{tabular}
  \label{tab:model-filter-accs}
\end{table}

\subsection{Filtering embeddings by aggregating scopes}
\label{section:rle-sca}

We now tackle embedding extraction by 
aggregating more node activations in a single feature.
Again, we strive for shorter embeddings and, at the same time, we
investigate if the {leaves do play a negligible role} as
feature extractors.
We propose to build embeddings by \emph{{averaging node
outputs having the same scope}}, leveraging the idea that all the nodes
sharing the same scope are {extracting different features for a
  single, shared,
latent factor}.
Thus computing for each possible scope $j$ in
$S$: \footnote{$S_{n}(\mathbf{x}^{i})$ values are in the $\exp$ domain
  and finally $e_{j}^{i}$ is projected in the $\log$ domain.}
\begin{equation}
  e_{j}^{i}=\frac{1}{|\{n|n\in S, \mathsf{sc}(n)=j\}|}\sum_{n\in
    \{n|n\in S, \mathsf{sc}(n)=j\}}S_{n}(\mathbf{x}^{i}). 
  \label{eq:scope-aggr}
\end{equation}
%
%
%
The question concerning which nodes to consider for each aggregation
can be answered in different ways.
Constructing embeddings according to
Eq.~\ref{eq:scope-aggr} requires collect the output of a \emph{{fictitious
complete sum node}} computing a uniform mixture
over all nodes sharing the same scope (cf. Figure~\ref{fig:spn-emb} (c)).
%
%
%
%
Hence, we decide to aggregate only sum
node outputs, since product nodes already
contribute to their sum node parent feature extractions. 
As an alternative, we propose to aggregate always by
scope both sum nodes and leaf nodes as well.
In this way we can verify if the additional information they provide
can be of some use (cf. Figure~\ref{fig:spn-emb} (l)).
{Such a scope aggregation criterion is derived from the
  recursive definition of SPNs: as a sub-network rooted at a certain
  node is a valid probabilistic model over the RVs in that node scope,
it is meaningful to look at the features extracted for each possible
scope---or \emph{resolution}---in the network.}


As one can see in Table~\ref{tab:model-scope-aggr}, leaf addition helps models with lower capacity like
\textsf{SPN-I}, scoring the best accuracy for them on
\textsf{CAL}.
As the model capacity increases, however, the contribution of leaves becomes
marginal or even zero.
Generally, aggregated embeddings are comparably accurate 
w.r.t. the best corresponding sum embeddings, while being smaller.
This  empirically confirms the utility of scope
aggregations as a heuristic to extract compact embeddings from an SPN.

%
{As a general guideline, one would seek embeddings that are as compact and as informative
as possible.
We summarize the general findings from our extensive experimental
suite.
Sum node activations alone act as
sufficient compressors for less regularized models, and as such they
shall be preferred over products.
Mid-level representations---embeddings belonging to nodes with
medium scope lengths---are  enough to preserve a good accuracy
while reducing the embedding size.
Contrary to classical deep models, only high-level representations are
somehow slightly less informative, even if they provide the best compression.
Scope aggregations prove
to be a very effective size reduction heuristic, leveraging the
recursive nature of SPNs as feature extractors, while deriving good
predictive performances.
Ultimately, our recommendation  would be to first look at scope
aggregations with SPNs, as they provide the best trade-off concerning
embedding size and informative power.
}

\begin{table}
  \caption[datasets]{Test accuracies for the embeddings
    extracted by aggregating node outputs
    with the same scope, when leaves are not counted (\textsf{no-leaves}) and when they are considered (\textsf{leaves}).
    Bold values denote significantly better scores than all the others.
    $\blacktriangle$ indicates a better score than competitor embeddings
    with greater or equal size. $\triangledown$ indicates worse
    scores than competitor embeddings with smaller or equal size.}
  \centering
  \small
  \setlength{\tabcolsep}{3pt}  
  \begin{tabular}{l llllll}
    \toprule
    & \multicolumn{2}{c}{\textsf{SPN-I}} &
                                           \multicolumn{2}{c}{\textsf{SPN-II}}
    & \multicolumn{2}{c}{\textsf{SPN-III}}\\
    
    \textsf{dataset}&\textsf{no-leaves}  & \textsf{leaves}& \textsf{no-leaves}& \textsf{leaves}& \textsf{no-leaves}& \textsf{leaves}\\
    \midrule
    \textsf{REC} & 72.47 & $75.92^{\triangledown}$ & $\mathbf{97.94}^{\blacktriangle}$ & $\mathbf{97.99}^{\blacktriangle}$ & $\mathbf{97.94}^{\blacktriangle}$ & $\mathbf{98.02}^{\blacktriangle}$ \\
    \textsf{CON} & 62.35 & $66.49^{\triangledown}$ & $77.21^{\blacktriangle}$ & 78.05 & $\mathbf{83.52}^{\blacktriangle}$ & $\mathbf{83.84}^{\blacktriangle}$ \\
    \textsf{OCR} & 74.32 & 81.85 & $89.71^{\blacktriangle}$ & $89.68^{\blacktriangle}$ & $\mathbf{89.90}^{\blacktriangle}$ & $\mathbf{89.91}^{\blacktriangle}$ \\
    \textsf{CAL} & 38.10 & $63.19^{\triangledown}$  & 62.59 & $62.76^{\triangledown}$ & $\mathbf{66.49}^{\triangledown}$ & $\mathbf{66.58}^{\triangledown}$ \\
    \textsf{BMN} & 93.51 & $94.83^{\triangledown}$ & $97.64^{\blacktriangle}$ & $97.62^{\blacktriangle}$ & $\mathbf{97.80}$ & $\mathbf{97.80}$ \\
    \bottomrule
  \end{tabular}
  \label{tab:model-scope-aggr}
\end{table}

\subsection{Semi-supervised representation learning}
\label{section:rle-semi}

Up to now, we empirically demonstrated the meaningfulness and
effectiveness of SPNs representations when plugged into 
supervised tasks. In real world scenarios, however, it is more likely that only a portion of the 
samples are labeled. 
Here we investigate whether in such a \emph{semi-supervised learning}
scenario these representations are still exploitable.
Formally, we consider a set of samples
$\{\mathbf{x}^{i}\}_{i=1}^{m}$ for
which only a reduced set of $l<m$ labels $\{y^{j}\}_{j\in\mathcal{L}}$,  $\mathcal L \subset\{i\}_{i=1}^{m}, |\mathcal{L}|=l$ is available.
From our perspective on density estimation, nothing really
changes---we will still exploit the same representation extracted on
the RVs $\mathbf{X}$---hence 
reusing the same embeddings previously generated with our reference models.

We employ the \emph{label spreading}
algorithm~\cite{Zhou2004} as the base classifier  over
all representations from all our models.
In a nutshell, $y^j$ from labeled samples are spread to
the unlabeled samples that are \emph{closer in the embedding space}.
In particular, we adopt a  $k$-nearest neighbor approach ($k=7$) to
classify samples.
We set the clamping factor to 0.2 and used up to 30 iterates to let
the label propagation process converge.
The meaningfulness of the extracted representations in this scenario is
still measured by the scored accuracy, however, it will now be more
correlated to the ability of the new geometric space to facilitate
label spreading by proximity.

To thoroughly evaluate the
meaningfulness of all embedding spaces, we repeat the classification
experiment by varying the number of available labels.
As a common scheme over the datasets, we run a
learning task allowing 1, 10, 60, 100 and 600 labeled training samples per class.
In the end, we evaluate the following learning regime: 2, 20,
120, 200, 600 labeled samples for \textsf{REC} and \textsf{CON};
26, 260, 1560, 2600, 7800 samples for \textsf{OCR}; 101, 505, 1010
 samples for \textsf{CAL}; and 10, 100, 600, 1000, 3000
labeled samples for \textsf{BMN}. We repeat each experiment ten times.

We employ embeddings from the
\textsf{RBM-5k}, \textsf{DBN-1k}, \textsf{MADE-1k} and \textsf{VAE-1k} models as
competitors, as they achieved the highest accuracies
in the supervised setting.
We use embeddings from \textsf{SPN-III} models comprising sum nodes, large scope lengths 
or scope aggregations without leaves since they provide a good compromise
between size and accuracy, as seen in Section~\ref{section:rle-sca}.
Lastly, as a baseline we run label spreading over the original feature
space $\mathbf{X}$, denoting it as \textsf{LP}.

\begin{table}
  \scriptsize
  \caption[Test set accuracy scores for 
  \textsf{SPN}, \textsf{RBM}, \textsf{MADE} and \textsf{DBN}
  embeddings for semi-supervised learning]{Mean and standard deviation
    (over ten runs) of test accuracies
for semi-supervised learning experiments with label propagation on embeddings
    extracted from the sum nodes of \textsf{SPN-III} (\textsf{sum}),
    from its nodes with large scopes (\textsf{L}) or from scope
    aggregations (\textsf{aggr}) compared against the baseline
    \textsf{LP} and  embeddings extracted from
    \textsf{RBM-5k}, \textsf{DBN-1k}, \textsf{MADE-1k} and {\textsf{VAE-1k}} models on all
    datasets and for a different number of available labels ($l$).
    Bold values denote significantly better scores
    than all the others for a dataset and a certain number of labeled
    examples.}
 \setlength{\tabcolsep}{2pt}  
  \begin{tabular}{ll c@{\hskip 3pt} c c c@{\hskip 3pt} c @{\hskip 3pt} c @{\hskip 3pt} c@{\hskip 3pt} c }
    \toprule

     &&  & \multicolumn{3}{c@{\hskip 8pt}}{\textsf{SPN-III}}
    & \textsf{RBM} &\textsf{DBN} & \textsf{MADE} & {\textsf{VAE}}\\
    &$l$& \textsf{LP}&\scriptsize\textsf{sum}&\scriptsize\textsf{L} & \scriptsize\textsf{aggr}
    & \scriptsize\textsf{5k} & \scriptsize\textsf{1k}&\scriptsize\textsf{1k}&\scriptsize{\textsf{1k}}\\

     \midrule
    \multirow{5}{*}{\rotatebox[origin=c]{90}{\textsf{REC}}}&\textbf{2}&$50.73 {\scriptstyle\pm 3.7}$&$65.17 {\scriptstyle\pm 12.2}$&$61.51 {\scriptstyle\pm 10.0}$&$\mathbf{65.53} {\scriptstyle\pm 11.7}$&$53.19 {\scriptstyle\pm 2.9}$&$56.75 {\scriptstyle\pm 6.6}$&$51.76 {\scriptstyle\pm 4.4}$&$54.99 {\scriptstyle\pm 9.5}$\\
    &\textbf{20}&$63.18 {\scriptstyle\pm 8.1}$&$79.16 {\scriptstyle\pm 4.5}$&$\mathbf{80.77} {\scriptstyle\pm 3.9}$&$78.73 {\scriptstyle\pm 4.3}$&$66.87 {\scriptstyle\pm 5.6}$&$77.92 {\scriptstyle\pm 4.6}$&$58.78 {\scriptstyle\pm 2.8}$&$65.27 {\scriptstyle\pm 4.4}$\\
    &\textbf{120}&$81.25 {\scriptstyle\pm 6.2}$&$88.89 {\scriptstyle\pm 2.2}$&$\mathbf{90.97} {\scriptstyle\pm 1.0}$&$88.95 {\scriptstyle\pm 2.2}$&$84.46 {\scriptstyle\pm 2.7}$&$90.34 {\scriptstyle\pm 2.3}$&$68.56 {\scriptstyle\pm 3.5}$&$79.23 {\scriptstyle\pm 3.7}$\\
    &\textbf{200}&$84.95 {\scriptstyle\pm 5.5}$&$90.38 {\scriptstyle\pm 1.2}$&$\mathbf{91.77} {\scriptstyle\pm 1.7}$&$90.51 {\scriptstyle\pm 1.2}$&$86.90 {\scriptstyle\pm 5.4}$&$\mathbf{92.62} {\scriptstyle\pm 1.1}$&$70.97 {\scriptstyle\pm 2.3}$&$83.20 {\scriptstyle\pm 3.1}$\\
    &\textbf{600}&$85.89 {\scriptstyle\pm 6.2}$&$87.83 {\scriptstyle\pm 0.9}$&$92.45 {\scriptstyle\pm 0.9}$&$87.68 {\scriptstyle\pm 1.0}$&$88.09 {\scriptstyle\pm 4.6}$&$\mathbf{93.36} {\scriptstyle\pm 0.6}$&$71.83 {\scriptstyle\pm 3.9}$&$81.97 {\scriptstyle\pm 2.9}$\\
    \midrule
    \multirow{5}{*}{\rotatebox[origin=c]{90}{\textsf{CON}}}&\textbf{2}&$\mathbf{49.83} {\scriptstyle\pm 0.5}$&$48.66 {\scriptstyle\pm 1.5}$&$49.11 {\scriptstyle\pm 1.0}$&$48.63 {\scriptstyle\pm 1.4}$&$49.58 {\scriptstyle\pm 0.4}$&$48.97 {\scriptstyle\pm 1.1}$&$49.26 {\scriptstyle\pm 0.7}$&$49.43 {\scriptstyle\pm 1.5}$\\
    &\textbf{20}&$49.82 {\scriptstyle\pm 0.5}$&$\mathbf{51.10} {\scriptstyle\pm 0.9}$&$50.51 {\scriptstyle\pm 0.7}$&$\mathbf{51.11} {\scriptstyle\pm 0.9}$&$49.92 {\scriptstyle\pm 0.4}$&$50.18 {\scriptstyle\pm 0.6}$&$50.43 {\scriptstyle\pm 0.6}$&$50.37 {\scriptstyle\pm 1.2}$\\
    &\textbf{120}&$49.82 {\scriptstyle\pm 0.3}$&$\mathbf{52.99} {\scriptstyle\pm 1.1}$&$52.09 {\scriptstyle\pm 0.7}$&$\mathbf{53.02} {\scriptstyle\pm 1.2}$&$49.91 {\scriptstyle\pm 0.3}$&$50.29 {\scriptstyle\pm 0.4}$&$51.16 {\scriptstyle\pm 0.6}$&$50.88 {\scriptstyle\pm 0.9}$\\
    &\textbf{200}&$49.86 {\scriptstyle\pm 0.2}$&$\mathbf{53.54} {\scriptstyle\pm 1.0}$&$52.47 {\scriptstyle\pm 0.9}$&$\mathbf{53.55} {\scriptstyle\pm 1.0}$&$49.96 {\scriptstyle\pm 0.3}$&$50.38 {\scriptstyle\pm 0.5}$&$51.35 {\scriptstyle\pm 0.8}$&$51.15 {\scriptstyle\pm 0.8}$\\
    &\textbf{600}&$49.84 {\scriptstyle\pm 0.2}$&$\mathbf{54.73} {\scriptstyle\pm 0.5}$&$53.58 {\scriptstyle\pm 0.6}$&$\mathbf{54.75} {\scriptstyle\pm 0.5}$&$50.03 {\scriptstyle\pm 0.2}$&$50.25 {\scriptstyle\pm 0.2}$&$51.92 {\scriptstyle\pm 0.5}$&$52.30 {\scriptstyle\pm 0.9}$\\
    \midrule
    \multirow{5}{*}{\rotatebox[origin=c]{90}{\textsf{OCR}}}&\textbf{26}&$29.51 {\scriptstyle\pm 4.1}$&$\mathbf{40.69} {\scriptstyle\pm 3.4}$&$\mathbf{41.00} {\scriptstyle\pm 3.6}$&$\mathbf{41.23} {\scriptstyle\pm 3.0}$&$38.74 {\scriptstyle\pm 4.4}$&$37.70 {\scriptstyle\pm 3.5}$&$20.69 {\scriptstyle\pm 3.3}$&$36.08 {\scriptstyle\pm 3.4}$\\
    &\textbf{260}&$50.29 {\scriptstyle\pm 1.3}$&$\mathbf{65.45} {\scriptstyle\pm 1.9}$&$\mathbf{66.61} {\scriptstyle\pm 1.5}$&$\mathbf{66.06} {\scriptstyle\pm 1.8}$&$62.67 {\scriptstyle\pm 1.6}$&$63.48 {\scriptstyle\pm 1.5}$&$37.22 {\scriptstyle\pm 1.3}$&$61.10 {\scriptstyle\pm 1.2}$\\
    &\textbf{1560}&$62.75 {\scriptstyle\pm 0.9}$&$\mathbf{76.29} {\scriptstyle\pm 0.4}$&$\mathbf{76.97} {\scriptstyle\pm 0.3}$&$\mathbf{76.27} {\scriptstyle\pm 0.4}$&$73.61 {\scriptstyle\pm 0.3}$&$74.67 {\scriptstyle\pm 0.3}$&$52.37 {\scriptstyle\pm 0.9}$&$72.32 {\scriptstyle\pm 0.3}$\\
    &\textbf{2600}&$65.19 {\scriptstyle\pm 0.4}$&$78.50 {\scriptstyle\pm 0.4}$&$\mathbf{79.24} {\scriptstyle\pm 0.2}$&$78.86 {\scriptstyle\pm 0.3}$&$75.94 {\scriptstyle\pm 0.4}$&$76.98 {\scriptstyle\pm 0.4}$&$56.20 {\scriptstyle\pm 0.8}$&$74.74 {\scriptstyle\pm 0.3}$\\
    &\textbf{7800}&$69.10 {\scriptstyle\pm 0.4}$&$81.93 {\scriptstyle\pm 0.3}$&$\mathbf{82.48} {\scriptstyle\pm 0.3}$&$\mathbf{82.15} {\scriptstyle\pm 0.4}$&$79.54 {\scriptstyle\pm 0.1}$&$80.14 {\scriptstyle\pm 0.3}$&$62.65 {\scriptstyle\pm 0.2}$&$78.95 {\scriptstyle\pm 0.4}$\\
    \midrule
    \multirow{3}{*}{\rotatebox[origin=c]{90}{\textsf{CAL}}}&\textbf{101}&$33.94 {\scriptstyle\pm 4.6}$&$37.21 {\scriptstyle\pm 9.5}$&$\mathbf{39.71} {\scriptstyle\pm 3.6}$&$37.50 {\scriptstyle\pm 3.6}$&$28.84 {\scriptstyle\pm 3.7}$&$36.06 {\scriptstyle\pm 3.4}$&$37.67 {\scriptstyle\pm 4.0}$&$36.79 {\scriptstyle\pm 5.0}$\\
    &\textbf{505}&$50.19 {\scriptstyle\pm 0.6}$&$51.43 {\scriptstyle\pm 1.0}$&$52.09 {\scriptstyle\pm 0.8}$&$51.40 {\scriptstyle\pm 1.0}$&$44.48 {\scriptstyle\pm 1.1}$&$51.69 {\scriptstyle\pm 0.7}$&$52.39 {\scriptstyle\pm 0.7}$&$\mathbf{53.47}{\scriptstyle\pm 1.4}$\\
    &\textbf{1010}&$51.46 {\scriptstyle\pm 0.7}$&$54.56 {\scriptstyle\pm 0.4}$&$54.69 {\scriptstyle\pm 0.6}$&$54.38 {\scriptstyle\pm 0.5}$&$46.58 {\scriptstyle\pm 0.9}$&$53.06 {\scriptstyle\pm 0.7}$&$53.76 {\scriptstyle\pm 0.7}$&$\mathbf{56.03} {\scriptstyle\pm 0.6}$\\
    \midrule
    \multirow{5}{*}{\rotatebox[origin=c]{90}{\textsf{BMN}}}&\textbf{10}&$47.43 {\scriptstyle\pm 8.6}$&$\mathbf{61.70} {\scriptstyle\pm 9.5}$&$58.76 {\scriptstyle\pm 9.1}$&$60.43 {\scriptstyle\pm 8.4}$&$59.46 {\scriptstyle\pm 10.8}$&$\mathbf{61.71} {\scriptstyle\pm 9.7}$&$48.22 {\scriptstyle\pm 10.2}$&$54.30 {\scriptstyle\pm 10.6}$\\
    &\textbf{100}&$79.27 {\scriptstyle\pm 4.1}$&$89.98 {\scriptstyle\pm 1.3}$&$87.86 {\scriptstyle\pm 1.2}$&$88.90 {\scriptstyle\pm 1.1}$&$88.98 {\scriptstyle\pm 2.0}$&$90.68 {\scriptstyle\pm 1.0}$&$82.92 {\scriptstyle\pm 2.7}$&$\mathbf{91.04} {\scriptstyle\pm 2.4}$\\
    &\textbf{600}&$90.31 {\scriptstyle\pm 0.6}$&$94.36 {\scriptstyle\pm 0.2}$&$93.37 {\scriptstyle\pm 0.3}$&$93.65 {\scriptstyle\pm 0.3}$&$94.04 {\scriptstyle\pm 0.4}$&$94.45 {\scriptstyle\pm 0.1}$&$89.68 {\scriptstyle\pm 0.8}$&$\mathbf{94.93}{\scriptstyle\pm 0.4}$\\
    &\textbf{1000}&$91.15 {\scriptstyle\pm 0.5}$&$94.72 {\scriptstyle\pm 0.2}$&$93.96 {\scriptstyle\pm 0.3}$&$94.14 {\scriptstyle\pm 0.3}$&$94.32 {\scriptstyle\pm 0.3}$&$94.81 {\scriptstyle\pm 0.2}$&$90.44 {\scriptstyle\pm 0.6}$&$\mathbf{95.29}{\scriptstyle\pm 0.3}$\\
    &\textbf{3000}&$92.29 {\scriptstyle\pm 0.2}$&$95.21 {\scriptstyle\pm 0.1}$&$94.73 {\scriptstyle\pm 0.1}$&$94.75 {\scriptstyle\pm 0.1}$&$94.86 {\scriptstyle\pm 0.1}$&$95.11 {\scriptstyle\pm 0.1}$&$91.57 {\scriptstyle\pm 0.3}$&$\mathbf{95.85} {\scriptstyle\pm 0.1}$\\
    \bottomrule
   
  \end{tabular}

\label{tab:model-accs-ssl}
\end{table}

\begin{figure}[!t]
  {\hspace{65pt}\small\textbf{\textsf{REC}}}\hspace{90pt}{\small\textbf{\textsf{OCR}}}\hspace{90pt}{\small\textbf{\textsf{BMN}}}\\[5pt]
  {\raisebox{30pt}{\rotatebox[origin=c]{90}{\small\textbf{\textsf{original}}}}\hspace{5pt}\includegraphics[width=0.3\columnwidth]
    {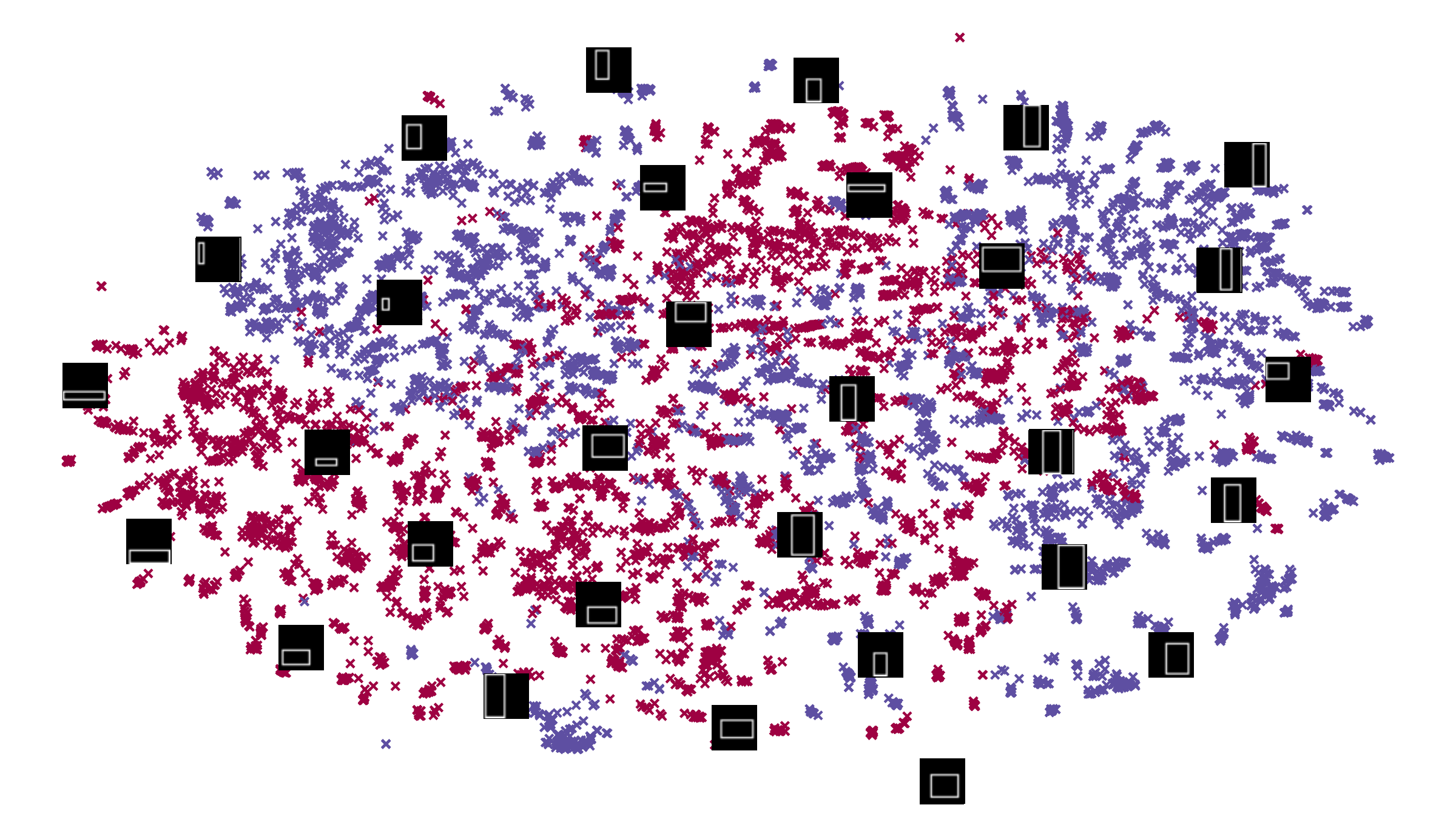}
    \label{fig:rec-lr-tsne}}
  {\includegraphics[width=0.3\columnwidth]
    {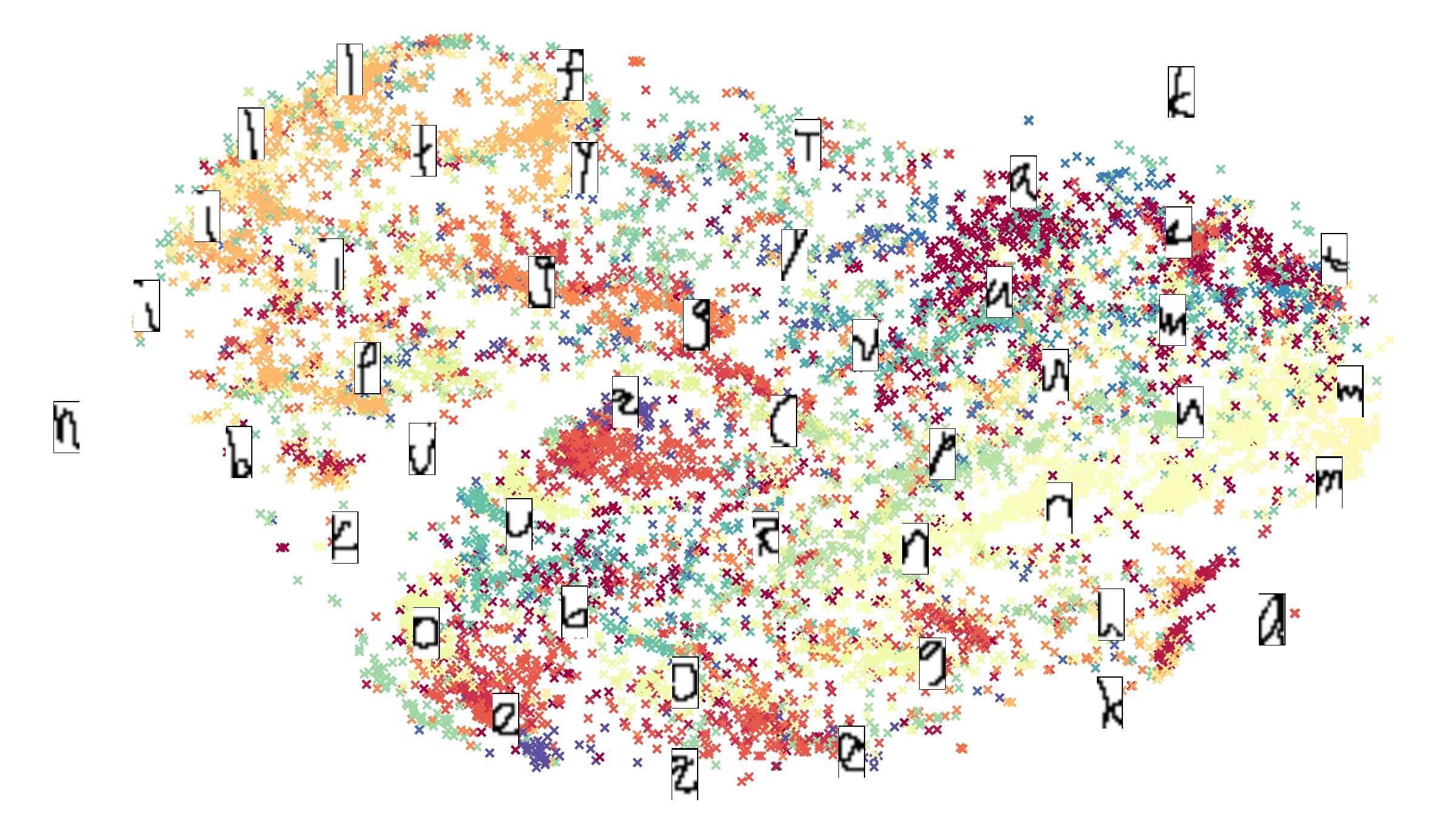}
    \label{fig:ocr-lr-tsne}}
  {\includegraphics[width=0.3\columnwidth]
    {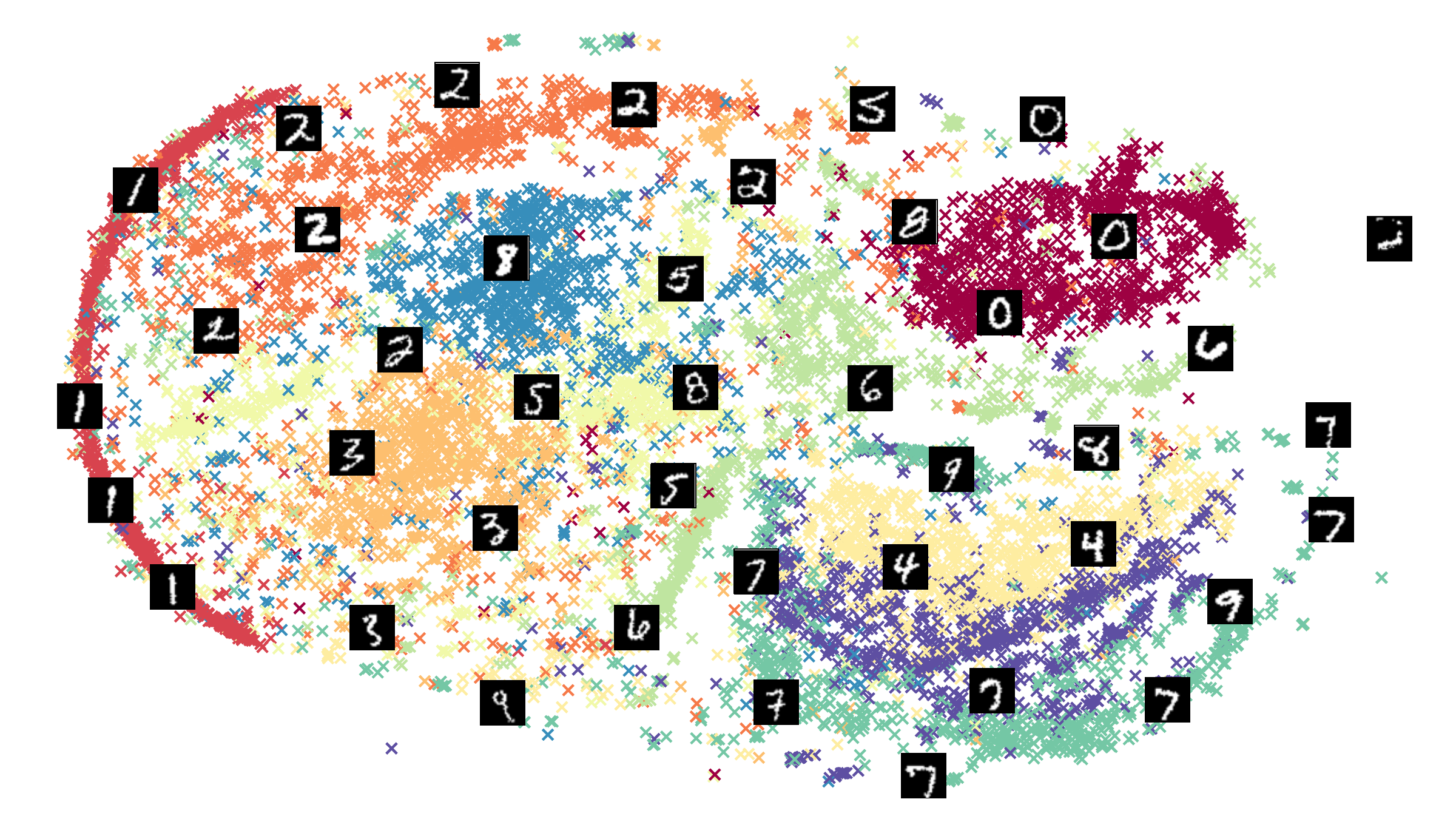}
    \label{fig:bmn-lr-tsne}}\\[1pt]
  {\raisebox{30pt}{\rotatebox[origin=c]{90}{\small\textbf{\textsf{SPN}}}}\hspace{5pt}\includegraphics[width=0.3\columnwidth]
    {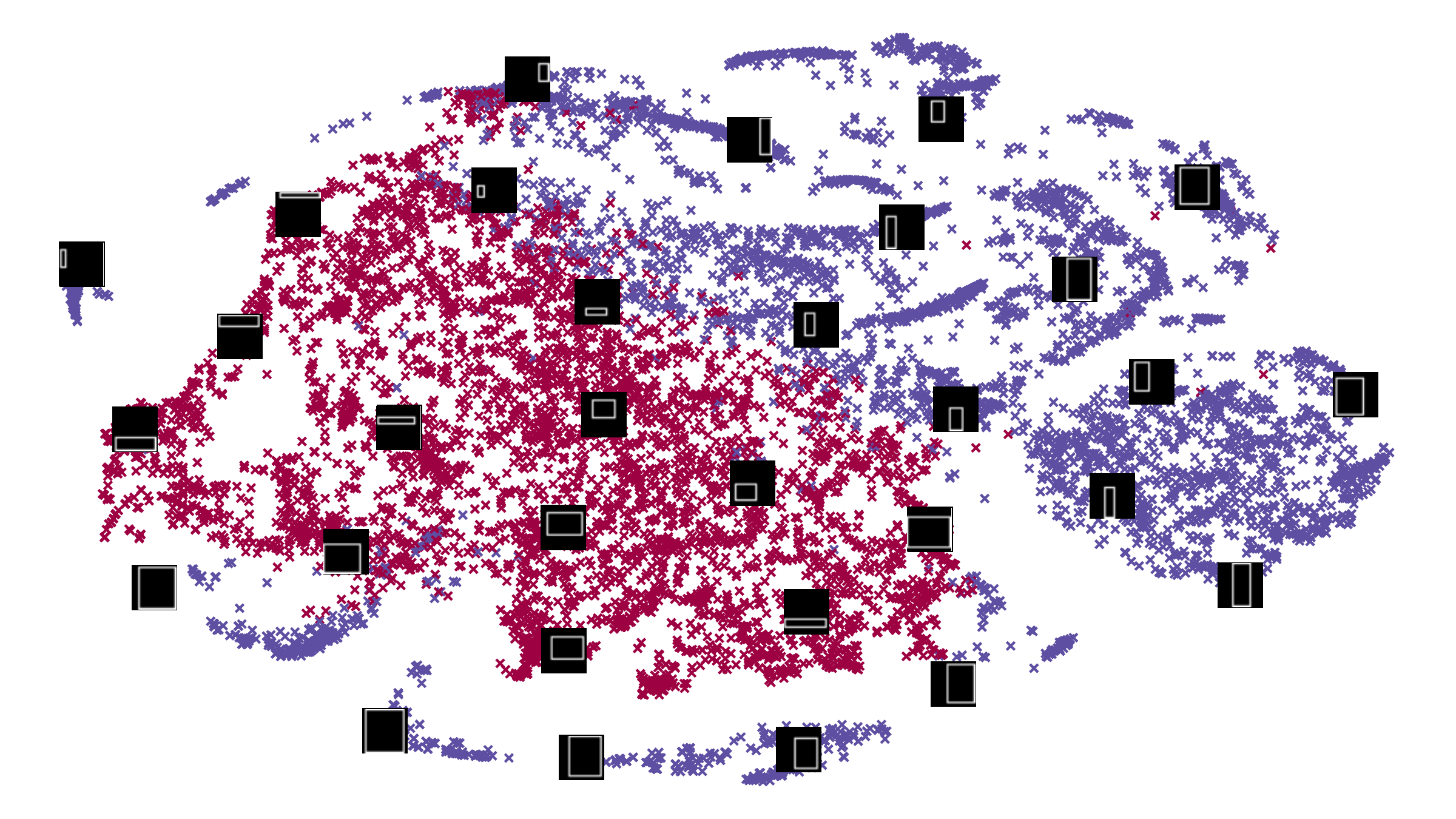}
    \label{fig:rec-spn-tsne}}
  {\includegraphics[width=0.3\columnwidth]
    {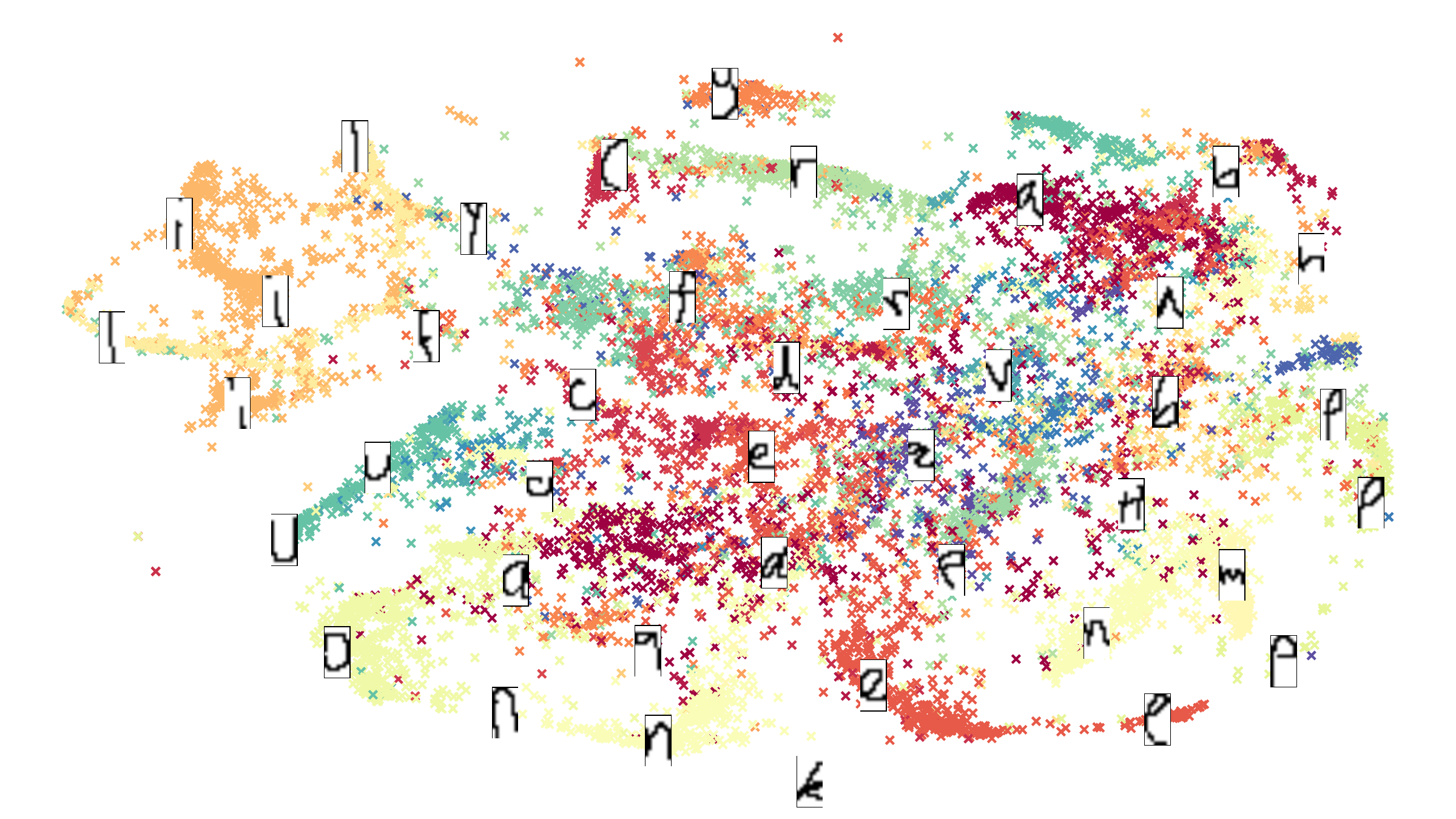}
    \label{fig:ocr-spn-tsne}}
  {\includegraphics[width=0.3\columnwidth]
    {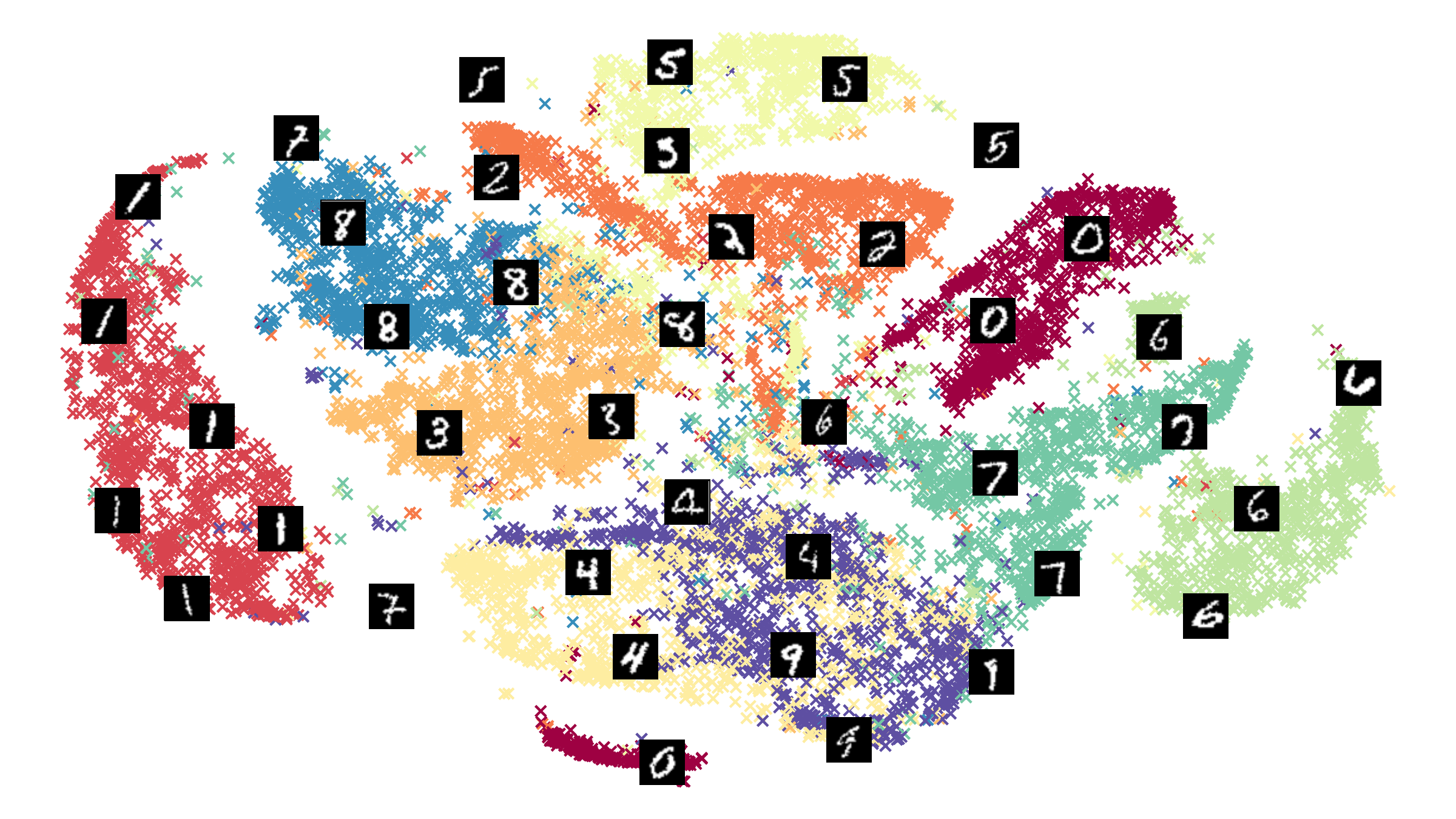}
    \label{fig:bmn-spn-tsne}}\\[1pt]
  {\raisebox{30pt}{\rotatebox[origin=c]{90}{\small\textbf{\textsf{MADE
          1k}}}}\hspace{5pt}\includegraphics[width=0.3\columnwidth]
    {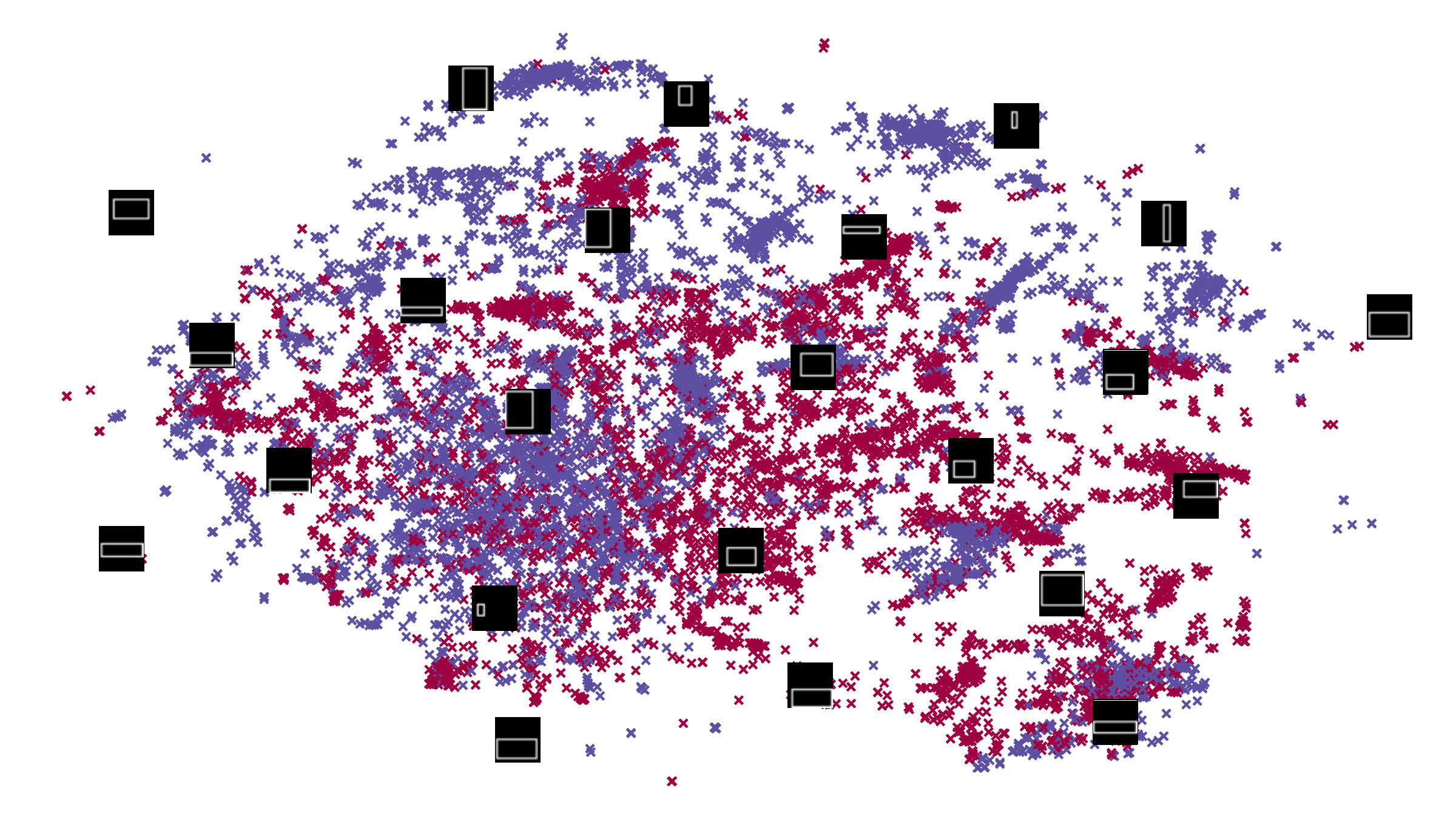}
    \label{fig:rec-made-tsne}}
  {\includegraphics[width=0.3\columnwidth]
    {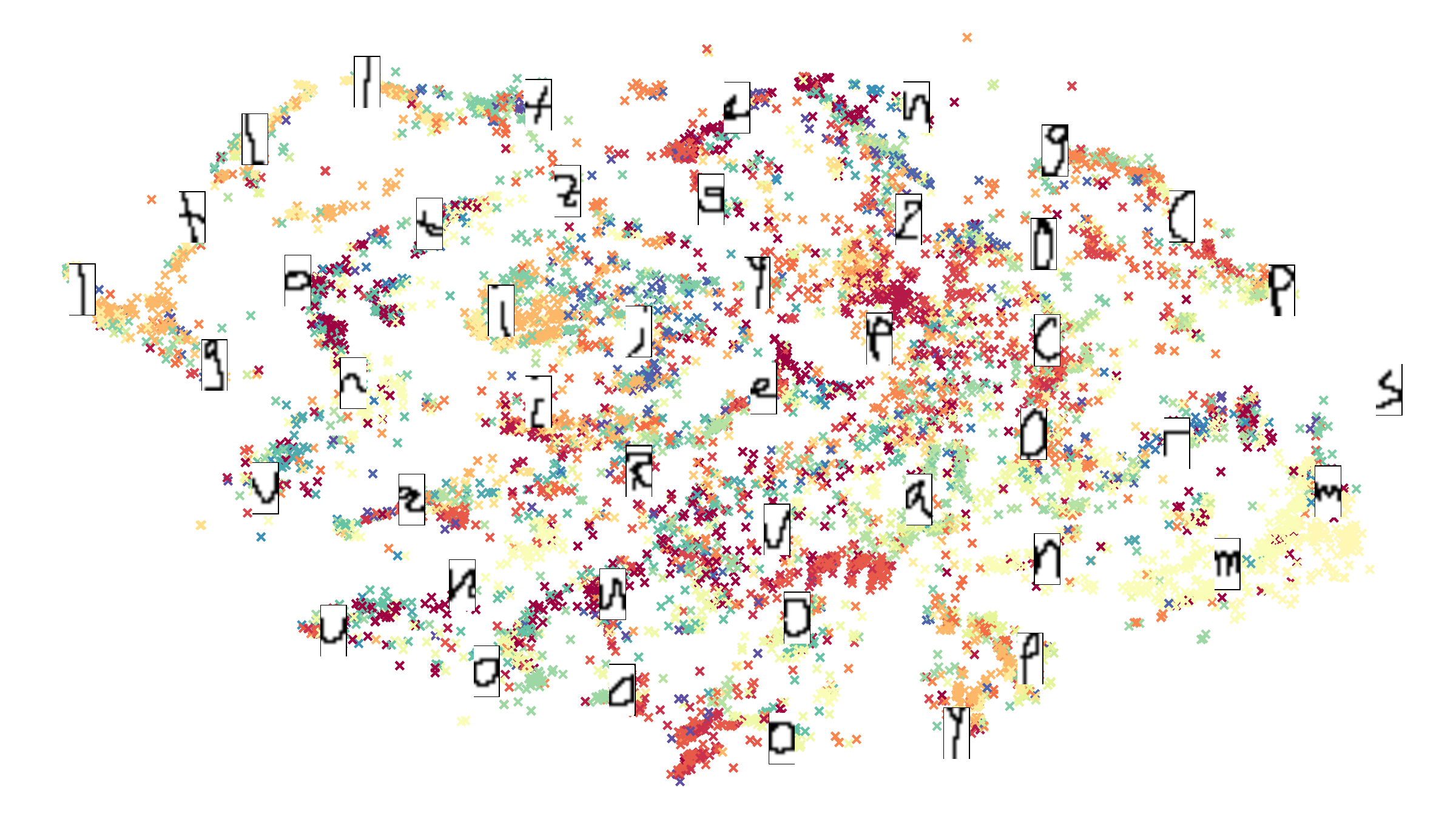}
    \label{fig:ocr-made-tsne}}
  {\includegraphics[width=0.3\columnwidth]
    {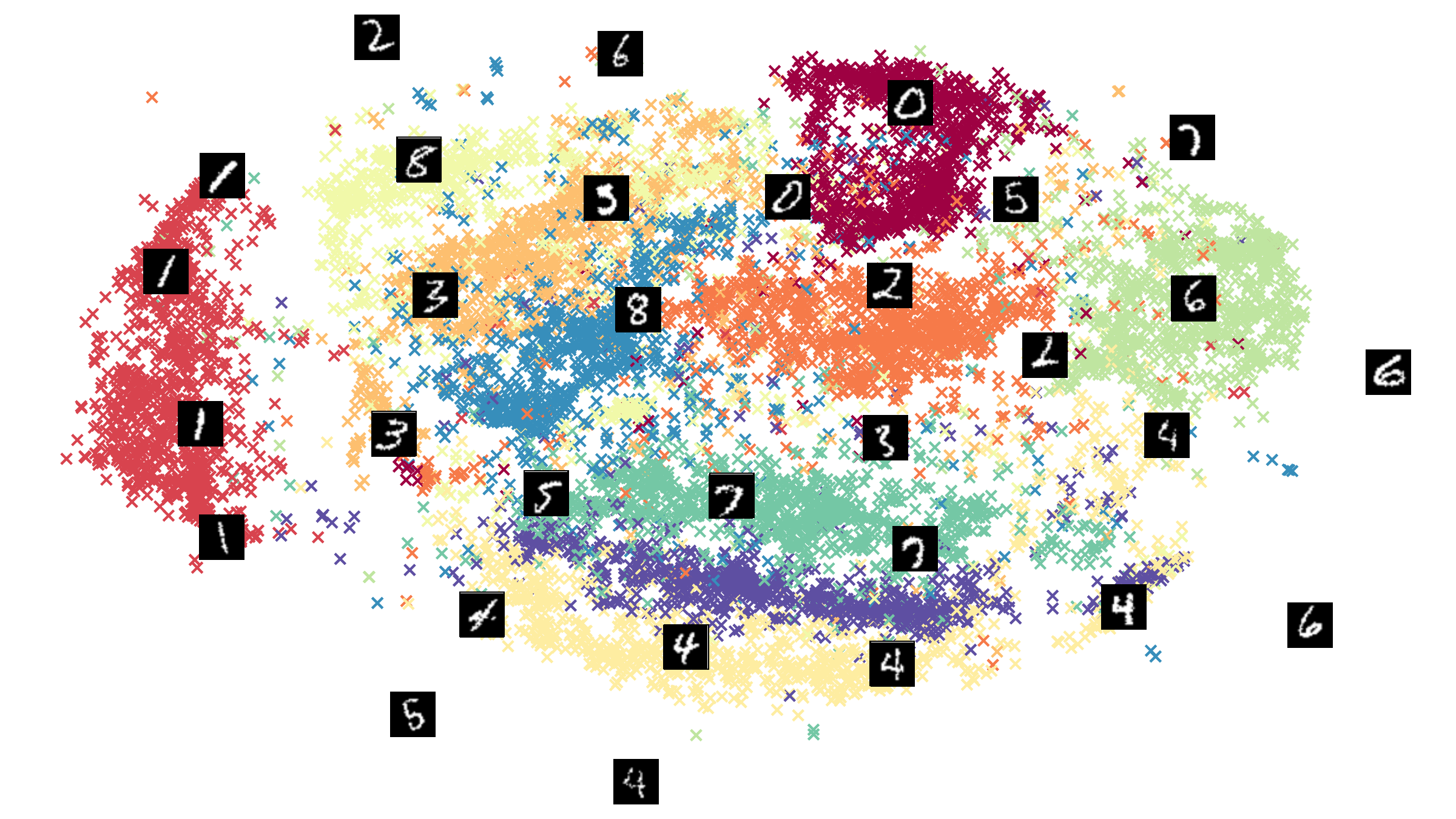}
    \label{fig:bmn-made-tsne}}\\[1pt]
  {\raisebox{30pt}{\rotatebox[origin=c]{90}{\small\textbf{\textsf{DBN 1k}}}}\hspace{5pt}\includegraphics[width=0.3\columnwidth]
    {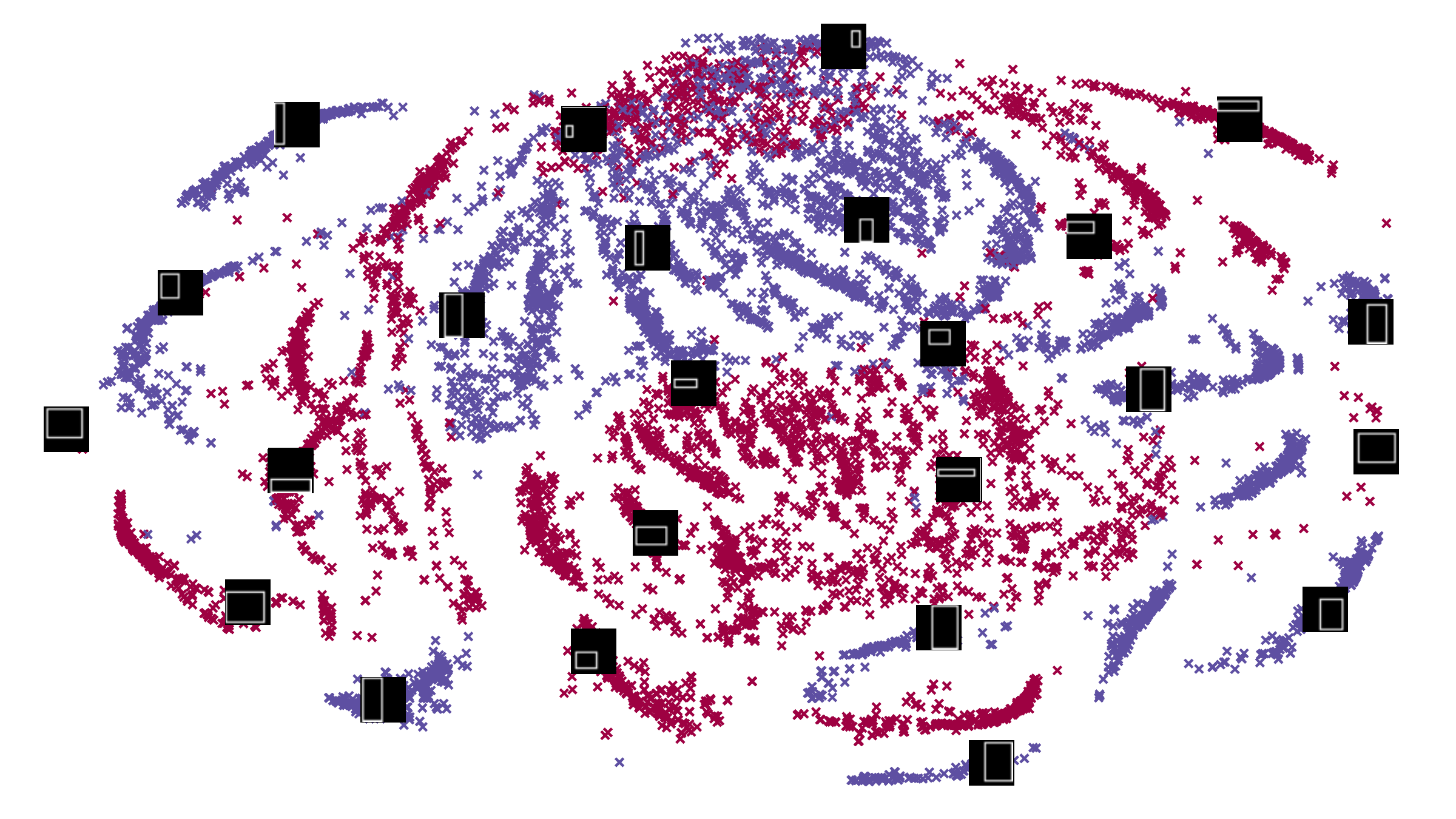}
    \label{fig:rec-dbn-tsne}}
  {\includegraphics[width=0.3\columnwidth]
    {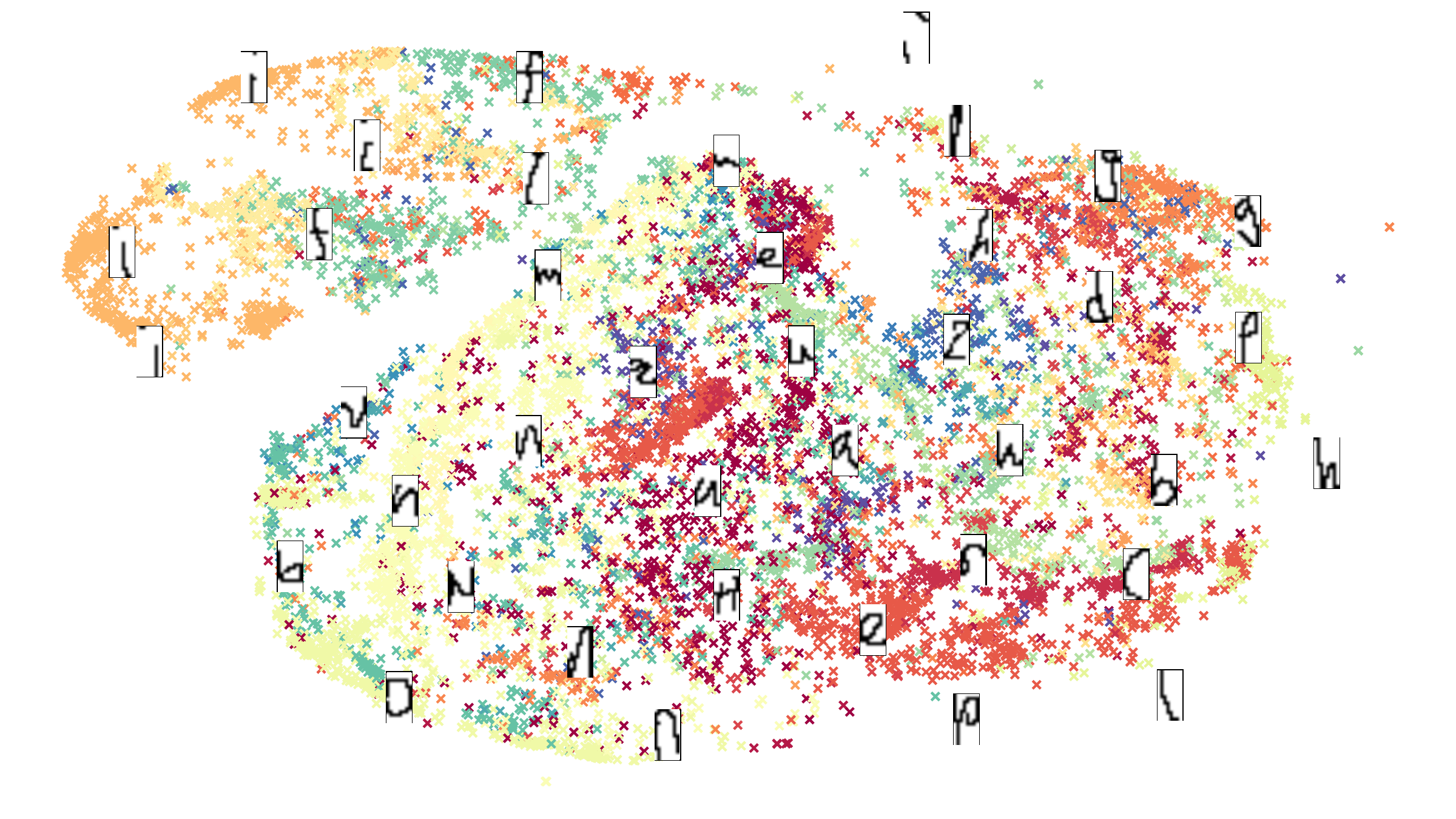}
    \label{fig:ocr-dbn-tsne}}
  {\includegraphics[width=0.3\columnwidth]
    {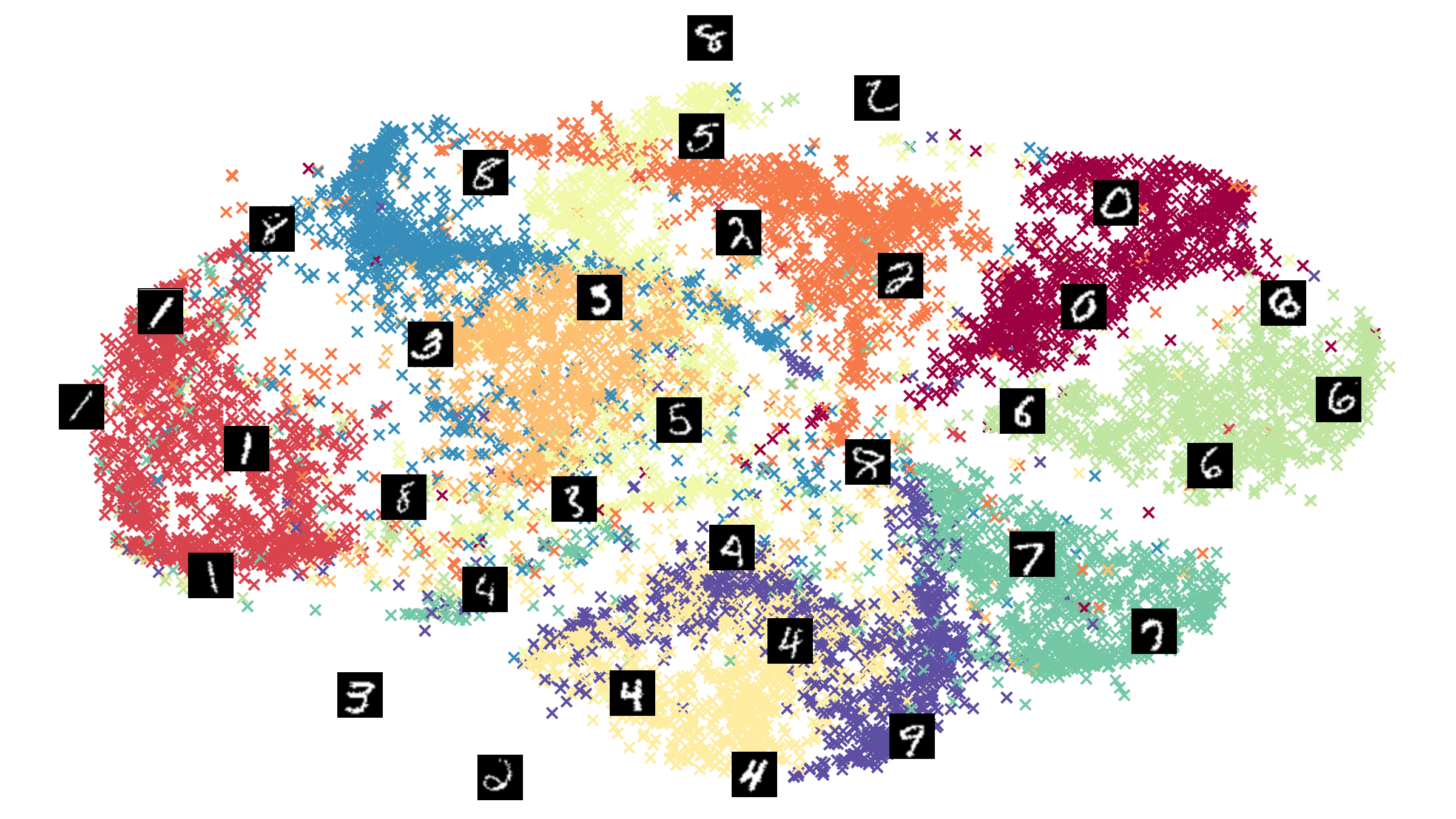}
    \label{fig:bmn-dbn-tsne}}\\[1pt]
  {\raisebox{30pt}{\rotatebox[origin=c]{90}{\small\textbf{\textsf{VAE 1k}}}}\hspace{5pt}\includegraphics[width=0.3\columnwidth]
    {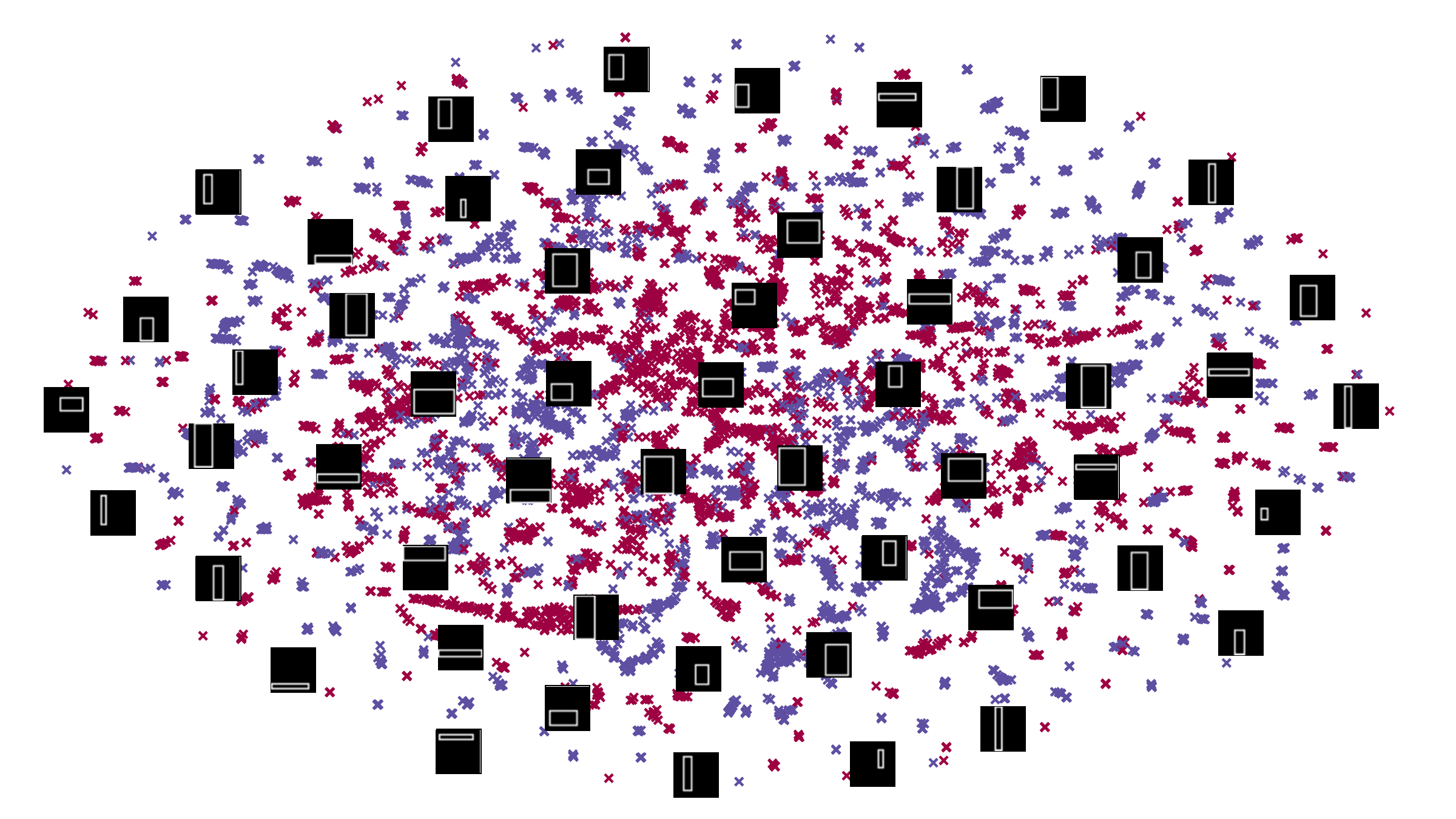}
    \label{fig:rec-vae-tsne}}
  {\includegraphics[width=0.3\columnwidth]
    {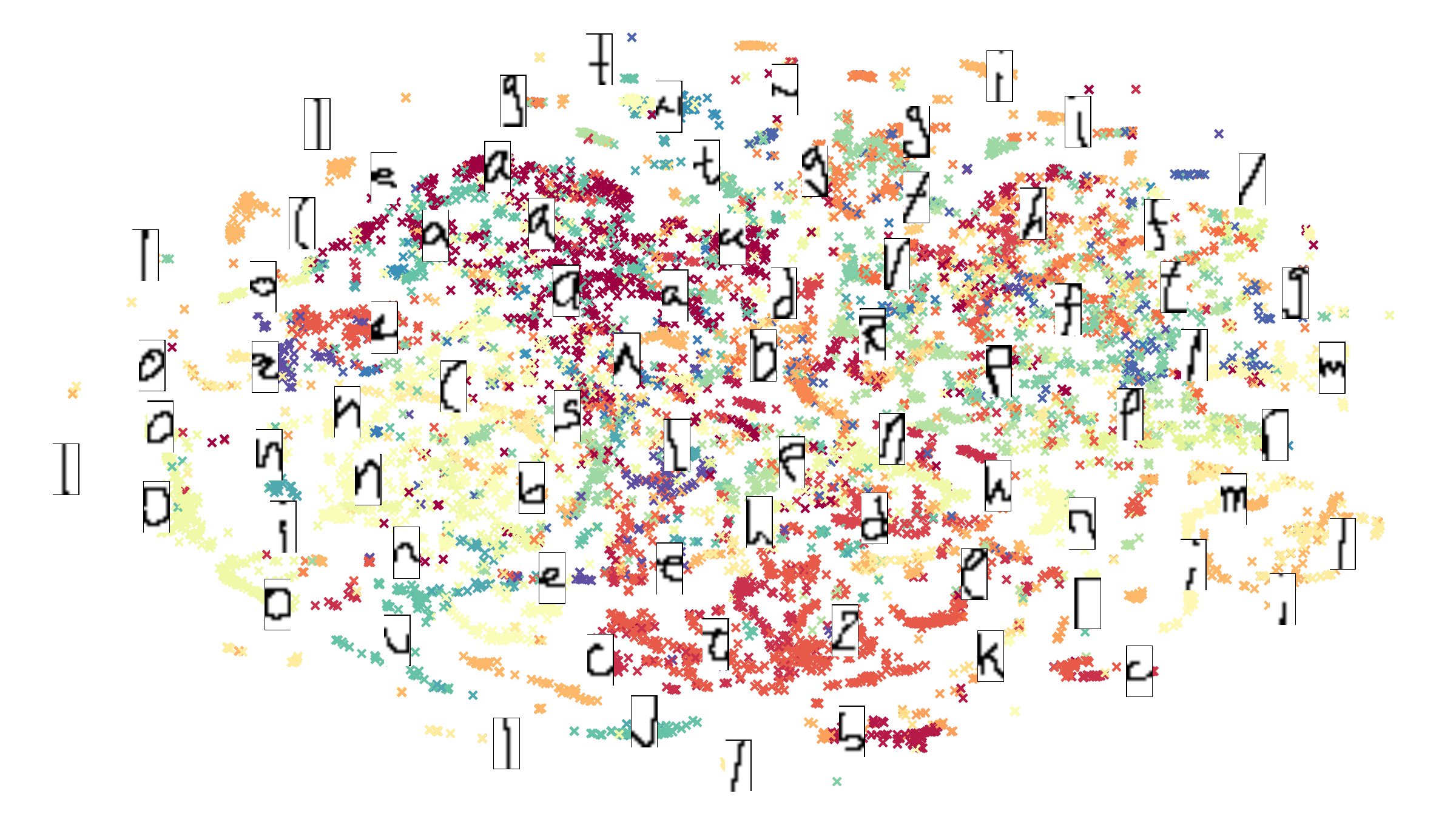}
    \label{fig:ocr-vae-tsne}}
  {\includegraphics[width=0.3\columnwidth]
    {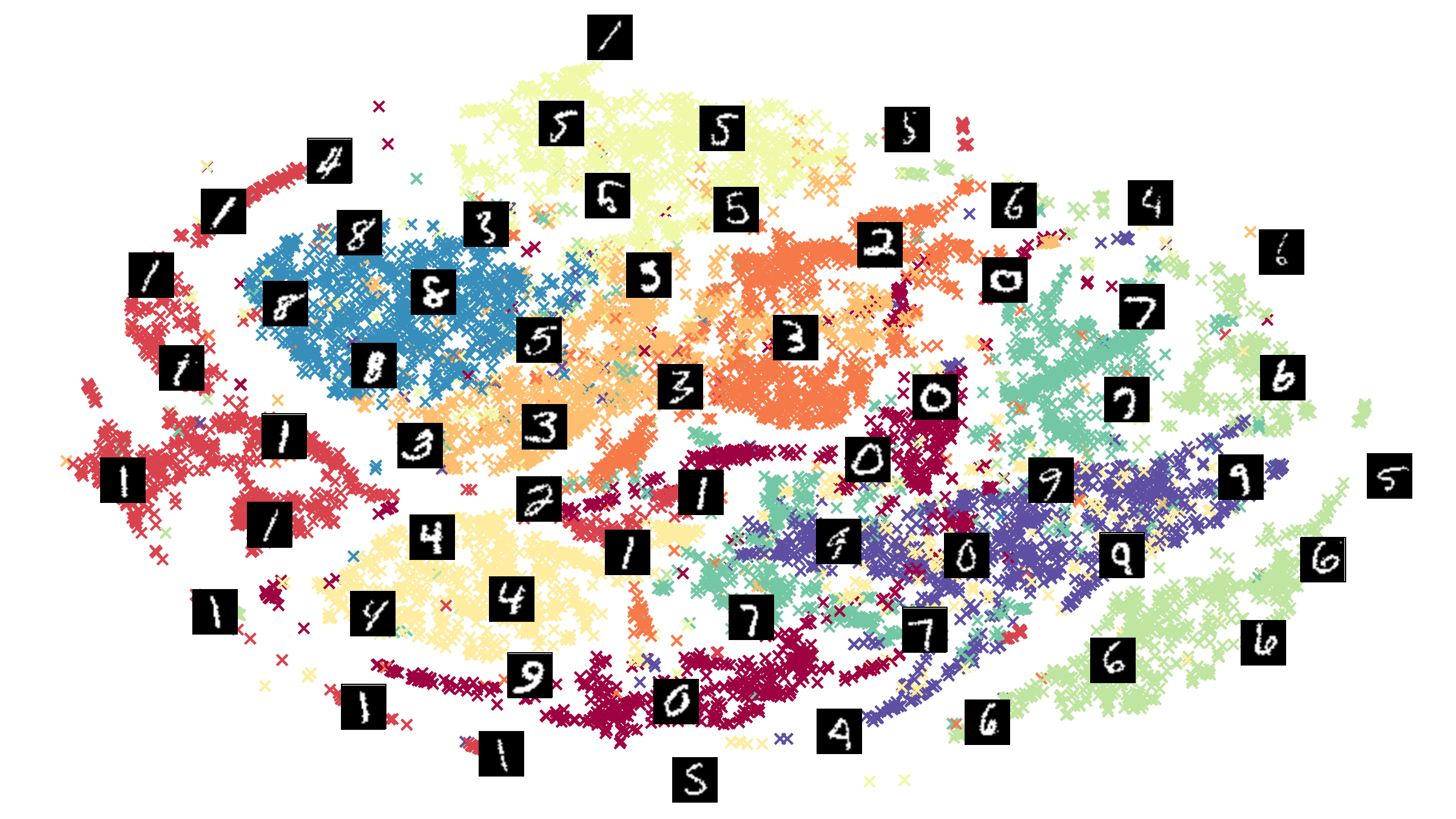}
    \label{fig:bmn-vae-tsne}}\\[1pt]
  \caption[Visualizing embeddings extracted from \textsf{SPNs},
  \textsf{RBMs}, \textsf{MADEs} and \textsf{DBNs} on \textsf{REC},
  \textsf{CON} and \textsf{OCR}]{
\emph{t-SNE plots.} The 2-d t-SNE plots of the \textsf{SPN-III} \textsf{L}arge scope embeddings against those from
\textsf{MADE-1k}, \textsf{DBN-1k}, {\textsf{VAE-1k}} and the
  original data for  
    \textsf{REC} (left),
    \textsf{CON} (center) and
    \textsf{BMN} (right). Colors indicate samples
    from different classes. Miniatures represent image samples.}
  \label{fig:tsne}
\end{figure}

As reported in Table~\ref{tab:model-accs-ssl}, 
SPNs embeddings 
provide a significant improvement over the baseline \textsf{LP}
leveraging the original features, going from random guessing ($50.73\%$)  to
$65.63\%$ by only using 2 labeled samples on the
\textsf{REC} dataset.
Compared to all other embeddings, the SPN ones are very competitive,
generally providing significantly better accuracy scores when labels
are very scarce.
In particular, they are competitive or statistically comparable to the
second best ones---\textsf{DBN-1k} representations.
{Only on \textsf{BMN}, does \textsf{VAE-1k} achieve a higher accuracy when
enough labels are provided.
Compared to the result in the supervised case, we can
argue that the manifold learned by VAEs on \textsf{BMN} is potentially
smoother, even if less separable class-wise.
While this is somehow expected by VAEs, since they are trained to
optimize a loss tailored towards such latent representations, the fact
that the generatively learned SPNs generally achieve similar
performance on \textsf{BMN} and even perform slightly better with on very few
labels is quite remarkable.}

The \textsf{CON} dataset has proved to be hard for all methods,
with no improvements from random guessing, with the exception
of SPN embeddings,
which are the only one yielding a slight
 accuracy improvement to $54.75\%$.
Concerning the filtering or aggregation criterion employed,
all  demonstrate the ability to create a meaningful
geometric space in which distances implicitly favor classification, even if class
information was unavailable when the density estimators were learned.
How well same-class samples are reachable by proximity is visible in Figure~\ref{fig:tsne}.
For instance, on \textsf{REC} the two red and blue classes are represented by two giant
``well connected'' components in the case of \textsf{SPN-III},
while for \textsf{DBN-1k} they are more fragmented,
thus explaining why SPN embedded spaces require a fewer number of
labeled examples to perform classification more accurately.

\subsection{Experimental wrap-up}
{In this Section, through our extensive set of
  experiments on (semi-)supervised tasks we definitely confirmed the
  usefulness of SPNs as tools for RL---embeddings extracted from SPNs
  have been proven to be competitive against those from RBMs, DBNs,
  MADEs and VAEs.
  The main advantage these tractable probabilistic models provide,
  w.r.t. all other competitors, is that, at the end of the day,
  one can still exploit the same model to perform exact inference for a wide
  range of queries, along employing it to extract informative feature
  representations.
  
  Concerning the reason of the effectiveness of these representations when employed
  in predictive tasks, again, one has to look at how SPNs are learned:
  structure learning as performing a form of hierarchical
  co-clustering (see Section~\ref{section:spn}).}
  It is definitely interesting evaluating how different structure
  learning algorithms could lead to different representations and how
  well these would perform in predictive tasks.

%


\section{Conclusions}
\label{section:con}

In this work we investigated how the internal representations learned by
SPNs can be understood, extracted and exploited.
We did that through visualization techniques exploiting
the peculiarity of inference and structure in SPNs.
We interpreted them as peculiar MLPs and extended their use to RL.
For this purpose, we devised several embedding extraction schemes, after noting
how classical layer or depth-wise criteria for SPNs are inadequate,
evaluating 
 their meaningfulness in a series of (semi-)supervised
classification task.
Concerning \textbf{Q1} and \textbf{Q2}, 
%
we confirmed the meaningfulness of a
scope length heuristics to correlate a node feature abstraction level
both visually and experimentally.
%
We investigated the impact of the learned structure on network
inference and on the learned representations.
Sum embeddings have been demonstrated to provide the best size versus
accuracy compromise, as well as scope aggregations.
Concerning \textbf{Q3}, the embedding extracted from SPNs have been proven to be
competitive against those from solid feature extractors such
as RBMs, DBNs, MADEs and VAEs.
%
All in all,
we provided a better understanding of the inner workings of SPNs
by uncovering what are their learned representations,
and how to effectively exploit them.
{As a result, we also provided alternative ways---to the classical
log-likelihood comparison---to assess the value of a learned SPN by visualizing and exploiting its inner representations.}

\bibliographystyle{spbasic}
\bibliography{referomnia}



\end{document}